\documentclass[journal]{IEEEtran}
\ifCLASSINFOpdf
\else
\fi

\usepackage{graphicx}
\usepackage{graphicx}
\usepackage{amsmath}
\usepackage{amssymb}
\usepackage{longtable}
\usepackage{bookmark}
\usepackage{amsthm}

\usepackage{mystyle}
\usepackage{tabularx}
\usepackage{arydshln}
\usepackage[normalem]{ulem}
\usepackage[export]{adjustbox}
\usepackage{subcaption}
\usepackage{resizegather}

\usepackage{bm}

\usepackage{float}
\usepackage{multirow}
\usepackage{xcolor,colortbl}
\usepackage{color}
\usepackage[linesnumbered,ruled]{algorithm2e}
\usepackage[skip=2pt]{caption}
\usepackage{wrapfig}
\usepackage{amsfonts}
\usepackage{tikz}

\usepackage{hyperref}
\usepackage[capitalize]{cleveref}

\providecommand{\realnum}{\mathbb{R}}
\providecommand{\naturalnum}{\mathbb{N}}
\providecommand{\bmcal}[1]{\bm{\mathcal{#1}}}

 \providecommand{\matnot}[1]{_{[{#1}]}}  
 \providecommand{\invar}{z}  
\providecommand{\binvar}{\bm{\invar}}  
\providecommand{\outvar}{x}  
\providecommand{\boutvar}{\bm{\outvar}}  

\begin{document}
\title{Tensor Methods in Computer Vision and\\Deep Learning}

\author{Yannis Panagakis$^{*}$, Jean Kossaifi$^{*}$, Grigorios G. Chrysos, James Oldfield, Mihalis A. Nicolaou, \\ Anima Anandkumar and Stefanos Zafeiriou
\thanks{
Yannis Panagakis is with the Department of Informatics and Telecommunications, National and Kapodistrian University of Athens, Greece.}
\thanks{
Jean Kossaifi is with NVIDIA.}

\thanks{Grigorios G. Chrysos is
 with the Department of Electrical Engineering, Ecole Polytechnique Federale de Lausanne (EPFL), Switzerland.}

\thanks{James Oldfield is with Queen Mary University of London} 

\thanks{Mihalis   A.   Nicolaou is with the Computation-based Science and Technology Re-search Center at The Cyprus Institute.} 

\thanks{Anima Anandkumar is with  Caltech and NVIDIA.} 

\thanks{Stefanos Zafeiriou is with the Department of Computing, Imperial College London.} 
 
\thanks{$^{*}$Corresponding authors: Yannis Panagakis, email: yannisp@di.uoa.gr, Jean Kossaifi: email: jean.kossaifi@gmail.com }
\thanks{}}

\markboth{Proceedings of the IEEE}%
{}

\maketitle

\begin{abstract}
Tensors, or multidimensional arrays, are data structures that can naturally represent visual data of multiple dimensions. Inherently able to efficiently capture structured, latent semantic spaces and high-order interactions, tensors have a long history of applications in a wide span of computer vision problems. With the advent of the deep learning paradigm shift in computer vision, tensors have become even more fundamental.  Indeed, essential ingredients in modern deep learning architectures, such as convolutions and attention mechanisms, can readily be considered as tensor mappings.  In effect, tensor methods are increasingly finding significant applications in deep learning, including the design of memory and compute efficient network architectures, improving robustness to random noise and adversarial attacks, and aiding the theoretical understanding of deep networks.

This article provides an in-depth and practical review of tensors and tensor methods in the context of representation learning and deep learning, with a particular focus on visual data analysis and computer vision applications. Concretely, besides fundamental work in tensor-based visual data analysis methods, we focus on recent developments that have brought on a gradual increase of tensor methods, especially in deep learning architectures, and their implications in computer vision applications. To further enable the newcomer to grasp such concepts quickly, we provide companion Python notebooks, covering key aspects of the paper and implementing them, step-by-step with TensorLy.
\end{abstract}


\IEEEpeerreviewmaketitle


\section{Introduction}
\label{sec:introduction}
\looseness-1
Tensors are a multidimensional generalization of matrices, and a core mathematical object in multilinear algebra, just like vectors and matrices are in linear algebra. The {\it order} of a tensor is the number of indices needed to address its elements. A matrix is a second-order tensor of two {\it modes}, namely rows and columns, requiring two indices to access its elements. In analogy, an $N\myth$ order tensor has $N$ number of modes, and $N$ indices index it. 

Tensors have found tremendous applications in a wide range of disciplines in science and engineering; from 
many-body quantum systems \cite{montangero2018introduction} to algebraic geometry and theoretical computer science \cite{landsberg2012tensors},  to mention a few. In data science and related fields such as machine learning, signal processing, computer vision, and statistics, tensors have been used to represent and analyze information hidden in multi-dimensional data, such as images and videos, or to capture and exploit higher-order similarities or dependencies among vector-valued variables. Just as pair-wise (i.e., second-order) similarities of vector samples are represented by covariance or correlation matrices, a third- (or higher)-order tensor can represent similarities among three (or more) samples in a set. Such tensors are often referred to as higher-order statistics. In this context, tensor methods mainly focus on extending matrix-based learning models such as component analysis \cite{lu2003book}, dictionary learning e.g., \cite{bahri2019robustKC}, and regression models e.g., \cite{guo2012tensor}, to handle data represented by higher-order tensors or estimating parameters of latent variable models by analyzing higher-order statistics \cite{anandkumar2014tensor}.  Such developments were for a long time independent of the rise of deep learning, where the concept of a tensor is central. Concretely, besides representing data and statistics, tensors can be viewed as multilinear mappings (i.e., functions), which are higher-order generalizations of linear mappings represented by matrices.  For instance, multichannel convolutional kernels, which are the fundamental building blocks of deep convolutional  networks, can be readily described using the language of tensors. 

This article provides an overview of tensors and tensor methods in the context of representation learning and deep learning, with a particular focus on visual data analysis and computer vision applications. 

\looseness-1\textbf{Tensor structures in visual data.}
\looseness-1 Visual data is a prominent example of multi-dimensional data where a tensor structure is inherent in two main forms: 
\begin{itemize}
    \item {\it Tensor structure of visual measurements:}
    Visual data consist of physically meaningful modes that describe spatial, color, and temporal aspects of the data. Indeed, visual sensors readily generate tensor data: A grayscale image is represented by a matrix whose elements capture the intensity of light along with the spatial coordinates of the image. Hence, a batch of grayscale images is conveniently represented by a third-order tensor with two spatial modes and one mode indexing different images in the batch. A third-order tensor is also a natural way to represent a color image where color channels (e.g., RGB) are stacked along a third mode. Similarly, in tensor representations of hyperspectral and multispectral images, different channels capturing different spectral bands of light are stacked along with the third mode. Tensors of fourth-order are also being used to represent visual data that includes a time mode, in addition to the spatial and spectral/color modes. Higher-order tensor structure also arises in medical imaging. For instance,  magnetic resonance imaging (MRI) and functional magnetic resonance imaging (fMRI), are naturally stored in a multidimensional array such as a matrix (e.g., a two-dimensional (2D) MRI image), a third-order tensor (e.g., a three-dimensional (3D) MRI image), or a fourth-order tensor (e.g., a 3D fMRI image over time).
    \item {\it Latent tensor structure in image formation:} Image formation relies upon the interactions of multiple latent factors of variation related to appearance (e.g., illumination, pose) or even image semantics (such as gender and age in human faces). Such factors are often entangled through a multilinear (tensor) mapping in an unobserved, latent space that gives rise to the rich visual data structure. This tensor mapping assumes that visual variation is linear if we keep all but one factor constant.  For example, in facial images, changing the illumination does not change the depicted person's identity.
\end{itemize}

Typical computer vision and visual data analysis applications, such as detecting objects and recognizing faces in images and videos, understanding human behavior in videos (e.g., action and activity recognition), enhancing the quality of images (i.e., image denoising and inpainting), and generating novel visual content (i.e., image synthesis), require the extraction of physically--or semantically--meaningful representations of visual data. Traditional machine learning models employed towards this end, such as component analysis, sparse coding, and regression functions, treat data points as vectors and datasets as matrices. Hence, multi-dimensional visual data samples need to be flattened into very high-dimensional vectors, where natural topological structures and dependencies among different modes (e.g., spatial and temporal) of measurements are lost. Besides structure loss, when high-dimensional vectors are employed in training matrix-based machine learning models, the number of data samples needed to estimate an arbitrary function (or parameters of a machine learning model) within a given level of accuracy grows exponentially with data dimensionality. This phenomenon is known as the curse of dimensionality \cite{bellman1961adaptive}. 

{\bf Tensor methods in representation learning for computer vision:}
Tensor methods have emerged as a powerful tool for learning representations from multidimensional data by mitigating the curse of dimensionality without discarding their structure, which carries useful information in the case of visual data. Indeed, being able to both recover latent factors in visual data and flexible enough to accommodate a large set of structural constraints and regularizations, tensor decomposition and tensor component analysis methods have had a transformative impact in a wide range of computer vision applications,  ranging from human sensing (face and body analysis) to medical and hyperspectral imaging. In this context, dimensionality reduction, clustering, and data compression rely on decompositions of tensors representing visual data \cite{lu2011survey}. Non-negative factorization of such data tensors results in representations that correspond to local parts of the visual object \cite{shashua2005non}. Decomposing visual data into shape and motion factors, or surface normal and reflectance factors, facilitates the recovery of structure from motion and provides solutions to photometric stereo, e.g., \cite{wang2018disentangling}. Tensor decompositions with sparse regularization  \cite{Hawe_2013_Separable,bahri2019robustKC} have proven to be very useful in inverse problems in imaging, such as denoising and inpainting.

\looseness-1Recently, deep learning models have led to qualitative breakthroughs on a wide variety of computer vision applications, among many other machine learning tasks \cite{LeCun_Bengio_Hinton_2015}.
Deep neural networks act as learnable approximations of high-dimensional, non-linear functions. For example, in image classification, one seeks to learn an unknown high-dimensional target function that maps an input image to the output label.   The key ingredient of their success is that deep learning models leverage statistical properties of data such as stationarity (e.g., shift-invariance)  and compositionality (e.g., the hierarchical structure of images) through local statistics, which are present in visual data~\cite{lecun1995convolutional,krizhevsky2012imagenet,he2016deep}. These properties are exploited by convolutional architectures~\cite{lecun1995convolutional,lecun1998gradient}, which are built of alternating multidimensional convolutional layers, point-wise nonlinear functions (e.g.,  ReLU), downsampling (pooling) layers, while also containing
tensor-structured fully-connected layers.  The use of multidimensional convolutions in deep neural networks allows extracting local features shared across the image domain.  In turn, this greatly reduces the number of learnable parameters and consequently alleviates the effect of the curse of dimensionality--without sacrificing the ability to approximate the target function.

{\bf Tensor methods in deep learning architectures for computer vision:}
Despite the compositional structure of deep neural networks that mitigates the curse of dimensionality, deep learning models are usually over-parameterized involving an enormous number (typically tens of millions or even billions) of unknown parameters. 
Even though over-parametrization of deep neural networks allows finding good local minima, particularly when they are trained with gradient descent~\cite{du2018power,soltanolkotabi2018theoretical}, such a large number of trainable parameters makes deep networks prone to overfitting and noise ~\cite{caruana2001overfitting}, hindering the analysis of their generalization properties (i.e., how well they perform on unseen data) \cite{mhaskar2016learning}.
 Tensor decompositions can significantly reduce the number of unknown parameters of deep models and further mitigate the curse of dimensionality. Indeed, tensor decompositions can be applied to the weights of neural network layers to compress them, and in some cases, speed them up~\cite{novikov2015tensorizing,lebedev2015speeding,astrid2017cp}. In addition, using tensor-based operations within deep neural networks allows to preserve and leverage the topological structure in the data while resulting in parsimonious models~\cite{tcl,trl}. In convolutional neural networks (CNNs), (non-separable) multichannel convolution kernels can be decomposed into a sum of mode-separable convolutions using low-rank tensor decompositions, resulting in network compression and a reduction in the number of parameters~\cite{kossaifi2019tnet,kossaifi2020factorized}.
Deep network compression is significant in the deployment of deep learning models 
in resource-limited devices, while reducing the number of parameters acts as implicit regularization that improves generalization across tasks and domains~\cite{yang2017deep,chen2017sharing,bulat2019incremental}. Furthermore, randomized tensor decomposition of neural networks' building blocks has been proved effective in robustifying neural networks against adversarial attacks~\cite{goodfellow2014generative,madry2018towards} and to various types of random noise--such as noise occurring naturally when capturing data such as MRI~\cite{kolbeinsson2020stochastically}.

Surveys that approach tensors from varying perspectives have emerged over the past years.  In \cite{kolda2009tensor}, one of the most prominent and influential overviews of tensor decompositions is presented, while in \cite{comon2014tensors}, a mathematical account for multilinear mappings is provided. While \cite{kolda2009tensor,comon2014tensors} target audiences mostly from an applied mathematics perspective, other surveys focus on applications of tensors in scientific and engineering fields;  indicatively, in areas such as signal processing and machine learning~\cite{sidiropoulos2017tensor, 7038247},  data mining and fusion~\cite{papalexakis2016tensors}, multidimensional data analysis~\cite{leardi2005multi, acar2008unsupervised}, blind source separation~\cite{cichocki2009nonnegative}, computer vision \cite{lu2003book,lu2011survey}, and spectral learning \cite{anandkumar2014tensor,janzamin2019spectral}.

While the surveys mentioned above cover an impressive spectrum of theoretical and algorithmic developments and applications, one can identify a gap in the literature concerning recent advancements in the use of tensor methods in the context of representation learning and deep learning for computer vision.  In this paper,  besides fundamental work in tensor learning for computer vision, we focus on recent developments that have brought on a gradual increase of tensor-related methods, especially in deep learning architectures, and their implications in vision applications.  To further enable the newcomer to grasp such concepts quickly, we provide thorough tutorial-style companion notebooks in Python, covering every section of the paper\footnote{The GitHub repository containing the notebooks can be found here: \url{https://github.com/tensorly/Proceedings_IEEE_companion_notebooks}. Specific notebooks are linked in a gray box under the relevant section title.}.

{\bf Structure of the paper.}
The remainder of the paper is organized as follows. In~\cref{sec:preliminaries}, tensor and matrix algebra fundamentals are presented.  In \cref{sec:comptools}, an overview of existing computational tools and the necessary infrastructure is discussed, focusing on the TensorLy library on which the companion notebooks of this paper are based. \cref{sec:replearning} provides an introduction to representation learning with tensors. In~\cref{sec:deeparch}, recent research that utilizes tensor methods in deep learning architectures are detailed, while related applications in computer vision are discussed in~\cref{sec:applCV}.
Finally, a summary of practical challenges is presented in~\cref{sec:challenges}, along with advice to practitioners.


\section{Preliminaries on Matrices and Tensors} 
\label{sec:preliminaries}
\notebook{tensor\_manipulation}

Multilinear algebra is much richer than its linear counterpart and hence tensor algebra notation and manipulation of tensors may appear complicated to newcomers in the field. To make the paper self-contained and give the reader an understanding of the mathematical tools employed in tensor methods for learning, we first introduce the notation conventions used in this paper and then review some of the fundamental concepts of linear and mutlilinear algebra. 
 
To ensure consistency in terminology that would be familiar to machine learning and computer vision researchers we have adopted the terminology  employed in relevant overview papers \cite{sidiropoulos2017tensor,kolda2009tensor,anandkumar2014tensor} and recent publications on these topics. The notation used here is very similar to that adopted in \cite{sidiropoulos2017tensor,kolda2009tensor,kiers2000towards}.
However, we differentiate on one point: we employ a different ordering of elements when tensors are flattened to matrices than that used in \cite{kolda2009tensor} and may be familiar to some of the readers (cf. Definition \ref{D:Tensor_unfolding}). 

\begin{figure*}
    \centering
    \begin{subfigure}[t]{0.2\textwidth}
    \includegraphics[width=1\linewidth]{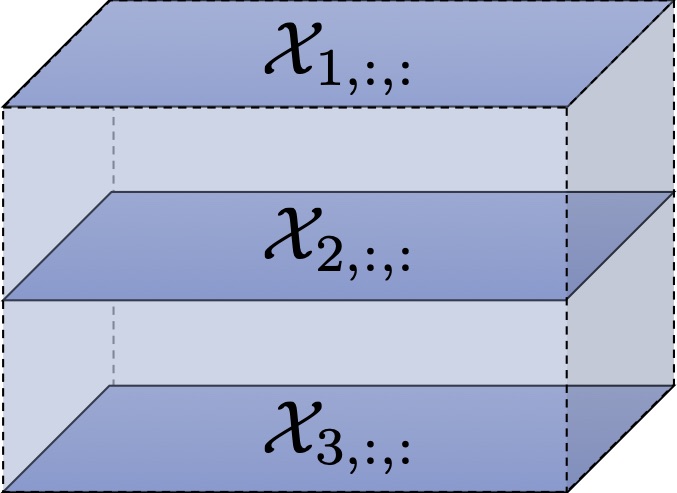}
    \caption{\textbf{Horizontal Slices}}
    \end{subfigure}
    \qquad
    \begin{subfigure}[t]{0.2\textwidth}
    \includegraphics[width=1\linewidth]{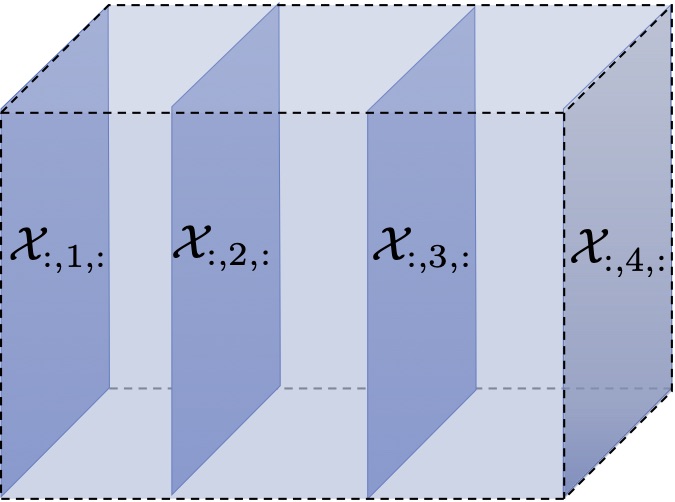}
    \caption{\textbf{Lateral Slices}}
    \end{subfigure}
    \qquad
    \begin{subfigure}[t]{0.2\textwidth}
    \includegraphics[width=1\linewidth]{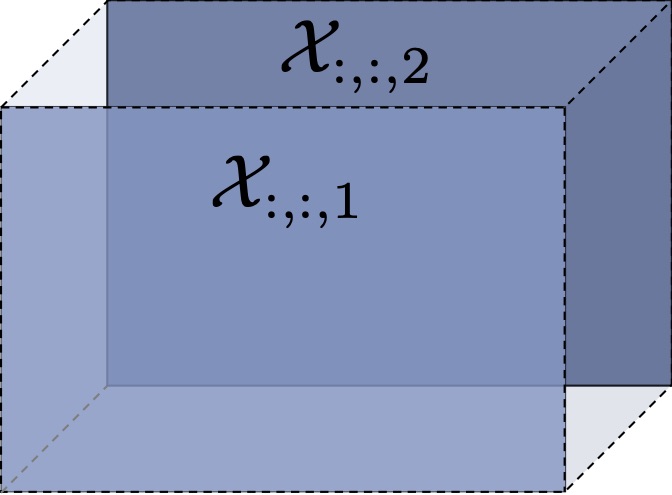}
    \caption{\textbf{Frontal Slices}}
    \end{subfigure}
    \caption{\textbf{Illustration of the slices of a third order tensor of size $3 \times 4 \times 2$}}
    \label{fig:tensor-slices}
\end{figure*}

\subsection{Notation}
{\it Scalars} are denoted by plain letters i.e., $x, I,  J$. First order tensors are {\it vectors}, denoted as \(\myvector{x}\). Second order tensors are {\it matrices}, denoted as  \(\mymatrix{X}\). 
\(\mymatrix{I}\) denotes the identity matrix of compatible dimensions. The transpose of \(\mymatrix{X}\) is denoted \(\mymatrix{X}\myT\). The $i$th column of \(\mymatrix{X}\) is denoted as \(\myvector{x}_i\). 

The sets of real and integer numbers are denoted by $\mathbb{R}$ and $\mathbb{Z}$, respectively. A set of $M$ real matrices (vectors) of varying dimensions is denoted by $\{\mathbf{X}^{(m)} \in \mathbb{R}^{I_m \times N} \}_{m=1}^{M}$ $( \{ \mathbf {x}^{(m)} \in \mathbb{R}^{I_m} \}_{m=1}^{M} )$. 
Finally, for any $i, j \in \mathbb{Z}$, with \(i < j\), \(\myrange{i}{j}\) denotes the set of integers \(\{ i, i+1, \cdots , j-1, j\}\).

 {\it Tensors} of order three or higher are denoted by  \(\mytensor{X}\). The reader is reminded that the order of a tensor is the number of indices (dimensions) needed to address its elements. Each dimension is called a mode.  Specifically, an $N\myth$ order tensor has $N$ indices, with each index addressing a mode of $\mytensor{X}$. Assuming  \(\mytensor{X}\) is real-valued, it is defined over the
tensor space $\mathbb{R}^{I_{1} \times I_{2} \times \cdots \times
I_{N}}$, where $I_{n} \in \mathbb{Z}$ for $n=1,2,\ldots,N$.
An element $(i_1, i_2, \cdots, i_N)$ of tensor $\mytensor{X} \in \myR^{I_1 \times I_2   \times \cdots \times I_{N}}$ is accessed as:
$\mytensor{X}_{i_1, i_2, \cdots, i_{N}}$
or
$\mytensor{X}( i_1, i_2, \cdots, i_{N})$.
This corresponds to viewing a tensor as a multi-dimensional array in $ \myR^{I_1 \times I_2 \times \cdots \times I_{N}}$. Furthermore, given a set of $N$ vectors (or matrices) that correspond to each mode of \(\mytensor{X}\), the $n\myth$ vector (or matrix) is denoted as \(\myvector{u}^{(n)}\) (or \(\mymatrix{U}^{(n)}\)).

{\it Fibers} are a generalization of the concept of rows and columns of matrices to tensors. They are obtained by fixing all indices but one. A colon is used to denote all elements of a mode e.g., the mode-1 fibers of \(\mytensor{X}\) are denoted as \(\mytensor{X}_{:, i_2, i_3}\).
 
{\it Slices} (Figure \ref{fig:tensor-slices}) are obtained, for a third-order tensor, by fixing one of the indices.

\newcolumntype{Y}{>{\centering\arraybackslash}X}
\renewcommand{\arraystretch}{1.5}
\begin{table*}[ht]
    \centering
    \begin{tabularx}{\linewidth}{c:Y:Y:Y}
        \parbox[t]{5mm}{
            \multirow{3}{*}{\rotatebox[origin=c]{90}{\textbf{Fibers}}}
        }
        & \textbf{mode-0 fibers} & \textbf{mode-1 fibers} & \textbf{mode-2 fibers}
         \\
         & \includegraphics[valign=M,width=0.45\linewidth]{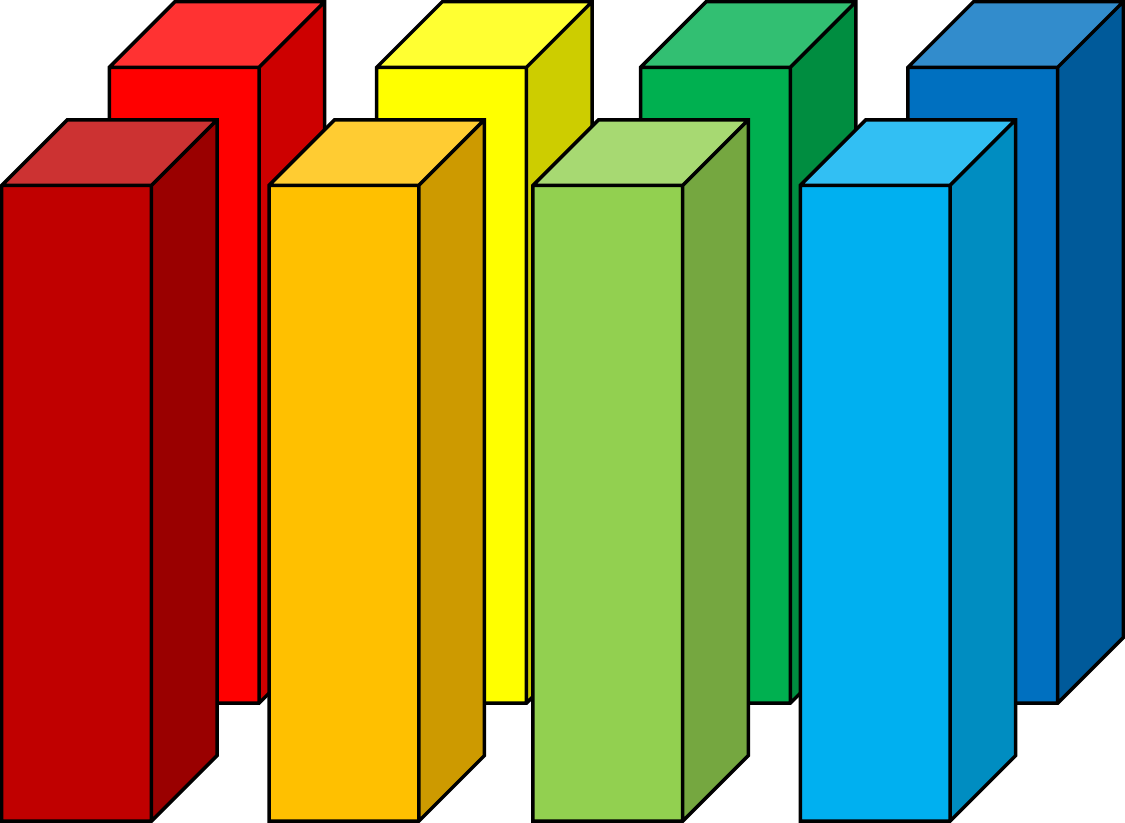}
         \quad &
         \includegraphics[valign=M,width=0.45\linewidth]{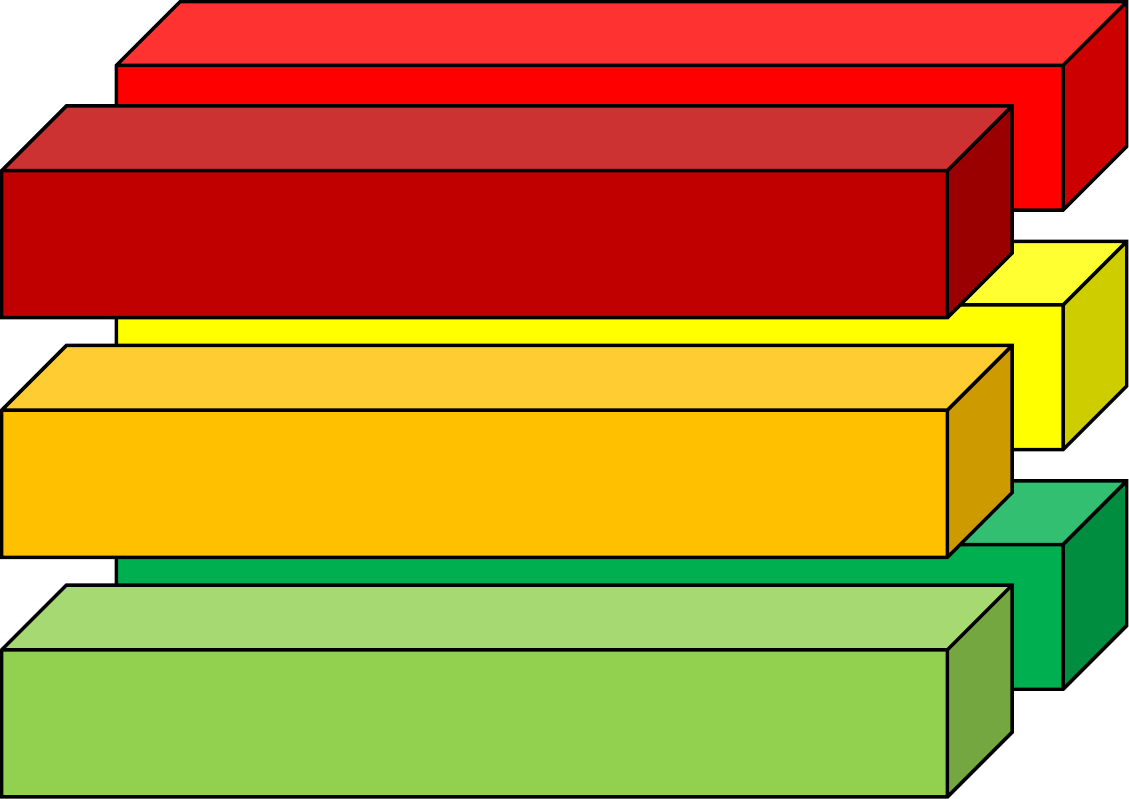}
         \quad &
         \includegraphics[valign=M,width=0.45\linewidth]{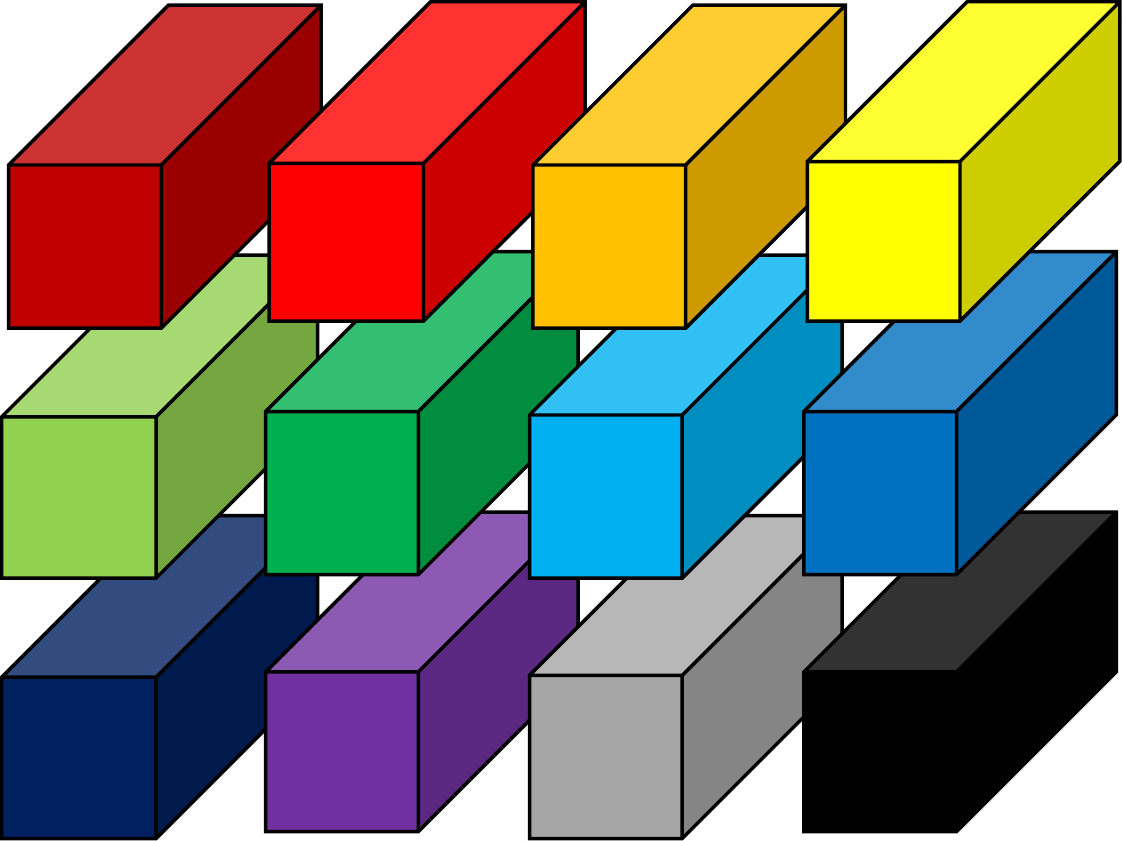}
         \\[1cm]
        \cdashline{2-4}
        \parbox[t]{5mm}{
            \multirow{3}{*}{\rotatebox[origin=c]{90}{\textbf{TensorLy}}}
        }
         &\textbf{mode-0 unfolding} & \textbf{mode-1 unfolding} & \textbf{mode-2 unfolding} 
         \\
         &\includegraphics[valign=M,width=0.45\linewidth]{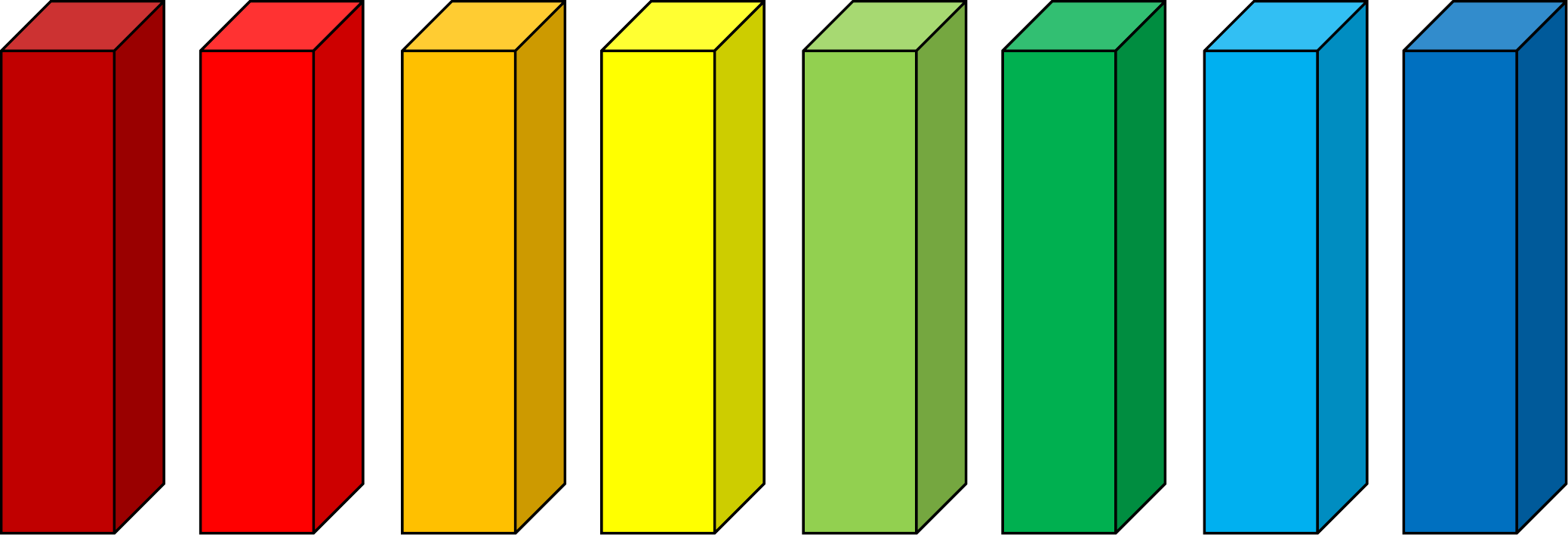}
         \quad &
         \includegraphics[valign=M,width=0.4\linewidth]{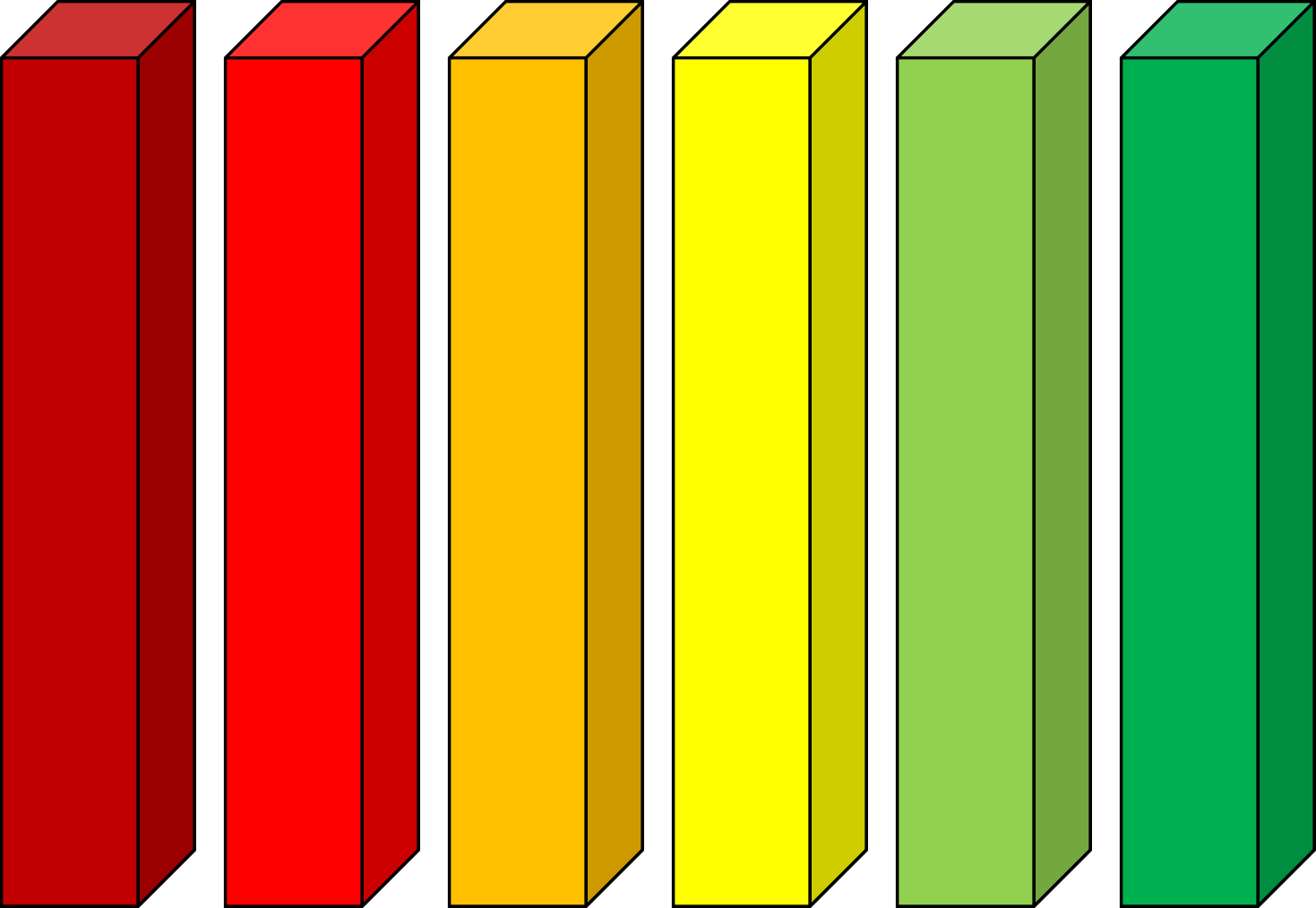}
         \quad &
         \includegraphics[valign=M,width=0.55\linewidth]{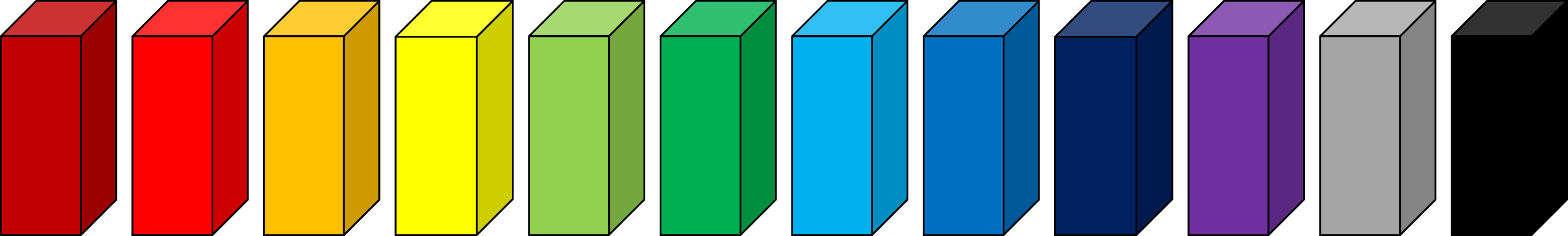}
         \\[1cm]
         \cdashline{2-4}
         \parbox[t]{5mm}{
            \multirow{3}{*}{\rotatebox[origin=c]{90}{\textbf{Matlab}}}
        }
         &\textbf{mode-0 unfolding} & \textbf{mode-1 unfolding} & \textbf{mode-2 unfolding} 
         \\
         &\includegraphics[valign=M,width=0.45\linewidth]{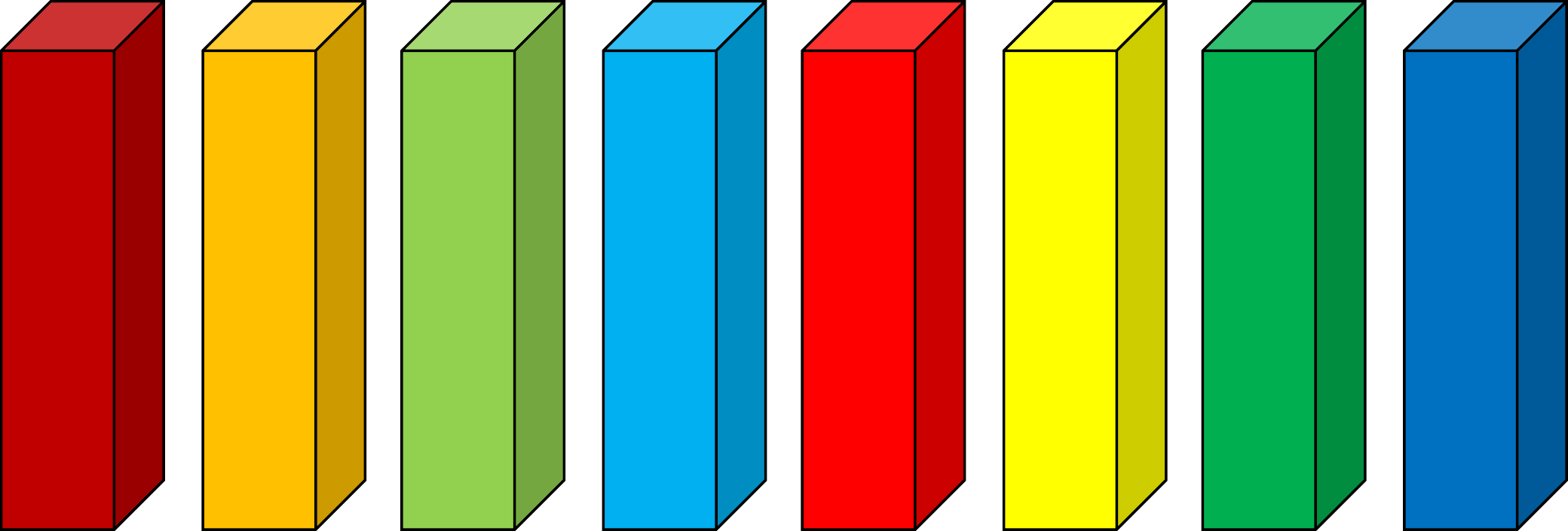}
         \quad &
         \includegraphics[valign=M,width=0.4\linewidth]{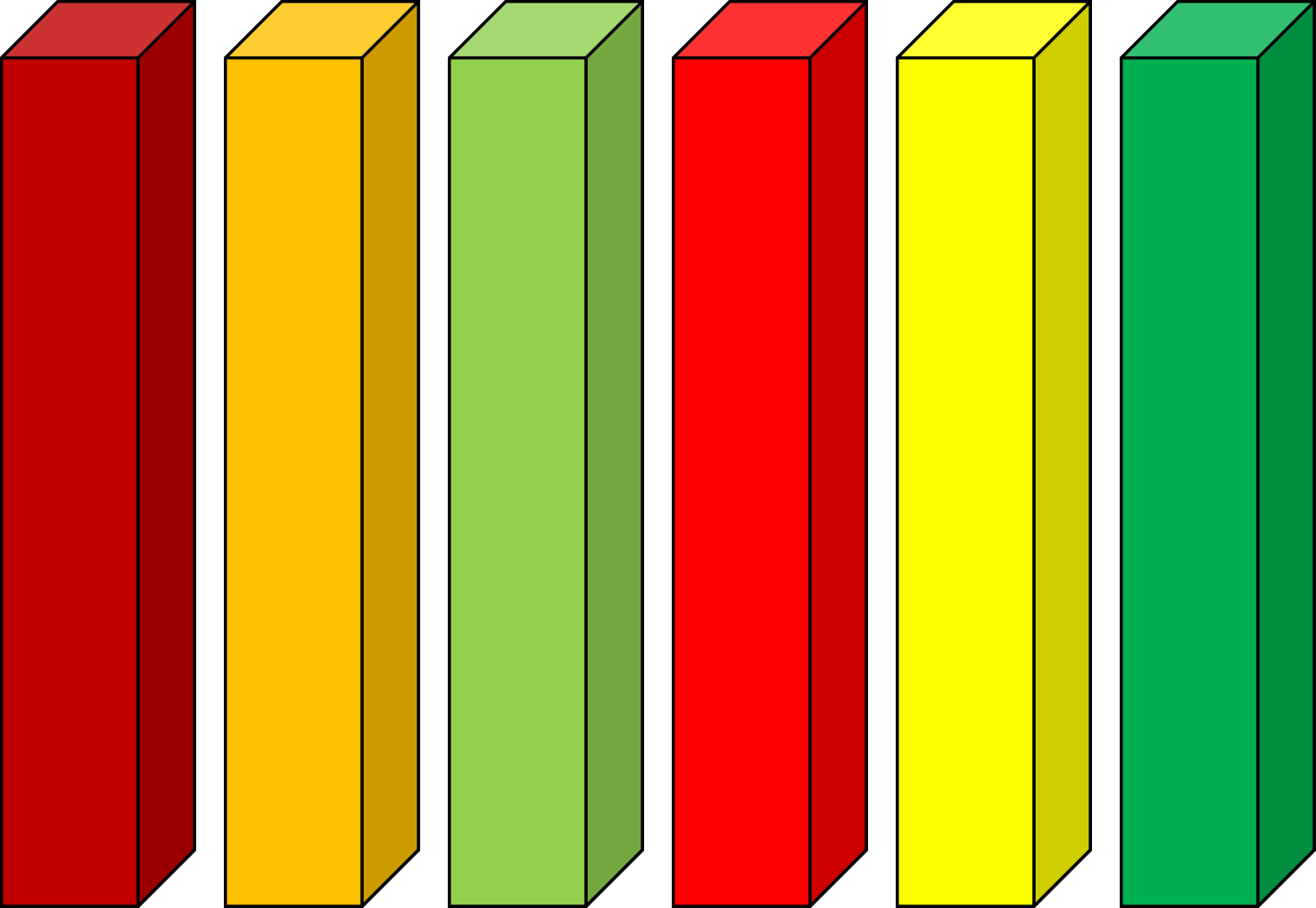}  
         \quad &
         \includegraphics[valign=M,width=0.55\linewidth]{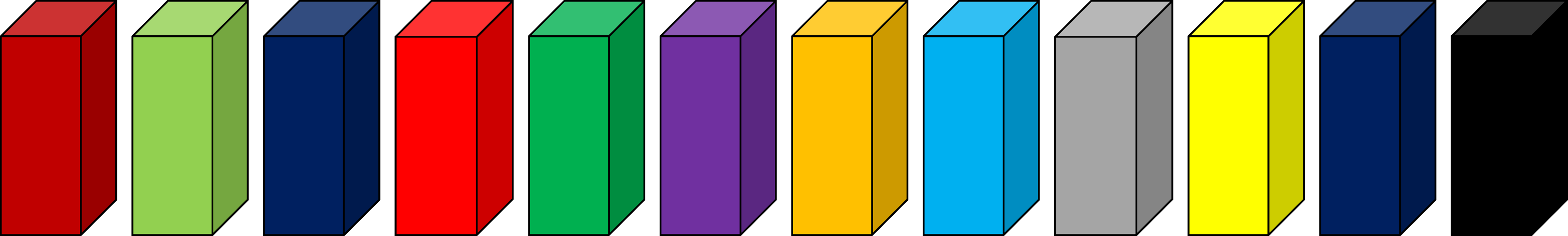}
    \end{tabularx}
    \caption{\textbf{Illustration of the fibers and unfolding of a third order tensor.}}
    \label{table:tensor-fibers}
\end{table*}

\subsection{Transforming tensors into matrices and vectors}

\begin{definition}[Tensor unfolding]\label{D:Tensor_unfolding}
	Given a tensor,
	\( \mytensor{X} \in \myR^{I_1 \times I_2 \times \cdots \times I_{N}}\),
	its mode-\(n\) unfolding is a matrix \(\mymatrix{X}_{[n]} \in \myR^{I_n \times I_M}\), 
	with \(M = \prod_{\substack{k=1,\\k \neq n}}^{N} I_k\)
	and is defined by the mapping from element
	\( (i_1, i_2, \cdots, i_{N})\) to \((i_n, j)\), with 
	\(
	j = 1 + \sum_{\substack{k=1,\\k \neq n}}^{N} (i_k - 1) \times \prod_{\substack{m=k+1,\\ m \neq n}}^{N} I_m 
	\).
\end{definition}
Tensor unfolding corresponds to a reordering of the fibers of the tensor as the columns of a matrix. There are more than one definition of tensor unfolding, which vary by the ordering of the fibers. In particular, our definition corresponds to a row-wise underlying ordering (or C-ordering ) of the elements. The main other definition, popularised by Kolda and Bader \cite{kolda2009tensor}, corresponds to a column-wise (or Fortran) ordering of the elements, which is more efficient for libraries using that ordering, such as MATLAB. Conversely, our definition naturally leads to better performance for implementation with C-ordering of the elements, which is the case with most Python/C++ software libraries as well as most GPU libraries~\cite{tensorly}. Table~\ref{table:tensor-fibers} illustrates the link between the fibers and the unfoldings, as well as the difference between our definition (row-wise, or C-ordering) and the column-wise, or Fortran ordering \cite{kolda2009tensor}.

\begin{definition}[Tensor vectorization]
	Given a tensor,
	\( \mytensor{X} \in \myR^{I_1 \times I_2 \times \cdots \times I_{N}}\), 
	we can flatten it into a vector \(\text{vec}(\mytensor{X})\) 
	of size \(I_1 \cdot I_2 \cdot \ldots \cdot I_{N}\) 
	defined by the mapping from element
	\( (i_1, i_2, \cdots, i_{N})\) of \(\mytensor{X}\) to element \(j\) of \(\text{vec}(\mytensor{X})\), with 
	\( j = 1 + \sum_{k=1}^{N} (i_k - 1) \times \prod_{m=k+1}^{N} I_m \).
\end{definition}
\subsection{Matrix and tensor products}
Here, we will define several matrix and tensor products employed by the surveyed methods.
\begin{definition}[Kronecker product]
Given two matrices \(\mymatrix{A} \in \myR^{I_1 \times I_2}\) and \(\mymatrix{B} \in \myR^{J_1 \times J_2}\), their Kronecker product is defined as:
\[
\mymatrix{A}\otimes\mymatrix{B} = \begin{bmatrix} a_{11} \mymatrix{B} & \cdots & a_{1I_2}\mymatrix{B} \\ \vdots & \ddots & \vdots \\ a_{I_11} \mymatrix{B} & \cdots & a_{I_1 I_2} \mymatrix{B} \end{bmatrix}
\in \myR^{I_1 \cdot J_1 \times I_2 \cdot J_2}
\]
\end{definition}

\begin{definition}[Khatri-Rao product]
Given two matrices \(\mymatrix{A} \in \myR^{I \times R}\) and \(\mymatrix{B} \in \myR^{J \times R}\) with the same number of columns, their Khatri-Rao product, also known as \emph{column-wise} Kronecker product, is defined as \(\mymatrix{A} \odot \mymatrix{B} \in \myR^{I\cdot J \times R}\):
\[
\mymatrix{A} \odot \mymatrix{B} = 
\left[
\begin{matrix}
  \mymatrix{A}_{:, 1} \otimes \mymatrix{B}_{:, 1},&
  \mymatrix{A}_{:, 2} \otimes \mymatrix{B}_{:, 2},&
  \cdots, &
  \mymatrix{A}_{:, R} \otimes \mymatrix{B}_{:, R}
\end{matrix}
\right]
\]
\end{definition}
Furthermore, the Khatri-Rao product of a set
of $M$ matrices  $\{\mathbf{X}^{(m)} \in \mathbb{R}^{I_m \times N} \}_{m=1}^{M}$ 
is denoted by $\mathbf{X}^{(1)} \odot \mathbf{X}^{(2)} \odot  \cdots \odot \mathbf{X}^{(M)}$.
 
 \begin{definition}[Hadamard product]
 The Hadamard product of $\mymatrix{A}, \mymatrix{B} \in \myR^{I \times J}$ is the element-wise product is symbolized as $\mymatrix{A} * \mymatrix{B}$ and is defined element-wise as $\mymatrix{A}_{(i, j)} \mymatrix{B}_{(i, j)}$. 
 \end{definition}

\begin{definition}[Outer product]
$\circ$ denotes  the vector outer product. Given a set of \(N \) vectors $\{ \mathbf {x}^{(n)} \in \mathbb{R}^{I_n} \}_{n=1}^{N}$ their outer product is denoted as \(
    \mytensor{X} = \myvector{x}^{(1)} \circ \myvector{x}^{(2)} \circ \cdots \circ \myvector{x}^{(N)} 
    \in \myR^{I_1 \times I_2 \times \cdots \times I_{N}}
\) 
and defines  a \textit{rank}-one $N\myth$-order tensor.
\end{definition}

\begin{definition}[$n$-mode product]
	For a tensor \(\mytensor{X} \in \myR^{I_1 \times I_2 \times \cdots \times I_{N}}\) and a matrix \( \mymatrix{M} \in \myR^{R \times I_n} \), the n-mode product of a tensor 
	is a tensor of size 
	\(\left(I_1 \times \cdots \times I_{n-1} \times R \times I_{n+1} \times \cdots \times I_{N}\right)\) 
	and can be expressed using unfolding of \(\mytensor{X}\) 
	and the classical dot product as:
	\begin{equation}\nonumber
		\mytensor{X} \times_n \mymatrix{M} = \mymatrix{M} \mymatrix{X}_{[n]} \in \myR^{I_1 \times \cdots \times I_{n-1} \times R \times I_{n+1} \times \cdots \times I_{N}}
	\end{equation}
	
The \textit{$n$- mode vector product} of a tensor  $\mytensor{X} \in
\mathbb{R}^{I_{1}\times I_{2}\times \ldots \times I_{N}}$ with a
vector $\mathbf{x} \in \mathbb{R}^{I_n}$ is denoted by
$\mytensor{X}  \times_{n} \mathbf{x} \in \mathbb{R}^{I_{1}\times
I_{2}\times\cdots\times I_{n-1}  \times I_{n+1} \times
\cdots \times I_{N}} $. The resulting tensor is of order $N-1$ and is defined element-wise as
\begin{equation}\label{E:Tensor_Mode_n}
(\mytensor{X} \times_{n} \mathbf{x})_{i_1, \ldots, i_{n-1}, i_{n+1},
\ldots, i_{N}} = \sum_{i_n=1}^{I_n} x_{i_1, i_2, \ldots, i_{N}} x_{i_n}.
\end{equation}
In order to simplify the notation, we denote
$\bm{\mathcal{X}} \times_{1} \mathbf{x}^{(1)} \times_{2} \mathbf{x}^{(2)} \times_{3}  \cdots \times_{N} \mathbf{x}^{(N)}  =
\bm{\mathcal{X}} \prod_{n=1}^N \times_{n} \mathbf{x}^{(n)}$.
\end{definition}

\begin{definition}[Generalized inner-product]
	For two tensors \(\mytensor{X}, \mytensor{Y} \in \myR^{I_1 \times I_2 \times \cdots \times I_N}\) of same size, their inner product is defined as 
	\(
	\myinner{\mytensor{X}}{\mytensor{Y}}=~\sum_{i_1=1}^{I_1}\sum_{i_2=1}^{I_2} \cdots \sum_{i_n=1}^{I_N} \mytensor{X}_{i_1, i_2, \cdots, i_n} \mytensor{Y}_{i_1, i_2, \cdots, i_n}
	\) 
	For two tensors \(\mytensor{X} \in \myR^{D_x \times I_1 \times I_2 \times \cdots \times I_N}\) and \( \mytensor{Y} \in \myR^{I_1 \times I_2 \times \cdots \times I_N \times D_y} \) sharing \(N\) modes of same size,
	we similarly define the generalized inner product along the \(N\) last (respectively first) modes of \(\mytensor{X}\) (respectively \(\mytensor{Y}\)) as 
	\begin{equation}\nonumber
	\myinner{\mytensor{X}}{\mytensor{Y}}_N =~\sum_{i_1=1}^{I_1}\sum_{i_2=1}^{I_1} \cdots \sum_{i_n=1}^{I_N} \mytensor{X}_{:, i_2, i_3, \cdots, i_n} \mytensor{Y}_{i_1, i_2, \cdots, i_{n-1}, :}
	\end{equation}
	with \( \myinner{\mytensor{X}}{\mytensor{Y}}_N \in \myR^{I_x \times I_y}\).
\end{definition}

\begin{definition}[Convolution]
We denote a regular convolution of \(\mytensor{X}\) with \(\mymatrix{W}\) as \(\mytensor{X} \myconv \mymatrix{W}\). For \(1\)--D convolutions, we write the convolution of a tensor \(\mytensor{X} \in \myR^{I_1\times I_2 \times \cdots \times I_N}\) with a vector \(\myvector{v} \in \myR^{K}\) along the \(n\myth\)--mode as \(\mytensor{X} \myconv_n \myvector{v}\). In practice, as done in current deep learning frameworks~\cite{pytorch},
we use cross-correlation, which differs from a convolution by a flip of the kernel. This does not impact the results since the weights are learned end-to-end. In other words, \(\left( \mytensor{X} \myconv_n \myvector{v}\right)_{i_1, \cdots, i_N}  = \sum_{k = 1}^K \myvector{v}_k \mytensor{X}_{i_1, \cdots, i_{n-1}, i_n + k, i_{n+1}, \cdots, I_N}\).
\end{definition}

\subsection{Tensor diagrams}
Most tensor methods involve a series of tensor operators, such as 
tensor contractions\footnote{The $n$-mode product of a tensor with a matrix is a tensor contraction between the $n$-th mode of the tensor and the second mode of the matrix.}, resulting in several sums over various modes and tensors, with as many indices, which can be cumbersome to read and write. 

\begin{figure}[h]
    \centering
    \begin{subfigure}[t]{0.2\textwidth}
        \centering
		\includegraphics[height=1.5cm]{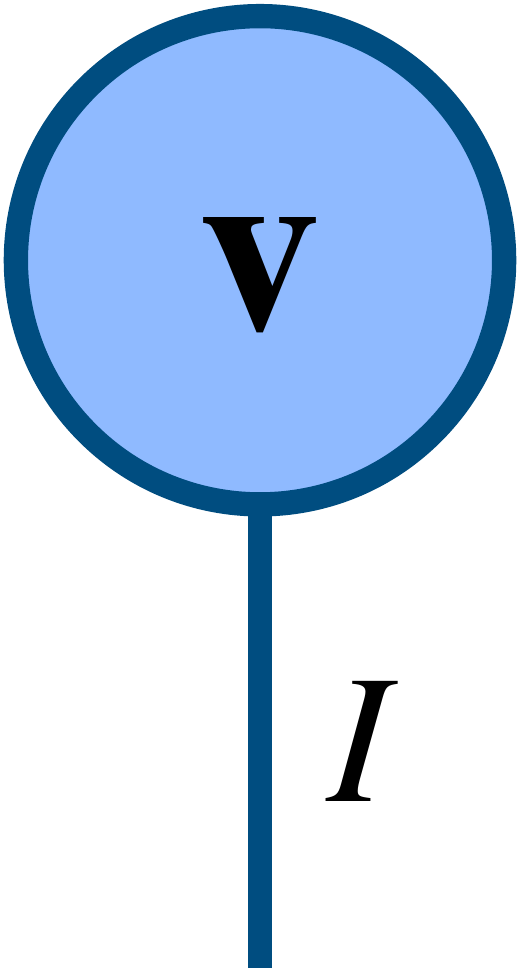}
		\caption{vector}\label{fig:diagram-vector}
	\end{subfigure}
	\qquad
	\begin{subfigure}[t]{0.24\textwidth}
	    \centering
		\includegraphics[height=2.3cm]{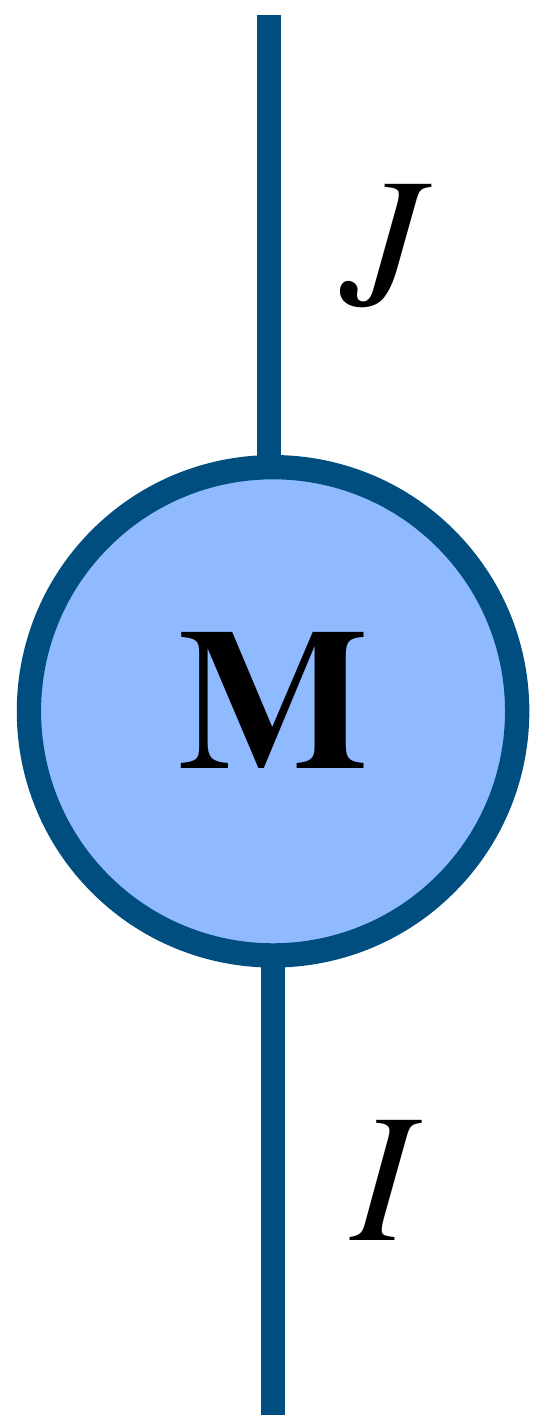}
		\caption{matrix}\label{fig:diagram-matrix}
	\end{subfigure}
	\begin{subfigure}[t]{0.24\textwidth}
	    \centering
		\includegraphics[height=2.5cm]{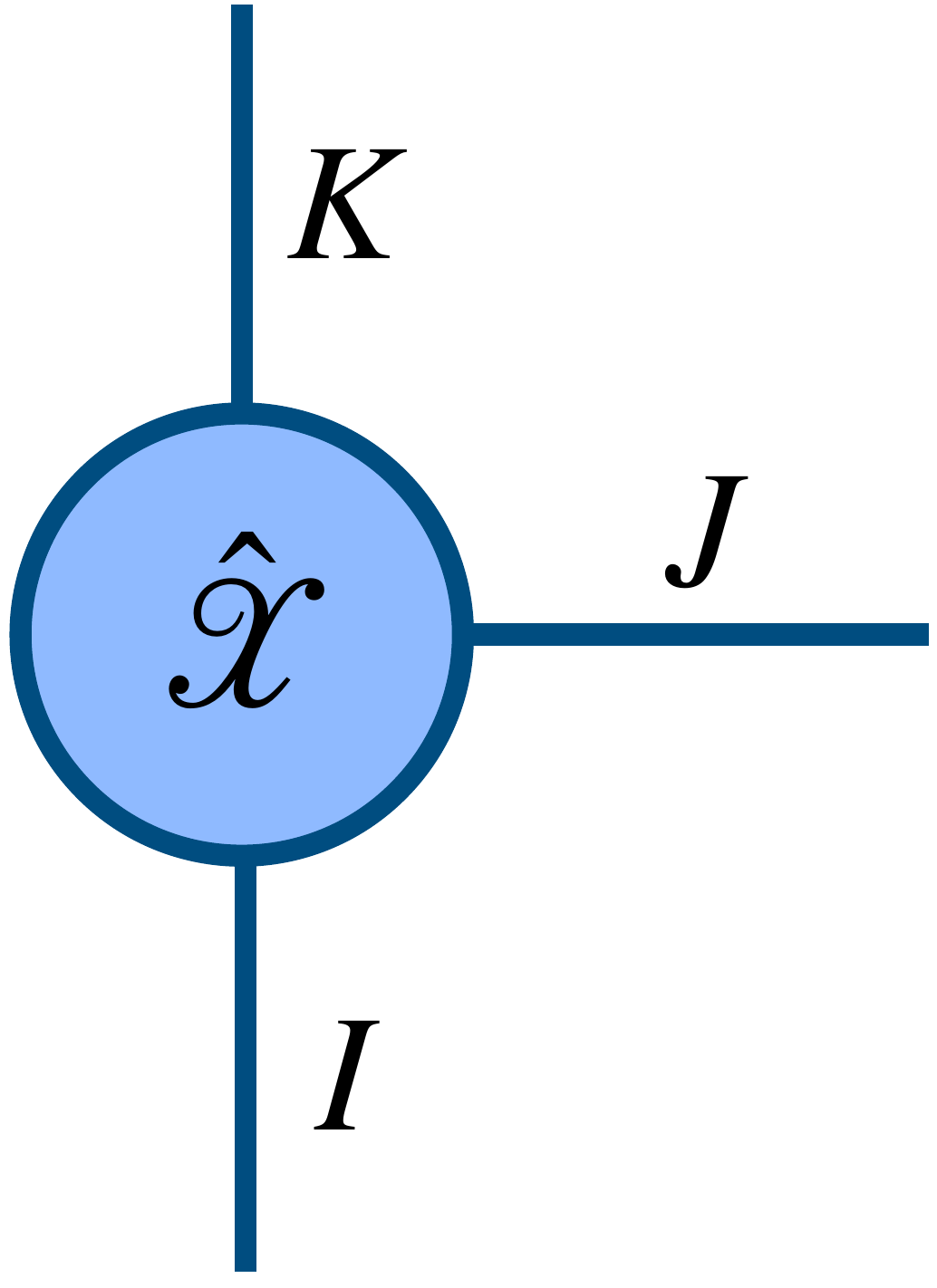}
		\caption{\(3^\text{rd}\) order tensor}\label{fig:diagram-tensor-3}
	\end{subfigure}
	\begin{subfigure}[t]{0.24\textwidth}
	    \centering
		\includegraphics[height=2.5cm]{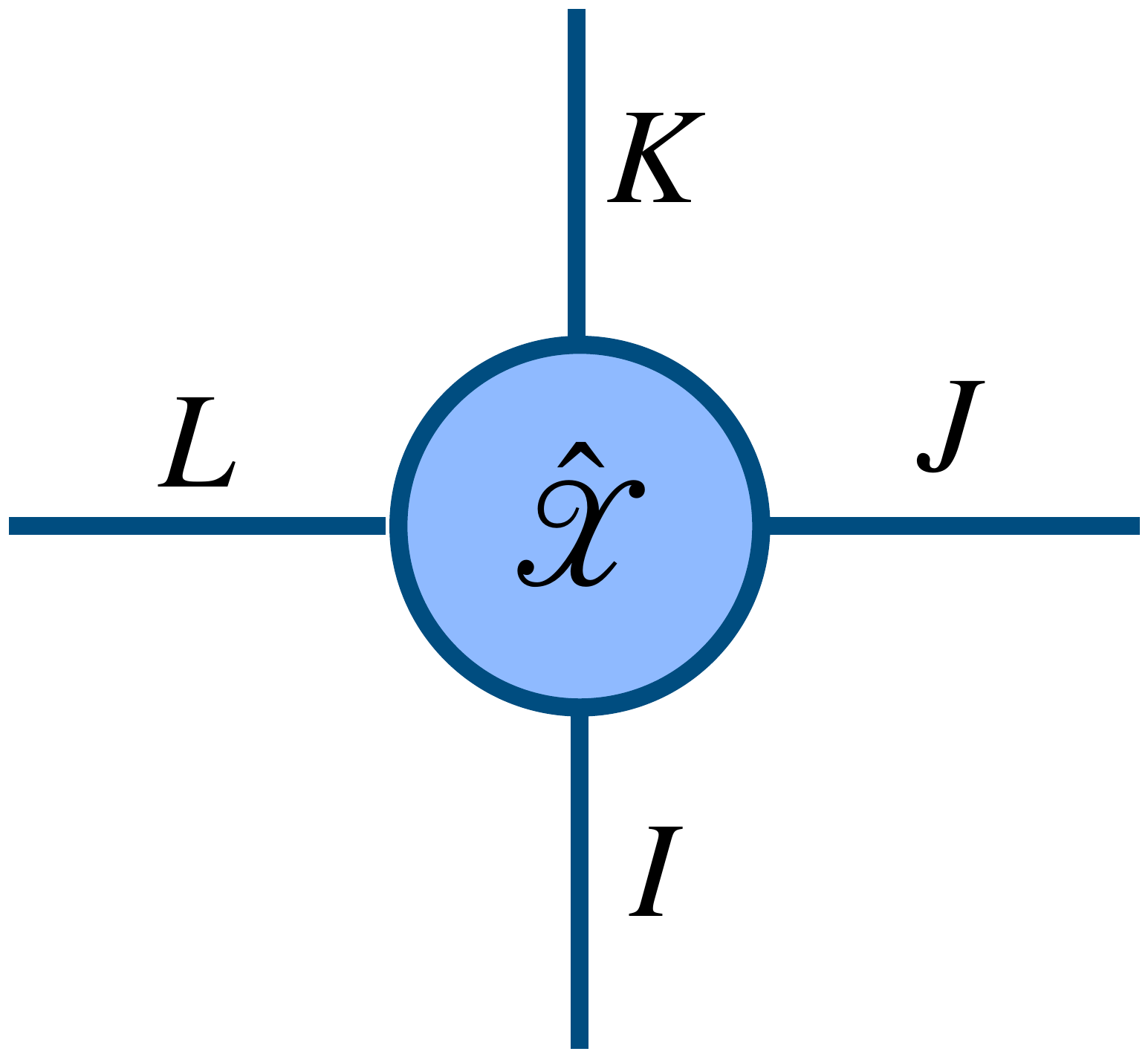}
		\caption{\(4^\text{th}\) order tensor}\label{fig:diagram-tensor-4}
	\end{subfigure}
	\caption{Representation of tensors of various orders using tensor diagrams. The vertices represent tensors and their degree the order of the tensors.}\label{fig:diagram-examples}
\end{figure}

{\it Tensor diagrams}  are pictorial representations of tensor algebraic operations, in the form of undirected graphs offering a more intuitive way to read and write tensor operators and design tensor methods.  The vertices (circles) represent tensors, and its edge the modes of that tensor. The degree of each vertex (the number of edges coming out of it) thus represents the order of the corresponding tensor. Consequently, using tensor diagrams a scalar~(Fig. \ref{fig:diagram-vector}) would simply be represented by a vertex while a vector (i.e., a tensor with one index) would have one edge. A matrix (Fig. \ref{fig:diagram-matrix}) would be a vertex with two edges, and so on for third-~(Fig. \ref{fig:diagram-tensor-3}) and fourth-order~(Fig. \ref{fig:diagram-tensor-4}) tensors.
 
\begin{figure}[h]
    \centering
    \begin{subfigure}[t]{0.23\textwidth}
        \centering
		\includegraphics[height=3cm]{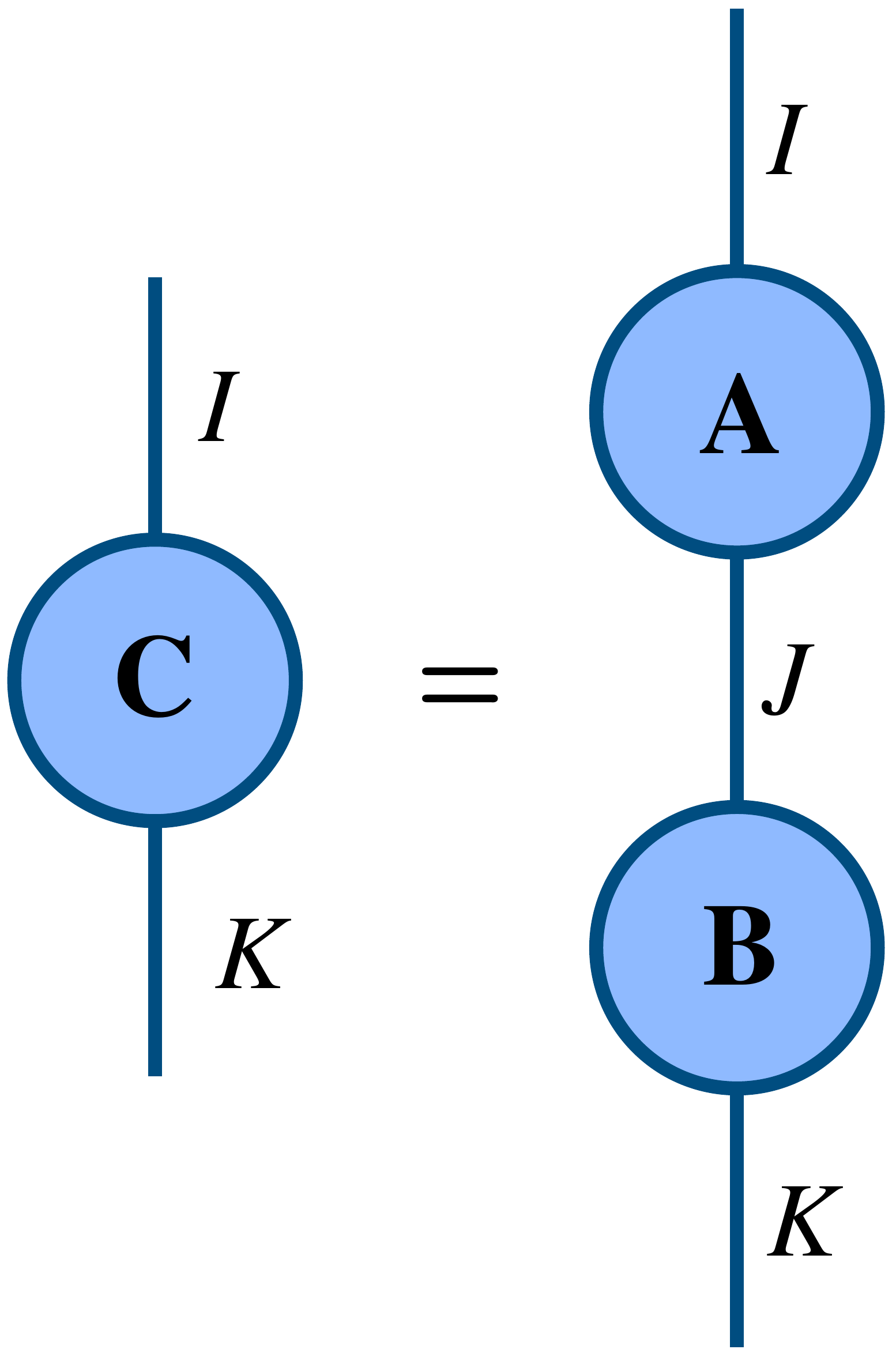}
		\caption{matrix product}\label{fig:diagram-matrix-product}
	\end{subfigure}
	\begin{subfigure}[t]{0.23\textwidth}
	    \centering
		\includegraphics[height=2.5cm]{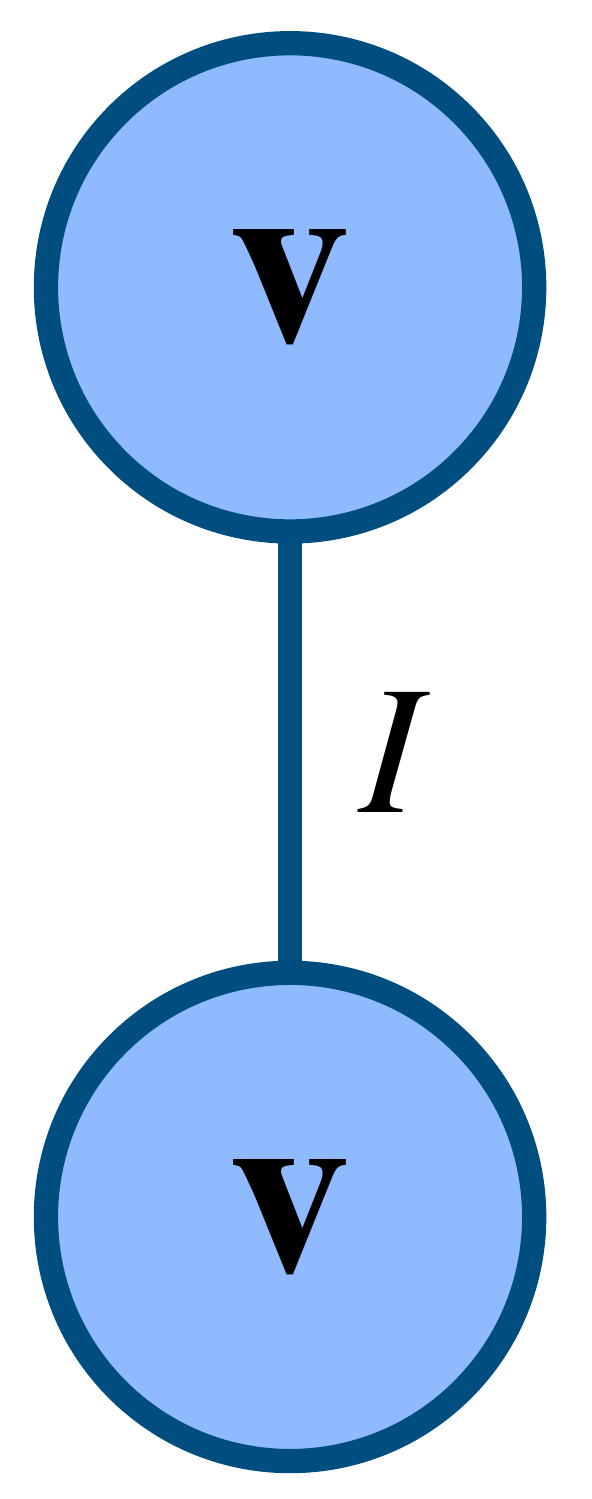}
		\caption{inner-product (vectors)}\label{fig:diagram-inner-vec}
	\end{subfigure}
	\begin{subfigure}[t]{0.23\textwidth}
	    \centering
		\includegraphics[height=2cm]{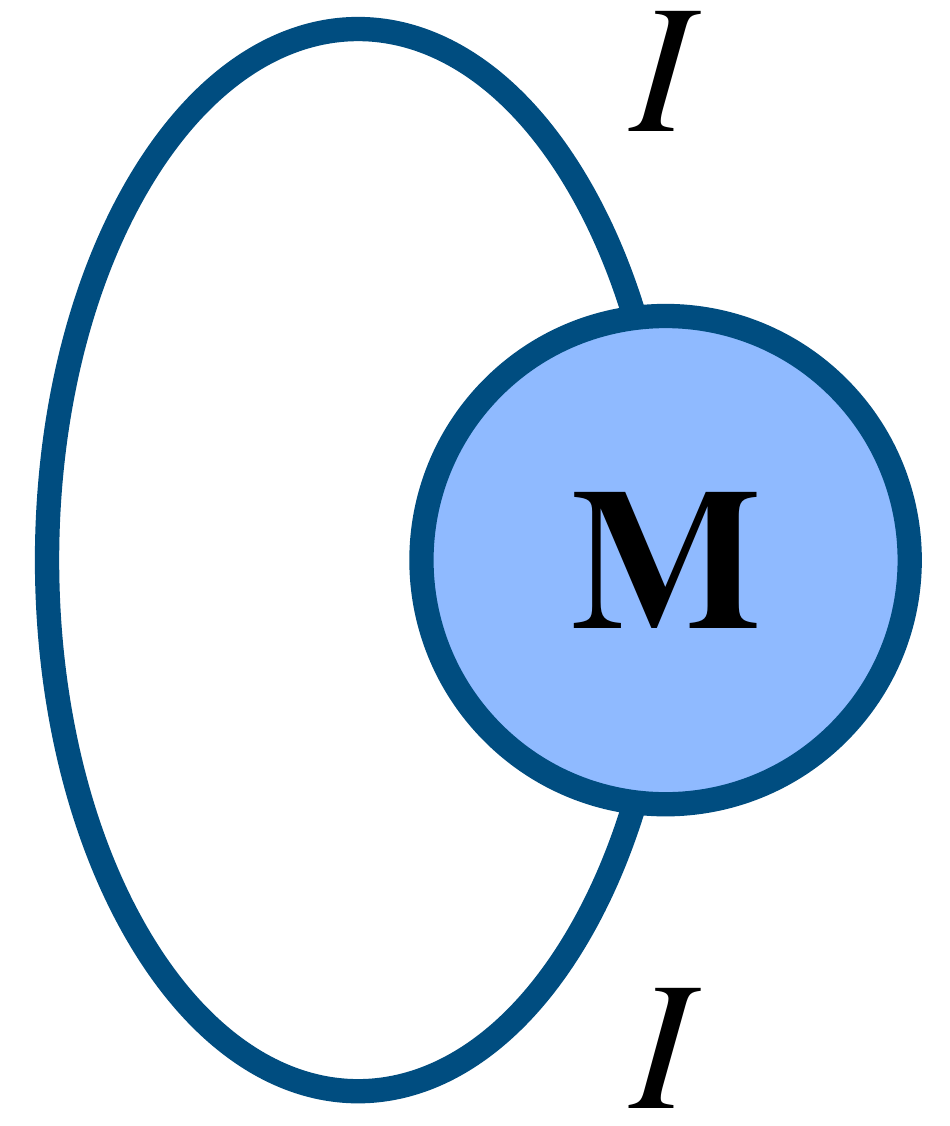}
		\caption{matrix trace}\label{fig:diagram-trace}
	\end{subfigure}
	\begin{subfigure}[t]{0.23\textwidth}
	    \centering
		\includegraphics[height=1.5cm]{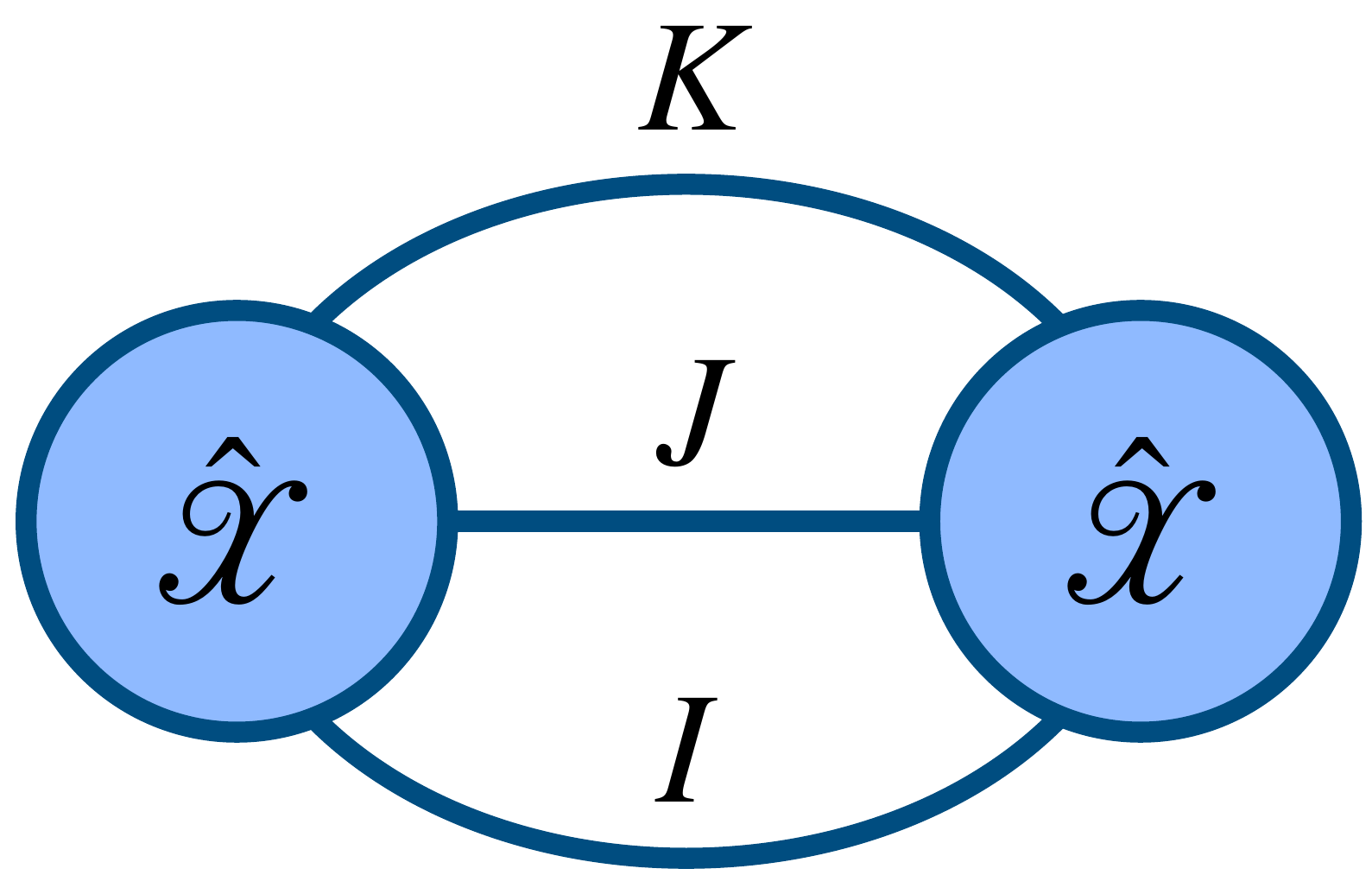}
		\caption{inner-product (tensors)}\label{fig:diagram-inner-tensor}
	\end{subfigure}
	\caption{Representation of some tensor contractions using tensor diagrams}\label{fig:diagram-contraction}
\end{figure}

By employing tensor diagrams, tensor contraction over a common dimension of two tensors is represented by connecting the two corresponding edges. For instance, given matrices \(\mymatrix{A} \in \myR^{I\times J}, \mymatrix{B} \in \myR^{J\times K}\), we can express simply their product \(\mymatrix{C} = \mymatrix{A} \mymatrix{B} = \sum_{j=1}^{J} \mathbf{a}_{:,j} \mathbf{b}_{j, :}^{\myT}\), as illustrated in Fig. \ref{fig:diagram-matrix-product}.

\subsection{Matrix rank}
The rank of a matrix can be defined in several equivalent ways. Hereby we give a simple definition:
\begin{definition}[Matrix rank]\label{D:matrix_rank}
Let $\mymatrix{X} \in \myR^{I \times J}$ an $I \times J$ matrix of real numbers. The \textbf{rank} of $\mymatrix{X}$ is denoted  as rank$(\mymatrix{X})$. It is  equivalently defined as:
\begin{itemize}
    \item The number of linearly independent columns of $\mymatrix{X} $
    \item The number of linearly independent rows of $\mymatrix{X} $
\end{itemize}
\end{definition}
The above definition immediately implies that if
$\mymatrix{X} \in \myR^{I \times J}$, then rank$(\mymatrix{X}) \leq \min(I, J)$.
If rank$(\mymatrix{X})  = \min(I, J)$ then $\mymatrix{X} $ is full-rank.

\subsection{Norms}

One of the major families of tensor norms, of any order, is \textit{element-wise} norms.   

\begin{definition}[$\ell_p$ norm]
The element-wise $\ell_p$ of $\mytensor{X}$ is:
\begin{equation}\nonumber
\Vert \mytensor{X} \Vert_p = \Vert \text{vec}(\mytensor{X}) \Vert_p = \left( \sum_{i_1,i_2,\cdots, i_{N}} \vert \mytensor{X}_{i_1,i_2,\cdots, i_{N}} \vert^p \right)^{1/p}
\end{equation}
\end{definition}
For $p=0$, the $\ell_0$ (pseudo)-norm, denoted $\Vert \mytensor{X}\Vert_0$ returns the number of non-zero elements of a tensor, acting as a measure of the density (and conversely of the sparsity) of the tensor. However, it is not exactly a norm in the mathematical sense (it is positive definite and satisfies the triangle inequality but is not absolutely homogeneous), hence the name pseudo-norm.  
When $p=1$ the $\ell_1$-norm is denoted by $\Vert \mytensor{X}\Vert_1$ and is defined as the sum of the absolute values of the tensor's elements. The $\ell_1$-norm is the tightest  convex surrogate of $\ell_0$-norm \cite{donoho2006sparse} and  is used as a convenient measure of sparsity in practice.
For $p=2$, the  {\it Frobenius norm}  is denoted as $ \Vert \mymatrix{X} \Vert_F$ for matrices and it is generalized to higher-order tensors by the {\it tensor norm}, denoted $\Vert \mytensor{X} \Vert$.

The family of \textit{Schatten-$p$} norms are defined for matrices and act as functions of the \textit{singular values} (i.e., spectrum) of the matrix. Therefore, Schatten-$p$ norms can be used to control the spectral properties of matrices.

\begin{definition}[Schatten-$p$ norm]
Let $[\sigma_1, \sigma_2, \ldots, \sigma_{\min(I,J)}]\myT$ be a vector of the singular values of $\mymatrix{X} \in \myR^{I\times J}$. A Schatten-$p$ norm is obtained by taking the $p$ norm of the singular values:
\begin{equation}\nonumber
\Vert \mymatrix{X}\Vert_p = \left( \sum_i^{\min(I,J)} \sigma_i^p \right)^{1/p}
\end{equation}
\end{definition}
For $p=1$, the Schatten-$1$ norm is referred to as the \emph{nuclear norm} and denoted by $\Vert \mymatrix{X} \Vert_*$.
It is defined as the sum of the singular values of  $\mymatrix{X}$ and is the tightest convex envelope of the rank function~\cite{fazel2002matrix}.


\section{Computational Infrastructure and Tools}
\label{sec:comptools}
In this section, we discuss the necessary infrastructure for computer vision and deep learning. Also, to accelerate the learning curve and enable researchers and practitioners to get started in the field of  tensor methods quickly, we review available software packages for tensor algebra and introduce 
the TensorLy library~\cite{tensorly},
with which we implement ready-to-use examples that accompany the paper.

\subsection{Hardware}

\looseness-1Enabling deep learning, computer vision, and machine learning start in general down at the hardware level. Indeed,  modern computer vision is enabled by the ``trinity of AI'': \emph{data}, with the introduction of large bodies of annotated data; \emph{algorithms}, with the introduction of convolutional neural networks, recurrent units, etc. and finally \emph{hardware}, with the use of graphical processing units (GPUs). The latter was crucial in scaling up computation and enabling training on big (visual) data consisting of millions of images.

Historically, datasets used in computer vision were relatively small and could easily fit in the memory of most commodity personal computers. However, with the advent of deep learning and large bodies of annotated images, such as ImageNet~\cite{krizhevsky2012imagenet}, it became crucial to have large amounts of memory. 
Most importantly, to be able to process the information efficiently, it became paramount to be able to process multiple bits of information in parallel. However, central processing units (CPUs) are inherently sequential. 
First proposed in~\cite{raina2009large}, then successfully used on MNIST~\cite{gpu_mnist} and more prominently, on ImageNet~\cite{krizhevsky2012imagenet}, it is the use of GPUs that revolutionized deep learning, making it feasible to train a very large model on millions of samples. While CPUs perform operations on tensors in a mostly sequential way and prioritize low latency, GPUs favor high throughput and accelerate operations by running them efficiently in parallel: modern CPUs typically contain up to $16$ cores, while a GPU has thousands of them, allowing for the performance of hundreds of TFLOPS. 

Efficient libraries then enable to efficient use of the tensor cores composing GPUs. In particular, extending Basic Linear Algebra Subprograms (BLAS) primitives allows to significantly speed up computation by leveraging hardware acceleration~\cite{shi2016tensor}. The recent cuTensor~\cite{cutensor} is a high-performance CUDA library that leverages GPU-acceleration for efficient tensor operations. In particular, it obviates the need for costly reordering of the elements and reshaping when performing operations such as tensor contraction for which it provides native implementations.

Further, large models typically need to be trained on multiple GPUs distributed in various machines. 
However, distributed training and inference across multiple machines introduces new challenges and can be incredibly complex to implement correctly. Fortunately, modern deep learning libraries abstract such technicalities and allow the end-user to focus on the model's logic. Notable such frameworks include PyTorch~\cite{pytorch}, which offers a functional approach to deep learning, and TensorFlow~\cite{tensorflow}, which uses a symbolic one. Others such as MXNet~\cite{mxnet} enable a hybrid approach.

While CPU and GPU are the most widely used mediums for training and inference in computer vision and deep learning tasks, other hardware types can be used towards this end.  Examples include application-specific integrated circuits (ASICs)  designed specifically for running a specialized set of instructions (e.g., TPUs~\cite{jouppi2017datacenter}), or field-programmable gate arrays (FPGAs) that conversely can be dynamically programmed depending on the application.

\subsection{Software}
In the previous section, we covered hardware support,
however, there is a considerable interplay between hardware and software, and one cannot function well without the other. Good software without optimized hardware support would result in slow training, while sub-par software with poor API, documentation, and tests is unusable in practice. An example of this interplay is the organization of elements in memory. Memory can be over-simplified as one long vector of numbers.  The way these elements are ordered in memory influences how fast operations are executed. For instance, to store a matrix, its elements can be arranged row-after-row (C-ordering) or column-after-column (also called Fortran ordering). In NumPy, elements are organized by default in row-order--the same for PyTorch. 
This, in turn, influences higher-level operations: unfolding, for instance, needs to be adapted to appropriate ordering to avoid expensive reordering of the data. As a result, much effort has been spent developing user-friendly libraries that provide state-of-the-art algorithms, neatly coded and wrapped in an intuitive, well-tested API. One notable such library is TensorLy~\cite{tensorly}. 

\begin{figure}[h]
    \centering
    \includegraphics[width=1\linewidth]{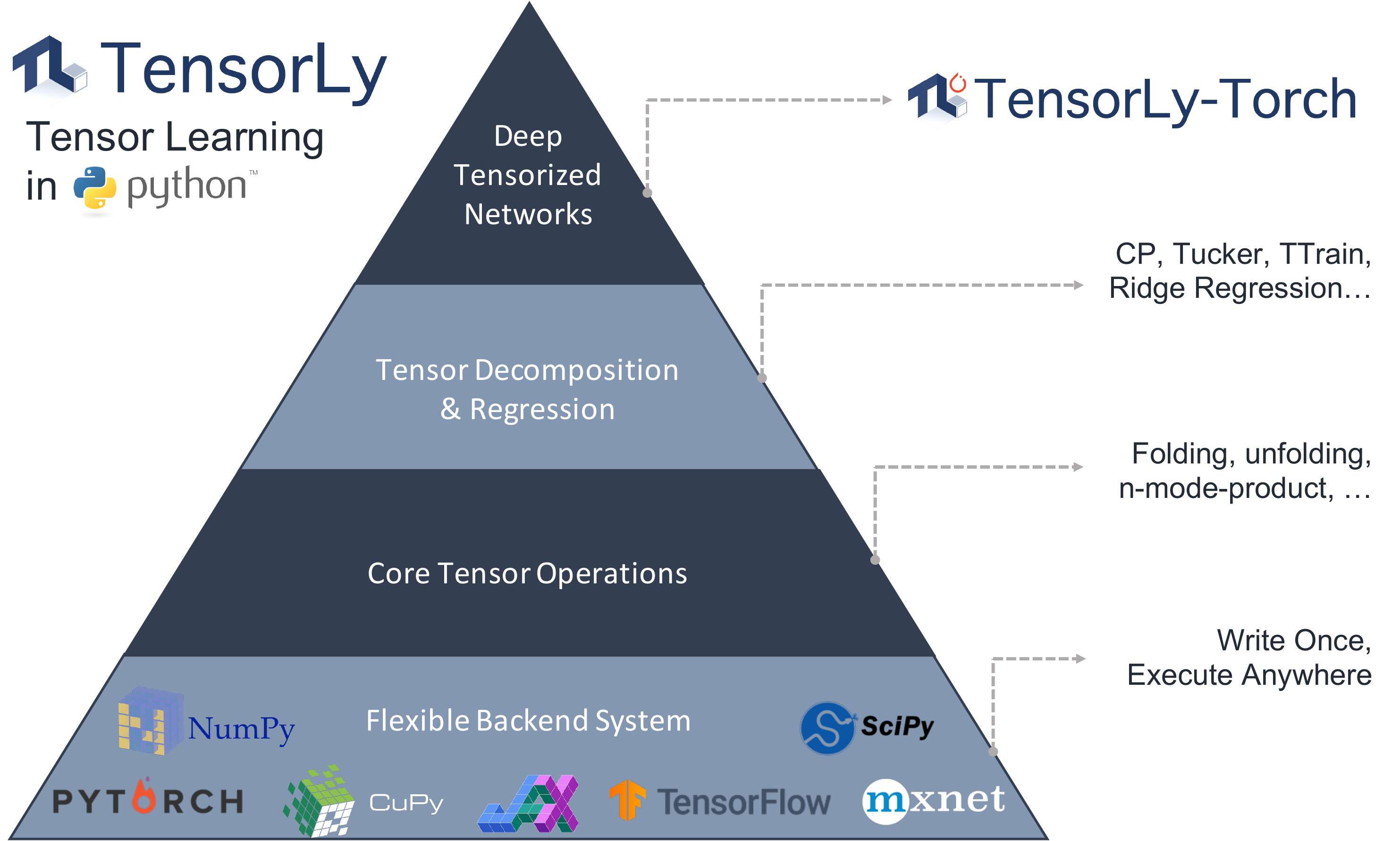}
    \caption{Overview of the TensorLy and TensorLy-Torch libraries}
    \label{fig:tensorly}
\end{figure}

TensorLy is a Python library that aims at making tensor learning simple and accessible. It provides a high-level API for tensor methods, including core tensor operations, tensor decomposition, regression, and deep tensorized architectures. It has a flexible backend that allows running operations seamlessly using NumPy, PyTorch, TensorFlow, etc. The project is also open-sourced under the BSD license, making it suitable for both academic and commercial use.
TensorLy is also optimized for Python and its ecosystem. For instance, the unfolding is redefined as introduced in Section \ref{sec:preliminaries}
to match the underlying C-ordering of the elements in memory, resulting in better performance. An overview of the functionalities of TensorLy can be seen in Figure~\ref{fig:tensorly}.
By default, it uses NumPy as its backend. However, this can be set to any of the deep learning frameworks, optimized for running large-scale methods, and in the companion, we also use the popular PyTorch library. Once PyTorch has been installed, it can be easily used as a backend for TensorLy, allowing all the operations to run transparently on multiple machines and GPU. Finally, TensorLy-Torch\footnote{\url{http://tensorly.org/torch/dev/}}~\cite{tensorly-torch} builds on top of TensorLy and PyTorch and implements out-of-the-box layers for tensor operations within deep neural networks.

Other libraries exist for tensor operations in Python. In particular, t3f~\cite{novikov2020tensor} is a TensorFlow library for working with the tensor-train decomposition on CPU and GPU. It also supports Riemannian optimization operations. Tntorch~\cite{tntorch} is a library for tensor networks in PyTorch, while TensorNetwork~\cite{tensornetwork} additionally supports JAX, TensorFlow and NumPy. Scikit-Tensor~\cite{sktensor} is a NumPy library for tensor learning, supporting several tensor decompositions.

In MATLAB, packages such as the Tensor Toolbox~\cite{tensor_toolbox_kolda} or TensorLab~\cite{tensorlab_lathauwer} provide extensive support for tensor decomposition. In C++, ITensor~\cite{itensor} provides efficient support for tensor network calculations.

Finally, efficient operations on sparse tensors require specialized implementations. Indeed, while libraries such as NumPy are highly-optimized for matrix and tensor contractions, they do not support operations on sparse tensors. Explicitly representing sparse tensors as dense arrays is very inefficient in terms of memory use and prohibitive for very large tensors. Instead, these can be represented efficiently, e.g., using the COOrdinate format which instead of storing all the elements only stores the coordinates and values of non-zero elements. As a result, implementing them efficiently can be tricky. While there is a large body of work focusing on sparse matrix support, support for sparse tensor algebra is more scant. Libraries such as the PyData Sparse\footnote{\url{https://sparse.pydata.org/}} for CPU and the Minkowski Engine~\cite{choy20194d} for GPU are being developed and provide an efficient way to leverage large, sparse multi-dimensional data. TensorLy also supports sparse tensor operations on CPU through the PyData Sparse library.


\section{Representation Learning with Tensor Decompositions}
\label{sec:replearning}
\subsection{Matrix decomposition and representation learning}
\label{sec:replearning:matdecomp}

Real-world visual data exhibit complex variability
due to many factors related to visual objects' appearance (e.g., illumination and pose changes, rigid and non-rigid deformations), image structure,  semantics, and noise. Hence, useful information is hidden in high-dimensional visual measurements.
Representation learning methods seek to extract useful information from such high-dimensional data through simple, low-dimensional representations (i.e., data are expressed as a linear combination of basic yet fundamental structures)  that simultaneously exhibit a set of desirable properties such as invariance to nuisance variability while at the same time recovering factors of variation in a meaningful manner for a specific task. In practice, low-rank and sparsity are popular simplicity measures used in representation learning models.

{\it Matrix decomposition} forms the mathematical backbone in learning parsimonious, low-dimensional representations from high-dimensional vector data samples, such as images flattened into vectors. {\it Component analysis}, e.g., \cite{delatorre2012leastsquares} and {\it sparse coding}, e.g., \cite{zhang2015sparseRepSurvey,mairal2010online_learning} are prominent examples of representation learning methods that extract simple data representations by means of low-rank matrix decompositions and sparse, possibly overcomplete representations, respectively.
Given $M$ vector data samples $\{\myvector{x}_m\}_{m=1}^{M}$,
where each sample $\myvector{x}_m$ is an $I$-dimensional vector, that is $\myvector{x}_m \in \myR^I_m$, such a dataset can be represented by
a data matrix $\mymatrix{X} \in \myR^{I \times J}$ which contains in its columns data vectors.  Matrix decompositions seek to factorize $\mymatrix{X}$ as
a product of two factor matrices $\mymatrix{U}^{(1)} \in \myR^{I \times R }$ and $\mymatrix{U}^{(2)} \in \myR^{J \times R }$, namely
 \begin{equation}\label{E:MF}
\mymatrix{X} = \mymatrix{U}^{(1)} {\mymatrix{U}^{(2)}}\myT.
\end{equation}

Assuming $\mymatrix{X}$ is of low-rank $R < \min\{I,J\}$,
Eq.~(\ref{E:MF}) is referred to as {\it low-rank matrix decomposition}. Hence, $\mymatrix{X}$  can be expressed as a sum of $R$ rank-one matrices:
$\mymatrix{X} = \mymatrix{U}^{(1)} {\mymatrix{U}^{(2)}}\myT  = \sum_{r=1}^R \myvector{u}^{(1)}_r \circ {\myvector{u}^{(2)}_r}  = \sum_{r=1}^{R}\myvector{u}^{(1)}_r {\myvector{u}^{(2)}_r}\myT.$
The concept of low-rank matrix decomposition 
is generalized for higher-order tensors by low-rank tensor decomposition which will be discussed next in Section \ref{sec:replearning:tensordecomp}. Interestingly, as opposed to matrix decomposition which is {\it not unique}\footnote{To verify this consider any invertible matrix
$\mymatrix{Q} \in \myR^{R\times R}$. Then we have  $\mymatrix{X} = \mymatrix{U}^{(1)} {\mymatrix{U}^{(2)}}\myT = 
\mymatrix{U}^{(1)} \mymatrix{Q} {\mymatrix{Q}}^{-1} {\mymatrix{U}^{(2)}}\myT.$} low-rank tensor decomposition, such as the Canonical-Polyadic  decomposition, also known as Parallel Factor Analysis (PARAFAC) \cite{hitchcock1927tensor_polyadic,carroll1970AnalysisOI,Hars1970}  is unique under mild algebraic assumptions~(see \cite{sidiropoulos2017tensor} and references therein).

In practice, data matrices representing visual data sets are not {\it exactly} low-rank. The observed matrix can deviate from the low-rank structure for several reasons, including noise, outliers, and non-linear structures underlying the data. In such cases, we are interested in {\it approximating} the observed matrix with a low-rank matrix expressed as a product of two factor matrices. Depending on the application,  these elements can be interpreted either as a basis of a low-dimensional subspace or as a mixing operator (i.e., mixing matrix) that combines hidden factors of variation, giving rise to the observed data. Such components comprise a low-dimensional, and hence a simplified representation of the data in the subspace spanned by the basis elements. To ensure uniqueness and make basis elements and/or components physically or semantically interpretable, constraints such as orthogonality, nonnegativity, statistical independence, smoothness, or sparsity are usually imposed on components. For instance, in principal component analysis (PCA) \cite{pearson1901on_lines_pca,Hotelling_1933}  basis elements are imposed to be column orthonormal. In independent component analysis (ICA) \cite{hy2000ica_algorithms} latent components should be statistically independent,  while in non-negative matrix factorization (NMF) \cite{lee1999nonneg_matrix} they are imposed to be non-negative. Furthermore, additional constraints can be used to reflect topological or class-specific properties of the data \cite{yan2007graph_embedding},
as well as spectral \cite{candes2011rpca}
or temporal patterns \cite{zhang2012slow_feature}.

The vast majority of the component analysis models mentioned above employ the least-squares criterion  \cite{delatorre2012leastsquares} which assumes that data are contaminated by Gaussian noise of small variance. However, such an assumption is often not valid for visual data, as we will see in Section \ref{sec:replearning:robust}, where robust to non-Gaussian noise tensor decomposition is discussed. Extensions of component analysis methods to high-order tensors are presented in Section \ref{sec:replearning:tensorcomponent}.

\looseness-1Apart from low-rank, sparsity is another widely employed low-dimensional structure in representation learning.
Sparse coding, in particular, aims at finding a sparse representation of the input vector data in the form of a sparse linear combination of basic elements. These elements are called atoms, and they compose a dictionary. Atoms in the dictionary (i.e., columns of $\mymatrix{U}^{(1)}$) are not required to be orthogonal, and they may comprise an over-complete spanning set. Learning the dictionary and the linear combination coefficients (i.e., the sparse code represented in columns of ${\mymatrix{U}^{(2)}}\myT$)  is referred to as dictionary learning \cite{zhang2015sparseRepSurvey,mairal2010online_learning,mensch2016massive_matrix}.
For signals such as natural images, it is now well established that sparse representations are well suited to  classification, and restoration tasks \cite{zhang2015sparseRepSurvey,mairal2010online_learning}. In Section \ref{sec:replearning:diclearn} we discuss tensor-structured dictionary learning that generalizes the concept of dictionary learning to higher-order tensors.

\looseness-1In multivariate linear regression models, enforcing low-rank constraints on the coefficient matrix offers an effective reduction of unknown parameters, which facilitates reliable parameter estimation and model interpretation. Likewise, low-rank structures can be imposed in tensor regression models such as those presented in Section \ref{ssec:tensor-regression}.

\subsection{Tensor decomposition}
\label{sec:replearning:tensordecomp}
Tensor decomposition extends the concept of matrix decomposition to higher-order tensors. Here, we give a brief overview of the three most fundamental tensor decomposition models that are most useful in the context of this paper. For more details on these models, the interested reader is referred to \cite{papalexakis2016tensors, kolda2009tensor, sidiropoulos2017tensor, 7038247}, where algorithmic matters are also covered in detail.  We introduce all three decompositions for an $N\myth$ order tensor \(\mytensor{X} \in \myR^{I_1 \times I_2 \times \cdots \times I_N} \).
 
\subsubsection{Canonical-Polyadic (CP) decomposition}\label{sssec:CP}
The CP decomposition, also referred to as PARAFAC, \cite{hitchcock1927tensor_polyadic,carroll1970AnalysisOI,Hars1970}, decomposes \(\mytensor{X}\)
into a sum of \(R\) rank-one tensors. The rank of a tensor \(\mytensor{X}\) is the minimum number of rank-one tensors that sum to  \(\mytensor{X}\) and generalizes the notion of matrix rank (cf. Definition \ref{D:matrix_rank}) to high-order tensors. However, computing the tensor rank is NP-hard \cite{haastad1990tensor,hillar2013most}.
Hence, a predefined rank should be known in advance or estimated from data using Bayesian approaches, e.g., \cite{zhao2015bayesiancp,rai2014scalable_bayesian_lowrank}
or deep neural networks \cite{zhou2019tensor_rank_cnn}. 
When directly learning the decomposition end-to-end with stochastic gradient descent, the rank parameter is validated along with other parameters.

\begin{figure}
    \centering
    \includegraphics[width=1\linewidth]{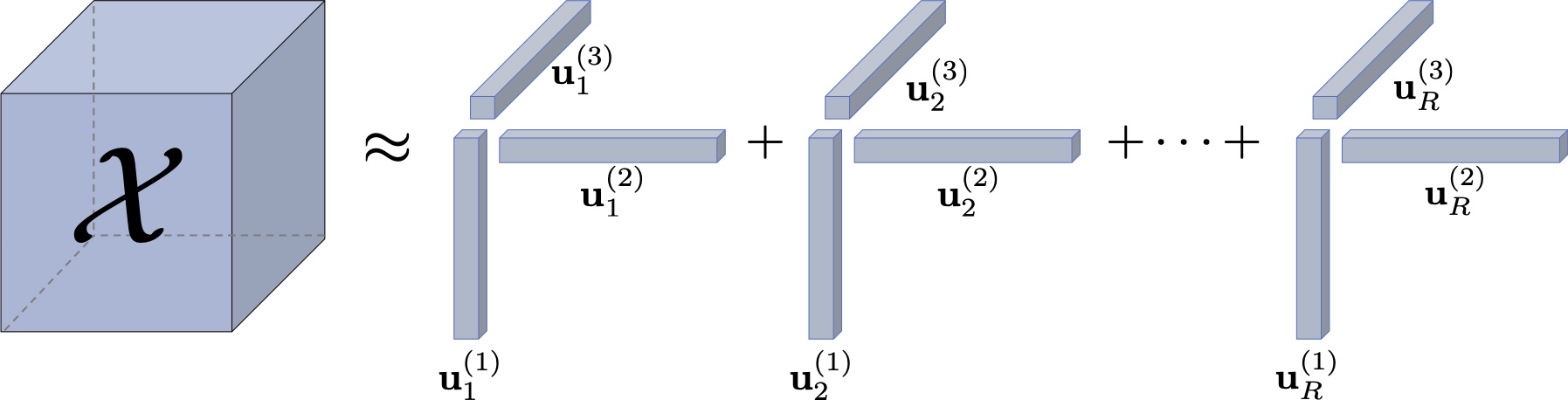}
    \caption{\textbf{Illustration of a CP decomposition} of third-order tensor  \(\mytensor{X}\)  into a sum of rank-1 tensors.}
    \label{fig:CP}
\end{figure}

Formally, the CP decomposition seeks to find the vectors 
\( \mathbf{u}^{(1)}_{r} , \mathbf{u}^{(2)}_{r}, \cdots, \mathbf{u}^{(N)}_{r}\),  such that:
\begin{equation}
\mytensor{X}
= \sum_{r=1}^{R}  \underbrace{ \myvector{u}^{(1)}_{r} \circ \myvector{u}^{(2)}_{r} \circ \cdots \circ \myvector{u}^{(N)}_{r} }_{\text{rank-1 components (tensors)}}.
\end{equation}
These vectors can be collected into $N$ matrices, $\{\mathbf{U}^{(n)}\in \mathbb{R}^{I_n \times R}\}_{n=1}^N$, with each matrix defined as:
\begin{align}
\mathbf{U}^{(n)} &= \left[ \begin{matrix} \mathbf{u}^{(n)}_{1}, \mathbf{u}^{(n)}_{2},  \cdots, \mathbf{u}^{(n)}_{R} \end{matrix} \right].
\end{align}
The CP decomposition is denoted more compactly as \(
\mytensor{X} = \mykruskal{\mathbf{U}^{(1)}, \mathbf{U}^{(2)}, \cdots, \mathbf{U}^{(N)}}
\) \cite{kolda2009tensor}. Representing a tensor using its CP decomposition is sometimes referred to as Kruskal format. A pictorial representation of the CP decomposition is shown in Fig.~\ref{fig:CP}.

\

\subsubsection{Tucker decomposition}

\begin{figure}
    \centering
    \includegraphics[width=0.9\linewidth]{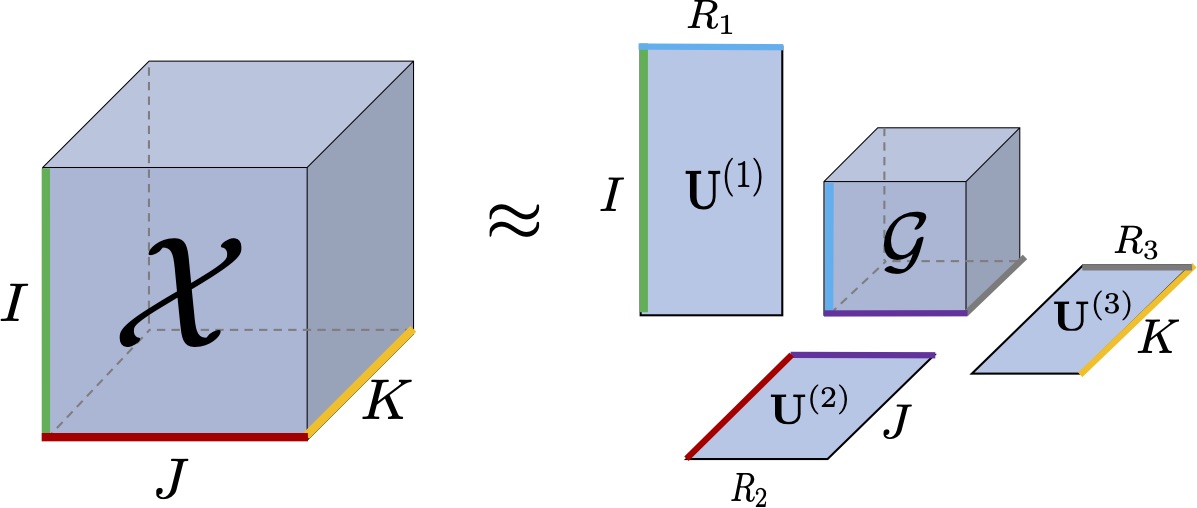}
    \caption{\textbf{Illustration of a Tucker decomposition} of a third-order tensor \(\mytensor{X}\) into a core tensor and three matrices.}
    \label{fig:tucker}
\end{figure}

The Tucker decomposition \cite{Tucker1966SomeMN} decomposes, non-uniquely,  \(\mytensor{X}\) into a core tensor
\(\mytensor{G} \in \myR^{R_1 \times R_2 \times \cdots \times R_N}\) and a set of factor matrices
\( \left( \mymatrix{U}^{(1)}, \mymatrix{U}^{(2)}, \cdots,\mymatrix{U}^{(N)} \right) \), with \(\mymatrix{U}^{(n)} \in \myR^{R_n \times I_n}, n=1,2,\ldots, N \) which are multiplied with \(\mytensor{G}\) along its modes  as follows:
\begin{align}
\mytensor{X} &= 
\mytensor{G} \times_1 \mymatrix{U}^{(1)} 
		  \times_2  \mymatrix{U}^{(2)} \times_3
		  \cdots
          \times_N \mymatrix{U}^{(N)}.
\end{align}
The core tensor captures interactions between the columns of factor matrices. If $R_n \ll I_n$, $\forall n$,
then the core tensor can be viewed as a compressed version of \(\mytensor{X}\). It is also worth noting that the CP decomposition can be expressed as Tucker decomposition, with $R_n = R$, $\forall n \in \{1,2, \ldots, N \}$ and the core tensor being superdiagonal. The Tucker decomposition is denoted compactly as \(
\mytucker{\mytensor{G}}{\mymatrix{U}^{(1)}, \mymatrix{U}^{(2)}, \cdots, \mymatrix{U}^{(N)}} 
\) \cite{kolda2009tensor}, and it is shown pictorially in Fig.~\ref{fig:tucker}. By imposing the factor matrices to be orthonormal, the Tucker model is known as higher-order singular value decomposition (HOSVD)
\cite{lathauwer2000msvd}.

In practice, data are corrupted by noise, and
as in the matrix case, both the Tucker and CP decompositions are not exact. Therefore, they are approximated by optimizing a suitable criterion representing fitting loss. Typically, the least-squares criterion is employed. In this case, assuming that $\{ R_n \}_{n=1}^N$ are known, the Tucker decomposition is computed by solving the following non-convex minimization problem:
\begin{equation}\label{E:Tucker_OPTIMIZATION}
    \min_{\mytensor{G}, \{\mymatrix{U}^{(n)}\}_{n=1}^N } \Vert \mytensor{X} -\mytucker{\mytensor{G}}{\mymatrix{U}^{(1)}, \mymatrix{U}^{(2)}, \cdots, \mymatrix{U}^{(N)}}  \Vert^2. 
\end{equation}

The most common optimization algorithm for the Tucker and CP decomposition is the Alternating Least Square (ALS)~\cite{mohlenkamp2013musings}, where the unknowns are estimated iteratively. More specifically, the ALS fixes,
sequentially, all but one factor, which is then updated by solving a linear least-squares problem. However, the ALS algorithm is not guaranteed to convergence towards a minimum; see \cite{comon2014tensors} for example.

\subsubsection{Tensor-Train}
\begin{figure}
    \centering
    \includegraphics[width=0.9\linewidth]{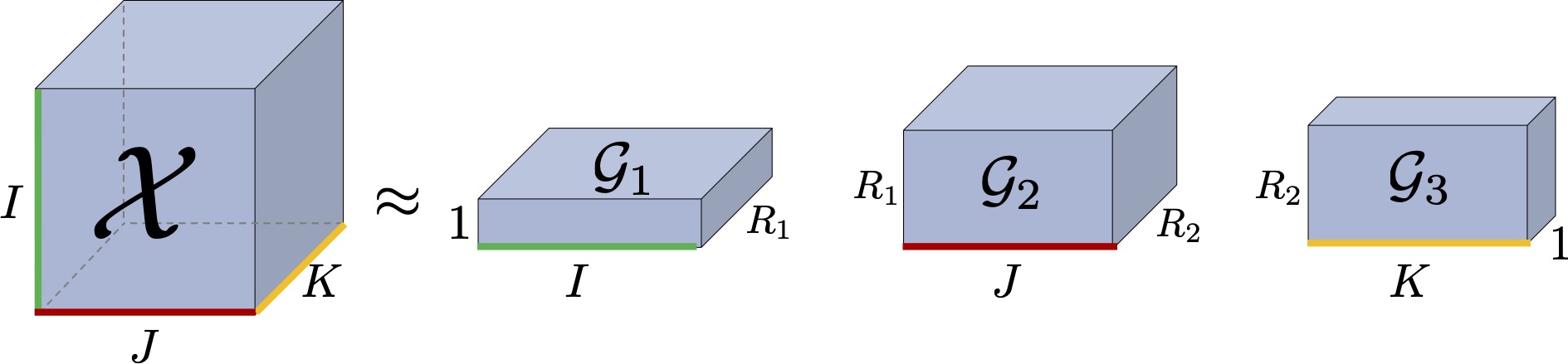}
    \caption{\textbf{Illustration of a Tensor-Train decomposition} of a third-order tensor \(\mytensor{X}\) into a series of third-order cores. The boundary conditions dictate $R_1 = R_{N+1} = 1$}
    \label{fig:mps}
\end{figure}

The Tensor-Train (TT) decomposition~\cite{oseledets2011tensor}, also known as the Matrix-Product-State~\cite{rommer1997class} in quantum physics, expresses \(\mytensor{X} \) as a product of third order tensors \(\mytensor{G}_1 \in \myR^{R_1 \times I_1 \times R_2}, \cdots,  \mytensor{G}_N \in \myR^{R_N \times I_N \times R_{N+1}}\) (the \emph{cores} of the decomposition), such that:
\begin{equation}
    \mytensor{X}(i_1, i_2, \cdots, i_N) =
        \underbrace{\mytensor{G}_1[i_1]}_{R_1 \times R_2}
        \times
        \underbrace{\mytensor{G}_2[i_2]}_{R_2 \times R_3}
        \times 
        \cdots
        \times
        \underbrace{\mytensor{G}_N[i_N]}_{R_{N} \times R_{N+1}}. \nonumber
\end{equation}
The boundary conditions of the tensor-train (\emph{open boundary conditions}) decomposition dictate \(R_1 = R_{N+1} = 1\). The decomposition is illustrated in Fig.~\ref{fig:mps}. The open boundary conditions can be replaced by periodic boundary conditions (the \emph{Tensor-Ring})~\cite{espig_note_2012,zhao2016tensor} which instead contracts together the first and last cores along \(R_1\) and \(R_{N+1}\).

\begin{figure}
    \centering
    \includegraphics[width=0.8\linewidth]{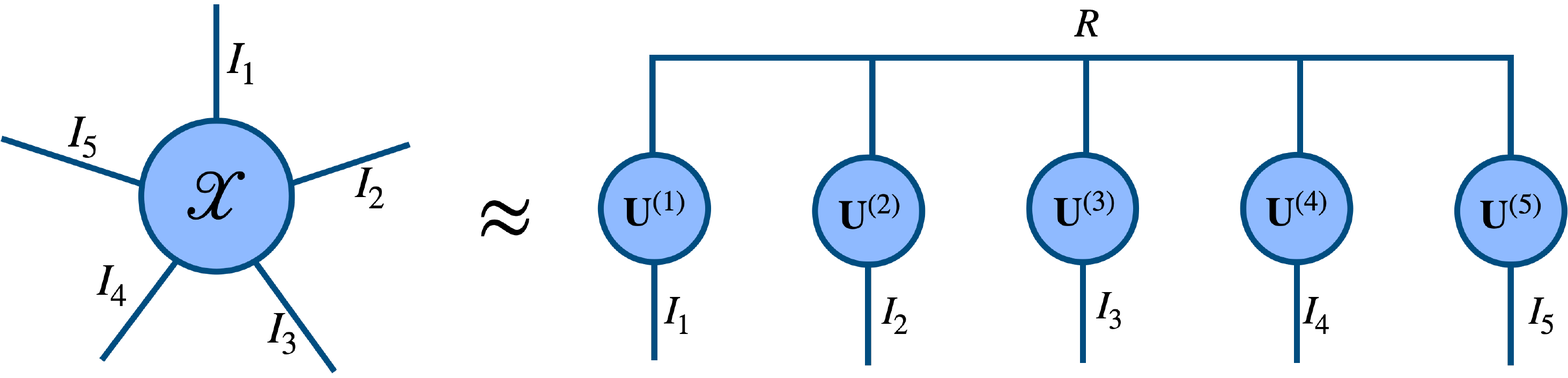}
    \caption{Representation of a CP decomposition using tensor diagrams.}\label{fig:diagram-cp}
\end{figure}

\begin{figure}
    \centering
    \includegraphics[width=0.8\linewidth]{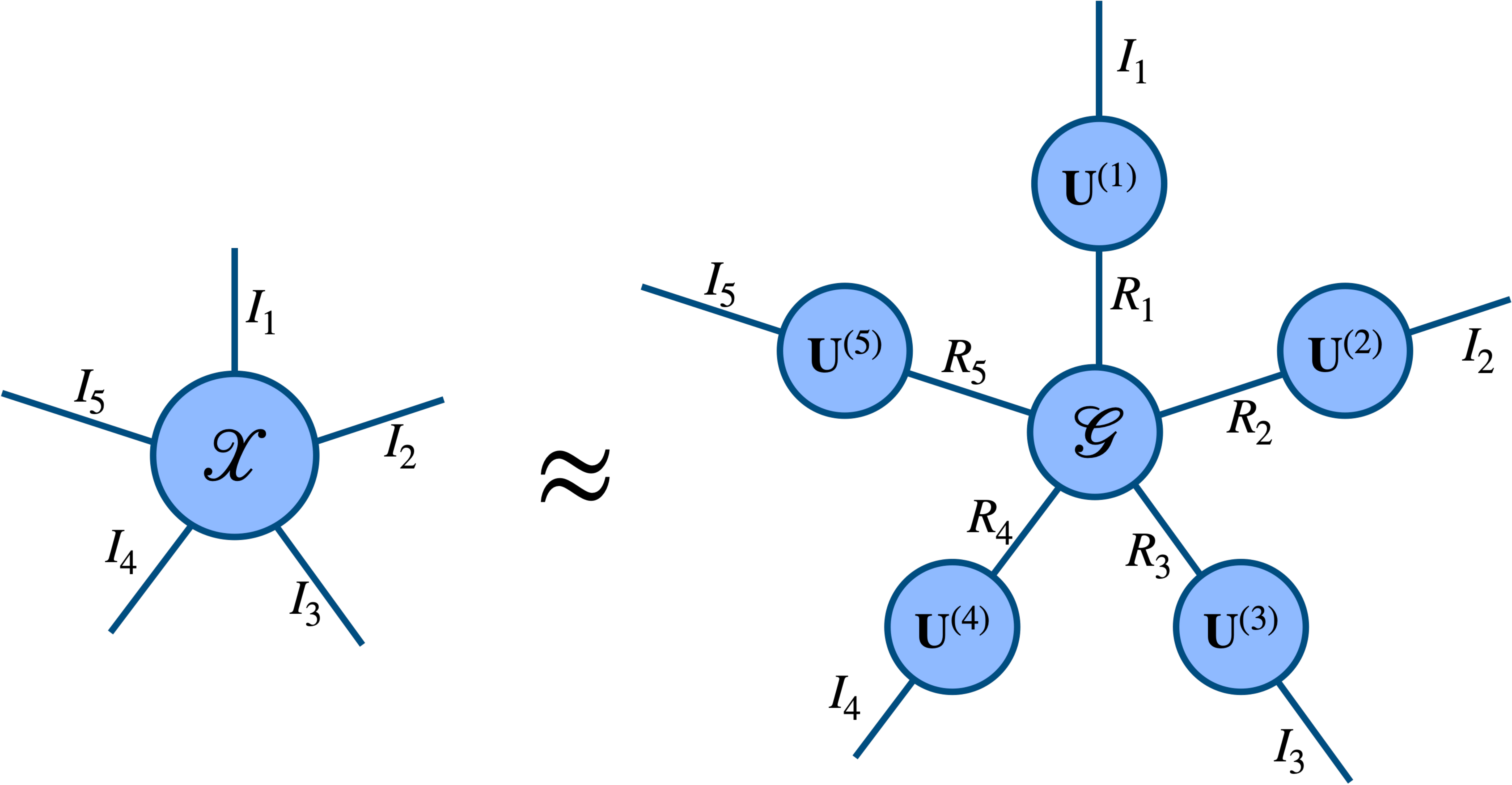}
    \caption{Representation of a Tucker decomposition using tensor diagrams.}\label{fig:diagram-tucker}
\end{figure}

\begin{figure}
    \centering
    \includegraphics[width=0.8\linewidth]{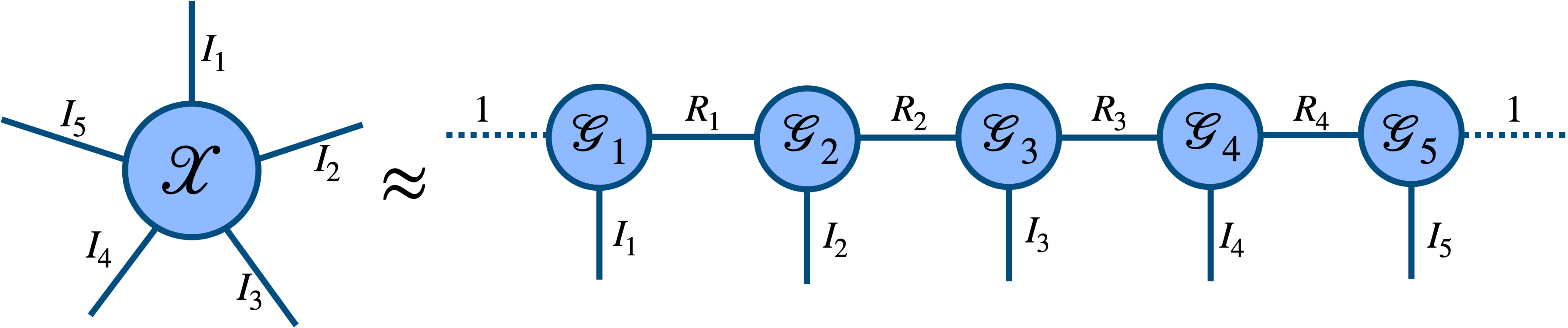}
    \caption{Representation of a Tensor-Train decomposition using tensor diagrams.}\label{fig:diagram-mps}
\end{figure}

Let us now depict the aforementioned tensor decompositions using tensor diagrams.   While representing tensor decompositions visually (e.g., Figures \ref{fig:CP} and \ref{fig:tucker}), we are inherently limited to a maximum of third-order tensors, tensor diagrams 
conveniently represent decompositions of higher-order tensors.
For instance, let \(\mytensor{X} \in \myR^{I_1\times I_2\times I_3\times I_4\times I_5}\) be a \(5^{\text{th}}\)-order tensor, its
CP decomposition is shown in Fig.~\ref{fig:diagram-cp}, Tucker decomposition
is shown in Fig.~\ref{fig:diagram-tucker},  and  Fig.~\ref{fig:diagram-mps} depicts its TT decomposition.

\subsection{Robust tensor decomposition}
\label{sec:replearning:robust}

\notebook{rpca}

Visual data of interest, such as facial images captured in unconstrained, real-world conditions, are often severely corrupted by nuisance factors of variation. Such factors include, but are not limited to, significant pose variations, illumination changes, and occlusions, as well as artifacts or missing data introduced during the acquisition process.  Collectively, these can be characterized as gross errors. From a statistical point of view, gross errors and outliers do not follow a Gaussian distribution, as is assumed in the vast majority of learning models employed in computer vision--including the tensor decompositions discussed above. Indeed, learning models that rely on optimization of algebraic criteria induced by second-order statistics of observations (e.g., covariance matrix), or goodness of fit measures such as the least-squares criterion, are unable to cope with gross errors and outliers \cite{huber2011robust}. As a result, the estimated components can be arbitrarily far from the true ones. Hence, their applicability in analyzing visual data captured in unconstrained conditions is rather limited.

\looseness-1Robust tensor decomposition methods aim to recover a low-rank tensor from noisy measurements, assuming the presence of non-Gaussian noise of possibly large magnitude. Those outliers only affect a small fraction of the measurements, and consequently, noise is considered sparse. Concretely, robust tensor modeling seeks a decomposition of the form:
\begin{equation}
\mytensor{X} = \mytensor{L} +\mytensor{S},
\end{equation}
where $\mytensor{X}$ is an $N\myth$-order data tensor usually composed by arranging a set of tensorial data samples of $N-1\myth$ order along the $N\myth$ mode of the tensor. $\mytensor{L}$ and
$\mytensor{S}$ are unknown $N\myth$-order tensors accounting for the low-rank tensor and sparse noise, respectively.

However, as mentioned previously optimizing tensor rank is intractable \cite{haastad1990tensor} in general. To address this \cite{gandy2011tensor,liu2012tensor,yang2013fixed,goldfarb2014robust_models} rely on approximating  tensor rank by a convex combination of the n-ranks of $\mytensor{L}$. That is,
$
\sum_{n=1}^{N} \alpha_{n} \operatorname{rank}_{n}(\mytensor{L}),
$
where 
$ \alpha_{n} \geq 0, \sum_{n=1}^{N} \alpha_{n}=1$ and $\operatorname{rank}_{n}(\mytensor{L})$
denotes the column rank of the mode-$n$ matricization  of $\mytensor{L}$. Therefore, the low-rank tensor is recovered by solving:
\begin{equation}\label{E:TRPCA_NC}
\min _{\mytensor{L}, \mytensor{S}} \sum_{n=1}^{N} \alpha_{n} \operatorname{rank}_{n}(\mytensor{L})+\lambda\Vert\mytensor{S}\Vert_{0} \quad \text {s.t.} \quad \mytensor{X}=\mytensor{L}+\mytensor{S},
\end{equation}
where $\lambda$ is a positive parameter.
However, both the $\operatorname{rank}_n$ function and the $\ell_0$-norm are discrete functions and hence hard to optimize \cite{boyd1996semidefinite,natarajan1995sparse}. To make (\ref{E:TRPCA_NC}) tractable, discrete norms are replaced by their convex surrogates \cite{fazel2002matrix,donoho2006sparse}. That is, the rank function is replaced by the nuclear norm and the $\ell_0$-norm by the $\ell_1$-norm, resulting into a convex optimization problem:
\cite{li2010optimum,goldfarb2014robust_models}:
\begin{equation}\label{E:TRPCA_C}
\min _{\mytensor{L}, \mytensor{S}} \sum_{n=1}^{N} \alpha_{n}\Vert\mymatrix{L}_{(n)}\Vert_{*}+\lambda\Vert\mathcal{S}\Vert_{1} \quad \text{s.t.} \quad \mathcal{X}=\mathcal{L}+\mathcal{S},
\end{equation}
which is an extension of principal components pursuit (PCP) \cite{candes2011rpca} to higher-order tensors.  It is also worth noting that under certain conditions (\ref{E:TRPCA_C}) is guaranteed to exactly recover the low-rank component \cite{bo2015provable}. Similar models has been introduced for the problems of tensor completion \cite{hu2015new_lowrank,Gandy_2011_low_n_rank,liu2014generalized_HOOI_completion,bo2015provable,liu2012tensor}.

As mentioned in Section~\ref{sec:replearning:tensordecomp}, the CP and Tucker decompositions are approximated in practice by minimizing the least-squares criterion, resulting in algorithms sensitive to sparse, non-Gaussian noise.
To alleviate this problem and obtain more robust approximations of tensor decompositions ~\cite{chachlakis2019l1,chachlakis2020l1,goldfarb2014robust_models} replace least-squares loss with the \(\ell_1\)-norm.

The vast majority of the aforementioned models are optimized  iteratively by employing alternating direction method of multipliers (ADMM) type of algorithms \cite{bertsekas1996lagrange}.

\subsection{Tensor component analysis}
\label{sec:replearning:tensorcomponent}

\notebooks{MPCA}{MDA}

Tensor component analysis (also referred to as multilinear subspace learning \cite{lu2003book}) extends the concept of component analysis to data represented by higher-order tensors (please refer to \cite{lu2003book,lu2011survey} for an overview on the topic). 

Let us assume a set of $M$ tensor data samples, that is
 $\{ \mytensor{X}_m \}_{m=1}^M$, where each data sample is a $N-1$-order
 tensor $\mytensor{X}_m \in \myR^{I_1 \times I_2 \times \cdots \times I_{N-1}}$. Therefore, as mentioned previously, a set of such data is represented by an $N\myth$ order tensor $\mytensor{X} \in \myR^{I_1 \times I_2 \times \cdots \times I_{N}}$. Tensor component analysis methods
seek to estimate  a set of $N-1$ projection matrices,  $\{\mathbf{U}^{(n)} \in  \myR^{R_n \times I_n} \}_{n=1}^{N-1}$ with $R_n \ll I_n \forall n \in \{1,2, \ldots, N-1\}$  such that the projection of tensor data samples onto the a low-dimensional multilinear subspace, namely $\{ \mytensor{X}_m \times_1 \mymatrix{U}^{(1)}  \times_2  \mymatrix{U}^{(2)} \times_3 \cdots \times_{N-1} \mymatrix{U}^{(N-1)} =\mytensor{X}_m  \prod_{n=1}^{N-1} \times_{n} \mymatrix{U}^{(n)}  \in \myR^{R_1 \times R_2 \times \cdots \times R_{N-1}} \}_{m=1}^M$, yields a low-dimensional tensor representation that satisfy some optimality criterion, such as minimizing reconstruction error e.g., \cite{shashua2005non}  or maximizing tensor scatter \cite{lu2008mpca}. Structural constraints, such as orthogonality in Multilinear PCA (MPCA) \cite{lu2008mpca} or non-negativity in Non-Negative Tensor Factorization (NTF) \cite{shashua2005non} and its variants \cite{panagakis2009locality}  are also imposed often on multilinear projections.

\looseness-1 Different tensor component analysis methods can be derived as different solutions to constrained optimization problems. For example, unsupervised tensor component analysis
can be formulated as a constrained low-rank tensor approximation problem:
\begin{equation}
\begin{aligned}\label{E:TCA_Unsupervised}
\min_{\{\mymatrix{U}^{(n)}\}_{n=1}^{N}} &\Vert \mytensor{X} -  \mytensor{G} \times_1 {\mymatrix{U}^{(1)}}\myT 
		  \times_2  {\mymatrix{U}^{(2)}}\myT \times_3 \cdots \times_N {\mymatrix{U}^{(N)}}\myT  \Vert^2 \\
		  &+ \sum_{n=1}^N \lambda_n r_n \left(\mymatrix{U}^{(n)}\right) \quad \mbox{s.t.} \;\; \{\mymatrix{U}^{(n)} \in \;\; \mathcal{C}_n\}_{n=1}^{N},
\end{aligned}
\end{equation}
where $r_n ( \cdot)$ is a regularization function (e.g., $\ell_1$-norm for inducing sparsity), $\lambda_n$ is a positive regularization parameter, and $\mathcal{C}_n$ denotes a set of structural constraints imposed onto the low-dimensional factors. For instance, MPCA \cite{lu2008mpca} is derived by (\ref{E:TCA_Unsupervised}) by setting $\mytensor{G}  = \mytensor{X} \times_1 {\mymatrix{U}^{(1)}} \times_2 {\mymatrix{U}^{(2)}} \times_3 \cdots \times_{N-1} {\mymatrix{U}^{(N-1)}} \times_N \mymatrix{I}$, requiring ${\mymatrix{U}^{(N-1)}} = \mymatrix{I}$,
and enforcing all the $N-1$ projection matrices to be column orthonormal. That is, 
$\{ \mathcal{C}_n = {\mymatrix{U}^{(n)}}\myT \mymatrix{U}^{(n)} = \mymatrix{I} \}_{n=1}^{N-1}$. Furthermore, by requiring  factor matrices to be orthogonal and nonnegative (\ref{E:TCA_Unsupervised})  yields the non-negative MPCA~\cite{panagakis2010nonnegative}.
By neglecting orthogonality constraints and requiring factor matrices to be only nonnegative, i.e., 
$\{ \mathcal{C}_n = \mymatrix{U}^{(n)} \geq \mymatrix{0}\}_{n=1}^{N-1}$ the NTF~\cite{shashua2005non}
is obtained by  (\ref{E:TCA_Unsupervised}) when $\mytensor{G}$ is  a superdiagonal tensor, i.e., $\mytensor{G} = \mytensor{I}$. 

Supervised tensor component analysis exploits available side information such as class labels or neighborhood relationships among samples to estimate a multilinear projection. Such models can be unified under the graph embedding framework \cite{yan2007graph_embedding} by solving:
\begin{equation}
\begin{aligned}
\min_{\{ \mymatrix{U}^{(n)} \}_{n=1}^{N-1}}  & \sum_{m,k \atop m \neq k }^M \mymatrix{W}_{m,k} \| \mytensor{X}_{m} \prod_{n=1}^{N-1} \times_{n} \mymatrix{U}^{(n)} -\mytensor{X}_{k} \prod_{n=1}^{N-1} \times_{n} \mymatrix{U}^{(n)} \|^{2} ,  \\
& \mbox{s.t} \quad \sum_{m,k}^M \mymatrix{B}_{m, k}\|\mytensor{X}_{m} \prod_{n=1}^{N-1} \times_{n} \mymatrix{U}^{(n)} \|^{2}  = 1,
\end{aligned}
\end{equation}
where $\mymatrix{W}$ denotes the affinity matrix capturing some sort of within-class neighborhood relationship among tensor samples $\mytensor{X}_m$
and $\mymatrix{B}$ is an affinity matrix representing
between-class neighborhood relationship. Multilinear extensions to  Linear Discriminant Analysis (LDA), such as \cite{yan2006multilinear,tao2007general,li2014multilinear}
for example, are obtained by setting 
$ \mymatrix{W}_{m k}=\delta_{c_{m}, c_{k}} / n_{c_{m}}$ and $ \mymatrix{B} = \mymatrix{I}- \frac{1}{M} \myvector{e} \myvector{e}\myT$, where
$\delta$ denotes the Kronecker delta function, $\myvector{e}$ is the standard basis and $n_c$ denotes the number of the samples belonging to the $c\myth$ class.
\subsection{Tensor-structured dictionary learning}
\label{sec:replearning:diclearn}

Sparse coding (or dictionary learning) methods  are central in numerous visual information processing tasks. In this context, a data matrix $\mymatrix{X} \in \myR^{I \times M}$ 
contains in its columns vectorized image patches of size $\sqrt{I}\times\sqrt{I}$,
rather than data samples. Assuming $\mymatrix{D}$ is over-complete and requiring $\mymatrix{R} =[\mymatrix{r}_1, \mymatrix{r}_2, \ldots,\mymatrix{r}_M]$ to be sparse, sparse dictionary learning solves the following, non-convex, optimization problem:

\begin{equation}\label{E:DL}
    \min_{\mymatrix{R},\mymatrix{D}} \Vert \mymatrix{X}-\mymatrix{D}\mymatrix{R}\Vert_{F}^2 + \lambda \sum_{m=1}^M\Vert \mymatrix{r}_m \Vert_0,
\end{equation}

However, extracting patches from images and flattening them into vectors makes (\ref{E:DL}) suboptimal
for multidimensional tensor data. Besides structure loss, typical solvers for (\ref{E:DL}),
such as the K-SVD family of algorithms \cite{zhang2015sparseRepSurvey}, suffer from a high computational burden, preventing their applicability to massive amounts of tensor data.

To preserve multilinear structures in dictionary learning, a separable structure on the dictionary can be enforced. For instance, Separable Dictionary Learning (SeDiL) \cite{Hawe_2013_Separable}
considers a set of samples in matrix form, namely, $\{ \mymatrix{X}_m\}_{m=1}^M$, admitting sparse representations on a pair of bases
$\mymatrix{U}^{(1)}, \mymatrix{U}^{(2)}$, of which the Kronecker product constructs the dictionary. More specifically, SeDiL solves:
\begin{align}
\min_{\mymatrix{U}^{(1)}, \mymatrix{U}^{(2)}, \mytensor{R} } &\frac{1}{2} \sum_{m=1}^{M}\Vert \mymatrix{X}_{m}-\mymatrix{U}^{(1)} \mymatrix{R}_{m} {\mymatrix{U}^{(2)}}\myT \Vert_{F}^{2} \nonumber \\
&+\lambda_1 g(\mytensor{R }) 
+\lambda_2 r(\mymatrix{U}^{(1)})+\lambda_3 r(\mymatrix{U}^{(2)}),
\end{align}
where the regularizers $g(\cdot)$ and $r(\cdot)$ promote sparsity in the representations, and low mutual-coherence of the dictionary $\mymatrix{D} = \mymatrix{U}^{(2)} \otimes \mymatrix{U}^{(1)}$, respectively. Here, $\mymatrix{D} $ is constrained to have orthogonal columns. 

A different approach is taken in \cite{hsieh2014twodsparse}, where
a separable 2D dictionary is learned following a two-step strategy similar to that of the K-SVD. Each observation matrix $\mymatrix{X}_m$ is approximated by 
$ \mymatrix{U}^{(1)} \mymatrix{R}_m {\mymatrix{U}^{(2)}}\myT $,
where $\mymatrix{R}_m$ are slices of a core tensor $\mytensor{R}$.
In the first step, the sparse representations $\mymatrix{R}_m$ 
are obtained by 2D Orthogonal Matching Pursuit (OMP)~\cite{fang2011matchingpursuit}.
In the second step, a CP decomposition is performed on a tensor of residuals. 

Theoretical analysis suggests that the sample complexity of tensor-structured dictionary learning  can be significantly lower than that for unstructured, vector data; see \cite{shakeri2016minimax}
for 2D data, and \cite{shakeri2018minimaxtensor}
for $N$-order tensor data. This suggests better performance is achievable with separable dictionary learning from tensor data compared to vector-based dictionary learning methods.

However, the aforementioned models learn  dictionaries on many small image patches whose number
can easily become prohibitively large, undermining the scalability benefits of learning separable dictionaries. Moreover, it is worth noting that none of the models mentioned above are robust to gross errors due to the  loss function used. Such limitations are alliviated by the Robust Kronecker Component Analysis (RKCA) \cite{bahri2017robustKDC,bahri2019robustKC}.

The RKCA combines ideas from separable dictionary learning and robust tensor decomposition. Instead of seeking overcompleteness, low-rank is promoted on the pair of dictionaries, and sparsity is imposed in the code yielding a sparse representation of the input tensor samples.
Specifically, RKCA solves:
\begin{align}\label{E:KC}
\min _{\mymatrix{U}^{(1)}, \mymatrix{U}^{(2)}, \mytensor{R}, \mytensor{S} } \sum_{m=1}^M \Vert \mymatrix{X}_{m}-\mymatrix{U}^{(1)} \mymatrix{R}_{m} {\mymatrix{U}^{(2)}}\myT-\mymatrix{S}_{m} \Vert_F^2 \nonumber \\
+\lambda \sum_{m=1}^M \Vert \mymatrix{R}_{m}  \Vert_{1}+\lambda \sum_{m=1}^M \Vert \mymatrix{S}_{m}\Vert_{1}+ \Vert \mymatrix{U}^{(2)} \otimes \mymatrix{U}^{(1)}  \Vert_{F},
\end{align}
where matrices $\{ \mymatrix{S}_m \}_{m=1}^M$ account for sparse noise and outliers.
Equivalently, enforcing the equality constraints and concatenating
the matrices $\mymatrix{X}_m$, $\mymatrix{R}_m$, and $\mymatrix{S}_m$  as the frontal slices of
a third-order tensor, (\ref{E:KC}) is written as:

\begin{align}\label{E:RKCA_T}
\min_{\mymatrix{U}^{(1)}, \mymatrix{U}^{(2)}, \mytensor{R}, \mytensor{S}  } & \alpha\|\boldsymbol{\mathcal { R }}\|_{1}+\lambda\|\mathcal{S}\|_{1}+\|\mymatrix{U}^{(2)} \otimes \mymatrix{U}^{(1)} \|_{\mathrm{F}} \nonumber \\
\text { s.t } & \mathcal{X}=\mathcal{R} \times_{1} \mymatrix{U}^{(1)}  \times_{2} \mymatrix{U}^{(2)}+\mathcal{S}.
\end{align}
Consequently, (\ref{E:RKCA_T}) indicate that separable dictionary can be expressed as a regularized tensor decomposition. As opposed to classic dictionary learning models, it is worth mentioning that following \cite{haeffele2014structured,haeffele2015global}, the global optimality of RKCA is ensured.
\subsection{Tensor regression}\label{ssec:tensor-regression}

Besides learning representations, low-rank tensor decomposition facilitates reducing the parameters of tensor regression models and prevents overfitting. Tensor regression models generalize linear regression to higher-order tensors. More specifically, in tensor regression a regression output \(y \in \myR\) is expressed as the inner product between the observation tensor \(\mytensor{X}\) and a weight tensor \(\mytensor{W}\) with the same dimension as \(\mytensor{X}\): \(y = \myinner{\mytensor{X}}{\mytensor{W}} + b\). Several low-rank tensor regression models have been proposed \cite{guo2012tensor,rabusseau2016low,zhou2013tensor,batmanghelich2011regularized,yu2016learning,song2017multilinear}. These methods share the assumption that the unknown regression weight tensor is low-rank. This is enforced by expressing the weights tensor in CP
\cite{guo2012tensor, zhou2013tensor}  or Tucker form \cite{paredes2013multilinear,yu2016learning,li_tucker_2018}.
 Alternatively, low-rank constraints on the weight tensor are enforced by spectral regularization, such as applying the nuclear norm on the unfoldings of the weight tensor along its modes \cite{paredes2013multilinear}, in a similar way to robust tensor decomposition described in Section \ref{sec:replearning:robust}. To further prevent overfitting, regularization is incorporated into low-rank tensor regression models. For instance, \cite{batmanghelich2011regularized,guo2012tensor, rabusseau2016low} employ tensor norm and hence extend the low-rank ridge regression to high-order tensors. $\ell_1$-norm and elastic-net regularization is adopted in \cite{zhou2013tensor}, while a combination of $\ell_1$-norm and nuclear norm regularizes 
 the weight tensor in \cite{song2017multilinear} forcing it to be simultaneously low-rank and sparse. It is worth mentioning that low-rank regression models extend support vector machines to higher-order tensors, e.g., \cite{guo2012tensor, song2017multilinear}.
 
 Furthermore, higher-order generalizations of Partial-Least Squares were proposed~\cite{hopls_nips_zhao2011,hopls_zhao2012}. The idea of Higher Order Partial Least Squares (HO-PSL) is to find a latent correlated subspace between the observation tensor and the response tensors. Alternatively, it is possible to perform regression in a subspace obtained with Tensor PCA~\cite{wu2006regression}. Finally, quantile regression has also recently been generalized to higher-order~\cite{lu2020high}, using a Tucker structure on the regression weights.


\section{Tensor Methods in Deep Neural Networks Architectures}
\label{sec:deeparch}
Tensors play a central role in modern deep learning architectures. The basic building blocks of deep neural networks, such as multichannel convolutional kernels and attention blocks, are essentially tensor mappings, represented by  tensors. Deep neural networks, as mentioned earlier, operate in a regime where the number of parameters is typically tens of millions or even billions, and is thus much larger than the number of training data. In this over-parameterized regime,  incorporating tensor decompositions into deep learning architectures leads to a significant reduction in the number of unknown parameters. This allows the design of networks that further preserve the topological structure in the data, being more parsimonious in terms of parameters, more data-efficient, and more robust to various types of noise and domain shifts. Besides the design of deep networks, tensor analysis can be beneficial in demystifying the success behind neural networks, having already been used to prove universal approximation properties of neural networks~\cite{cohen2016convolutional}, or to understand the inductive bias leveraged by networks in computer vision~\cite{levine2017deep}. This section overviews recent developments in the use of tensor methods to design efficient and robust deep neural networks and the theoretical understanding of their properties. 

\subsection{Parameterizing fully-connected layers}
\notebook{tt-compression}

Leveraging tensor decompositions within neural networks by re-parametrizing fully-connected layers has been first proposed in \cite{novikov2015tensorizing}.  Since the weights of fully-connected layers are represented as matrices, tensor decompositions cannot be applied directly. Therefore, the weight matrix needs to be reshaped to obtain a higher-order tensor.

Specifically, consider an input matrix 
$\mathbf{W}$ of size \(I \times J\), whose dimensions can be expressed as
\(I = I_1 \times I_2 \times \cdots \times I_N\) 
and 
\(J = J_1 \times J_2 \times \cdots \times J_N\). Hence,
$\mathbf{W}$ is tensorized first by reshaping it to a higher-order tensor of size 
$I_1 \times I_2 \times \cdots \times I_N \times J_1 \times J_2 \times \cdots \times J_N$. Next, 
by permuting the dimensions and reshaping it again,
an $N$\myth~order tensor of $I_1 J_1 \times I_2 J_2 \times \cdots \times I_N J_N$ is obtained. This tensor 
is then compressed in the TT format.
The approximated weights are reconstructed from that low-rank TT factorization during inference, and are then reshaped back into a matrix.

In this setting, other types of decompositions have also been explored, including Tensor-Ring~\cite{zhao2018learning}
and Block-Term Decomposition~\cite{block_term_nets}.
This strategy has also been extended to parametrize other types of layers~\cite{garipov2016ultimate}.

In the context of two-layer fully-connected neural networks, tensors have also been used for deriving alternative to (stochastic) gradient descent methods for  training, with guaranteed generalization bounds. Specifically, in~\cite{janzamin2015beating},  the CP decomposition is applied on the cross-moment tensor, which encodes the correlation between the third-order score function (i.e., the normalized derivative of the input pdf) to obtain the parameters of the first layer. The parameters of the second layer are obtained next via regression.  Under mild assumptions, this approach comes with guaranteed risk bounds in both the realizable setting (i.e., the target function can be approximated with zero error from the neural network under consideration) and the non-realizable one, where the target function is arbitrary.

\subsection{Tensor contraction layers}

In modern deep neural networks such as CNNs, each layer's output is an (activation) tensor. However, activation tensors are usually flattened and passed to subsequent fully-connected layers resulting in structural information loss.
A natural way to preserve the multilinear structure
of activation tensors is to incorporate 
tensor operations into deep networks.
For instance, tensor contraction can be applied to an activation tensor to obtain a low-dimensional representation of it~\cite{tcl}. In other words, an input activation tensor is contracted along each mode with a projection matrix (an \emph{n-mode product} as defined earlier). 

\begin{figure}[ht]
  \centering
  \includegraphics[width=0.8\linewidth]{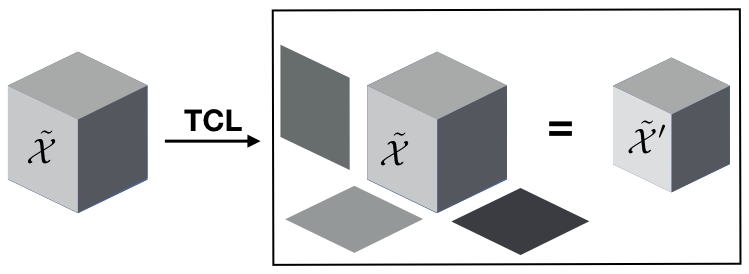}
  \caption{An input tensor \(\mytensor{X}\) is contracted into a smaller tensor \(\mytensor{X'}\) through a Tensor Contraction Layer~\cite{tcl}.}\label{TCL}
\end{figure}

Let us assume a network involves an input activation tensor \(\mytensor{X}\) of size \(  S \times  I_1 \times \cdots \times I_N  \), where \(S\) is the batch size.
A Tensor Contraction Layer (TCL) contracts the input activation  tensor, along each of its dimensions except the batch-size, with a series of factor matrices \(\{\mymatrix{V}^{(n)}\}_{n=2}^{N}\).The result is a compact core tensor \(\mytensor{X'}\) of
smaller size \( S \times R_1 \times \cdots \times R_N  \) defined as
\begin{equation}
\mytensor{X'} = 
\mytensor{X} \times_2 \mymatrix{V}^{(2)} 
		  \times_3  \mymatrix{V}^{(3)} \times
		  \cdots
          \times_{N} \mymatrix{V}^{(N)},
\end{equation}
with \(\mymatrix{V}^{(n)} \in \myR^{R_n \times I_n}, n \in \myrange{2}{N}\). 
Note that the projections start at the second mode as we do not contract along the first mode \(S\), which corresponds to the mini-batch size. These factors are learned in an end-to-end manner, concurrently with all the other network parameters by gradient backpropagation. A TCL that produces a compact core of smaller size is denoted as \emph{size--\(\left(R_2, \cdots, R_N\right)\) TCL}, or \emph{TCL--\(\left(R_2, \cdots, R_N\right)\)}.\\

In addition to preserving the multi-linear structure in the input, 
TCLs have much less parameters than a corresponding fully-connected layer. 
We can see this by establishing the link between a fully-connected layer with structured weights and a tensor regression layer.
Let us assume an input activation tensor \(\mytensor{X}\) of size \(  S_1 \times  I_1 \times \cdots \times  I_N\).  A size--\( \left(R_2, \cdots, R_N \right) \) TCL  
parameterized by weight factors \(\mymatrix{V}^{(2)}, \cdots, \mymatrix{V}^{(N)}\) has a total of \( \sum_{n=2}^{N} I_n \times R_n \) parameters. By unfolding the input tensor (i.e., vectorizing each sample of the batch), TCL can be equivalently expressed as a fully-connected layer parametrized by the weight matrix \( \mymatrix{W} = \left(\mymatrix{V}^{(2)}  \otimes \cdots \otimes \mymatrix{V}^{(N)} \right)\myT \), which computes
\[
    \mytensor{X}_{[1]} \mymatrix{W} =  \mytensor{X}_{[1]}  \left(\mymatrix{V}^{(2)}  \otimes \cdots \otimes \mymatrix{V}^{(N)} \right)\myT.
\]

However, since fully-connected layers have no structure on their weights, the corresponding fully-connected layer parametrized by a weight matrix \( \mymatrix{W} \) of the same size as above would have a total of \( \prod_{n=2}^{N} I_n \times R_n \) parameters, compared to the  \( \sum_{n=2}^{N} I_n \times R_n \) parameters of the TCL.

\subsection{Tensor regression layers}

\notebook{tensor\_regression\_layer}

In CNNs, after an activation tensor has been obtained through a series of convolutional layers, 
predictions (outputs) are typically generated 
by flattening this activation tensor or 
applying a spatial pooling, 
before using a fully-connected output layer.
However, both approaches discard the multilinear structure of the activation tensor. To alleviate this, tensor regression models
can be incorporated within deep neural networks and trained end-to-end jointly with the other parameters of the network~\cite{trl}.

\begin{figure*}[ht]
    \centering
    \includegraphics[width=0.9\linewidth]{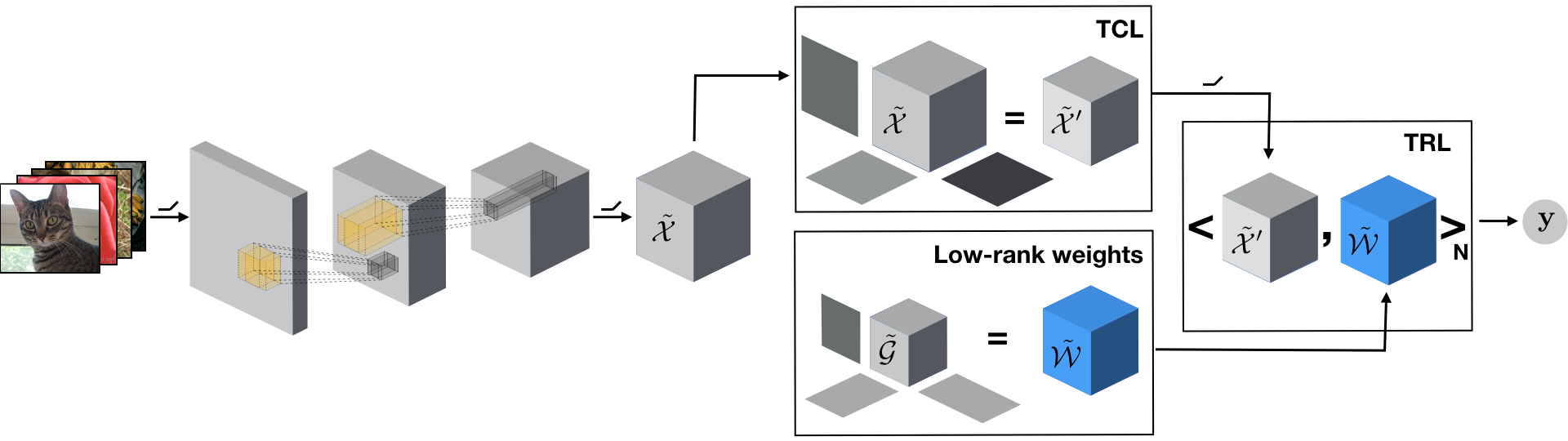}
    \caption{Illustration of a convolutional deep network where the activation tensor is processed through a TCL followed by a TRL to produce the outputs while preserving the topological structure~\cite{trl}.
   Note that \(y\), here, is a scalar.}
\end{figure*}

\looseness-1Specifically, let us assume we have a set of \(S\) input activation tensors
\(\mytensor{X}^{(s)} \in \myR^{I_1 \times I_2 \times \cdots \times I_{N}} \)
and corresponding target labels \( y^{(s)} \in \myR \), with \(s \in \myrange{1}{S}\). 
A Tensor Regression Layer (TRL) estimates the regression weight tensor 
\( \mytensor{W} \in 
   \myR^{I_1 \times I_2 \cdots \times I_N},
\)
by further assuming that it admits a low-rank decomposition. Originally, TRL has been introduced by assuming a Tucker structure on the regression weights. That is, for a rank--\(\left(R_1, \cdots, R_N\right)\) Tucker decomposition, we have:
\begin{align}\label{eq:trl}
    y^{(s)} & = \myinner{\mytensor{X}^{(s)}}{\mytensor{W}} + b \nonumber \\
    \text{with } & 
         \mytensor{W} 
         =  
         \mytensor{G} 
         \times_1 \mymatrix{U}^{(1)}
         \times_2 \mymatrix{U}^{(2)}
         \cdots
         \times_N \mymatrix{U}^{(N)}
\end{align}
with \(\mytensor{G} \in \myR^{R_1 \times \cdots \times R_N} \), 
\(\mymatrix{U}^{(n)} \in \myR^{I_n \times R_n}\) 
for each 
\(n\) in \(\myrange{1}{N}\)
and \(\mymatrix{U}^{(N)} \in \myR^{1 \times R_{N}}\).
For simplicity, here, we introduce the problem for a scalar output, but it is established, in general, for a tensor-valued response. The inner product between the activation tensor and the regression weight is then replaced by a tensor contraction along matching modes.

In addition to the implicit regularization induced by the low-rank structure of the regression weights, TRLs result in a much smaller number of parameters. For an input activation tensor
\(\mytensor{X} \in \myR^{S \times I_1 \times I_2 \times \cdots \times I_N} \),
a rank-\((R_1, R_2, \cdots, R_N, R_{N+1})\) TRL
and an \(d\)-dimensional output.
A fully connected layer taking the same input \(\mytensor{X}\) (after a flattening layer) would have \(n_{\text{FC}}\) parameters, with
\(
n_{\text{FC}} = d \times \prod_{n=1}^N I_n.
\)
By comparison, a rank-\((R_1, R_2, \cdots, R_N, R_{N+1})\) TRL taking \(\mytensor{X}\) as input has a number of parameters \(n_{\text{TRL}}\), with:
\begin{equation}\nonumber
n_{\text{TRL}} =  \prod_{n=1}^{N+1} R_n + \sum_{n=1}^N R_n \times I_n + R_{N+1} \times d.
\end{equation}
As previously observed for the TCL, TRL is much more parsimonious in terms of the number of parameters.
Note that, while we introduced the TRL here with a Tucker structure on the regression weight tensor, in general, any low-rank format can be used, including CP and TT~\cite{cao2017tensor}.

\subsection{Parametrizing convolutional layers}
\label{subsec:parametrizingconvl}

The weights of convolutional layers are naturally represented as tensors. For instance, a 2D convolution is represented by a 4\myth--order tensor. Hence convolutional layers can directly be compressed by employing tensor decomposition.  In fact, there is a close link between deep neural networks and efficient convolutional blocks, which we will highlight in this section.

\subsubsection{\(1 \times 1\) convolutions}
The first thing to notice is that \(1 \times 1\) convolutions are tensor contractions. This type of convolution is frequently used within deep neural networks in order to create a data bottleneck.
For a \(1\times1\) convolution \(\Phi\), defined by kernel \(\mytensor{W}  \in \myR^{T \times C \times 1 \times 1}\) and applied to an activation tensor \(\mytensor{X} \in \myR^{C \times H \times W}\). We denote the squeezed version of \(\mytensor{W}\) along the first mode as \(\mymatrix{W} \in \myR^{T \times C}\). We then have:
\begin{equation}
    \label{eq:n-mode-1x1}\nonumber
    \Phi(\mytensor{X})_{t, y, x} =
    \mytensor{X} \myconv \mytensor{W} = 
    \sum_{k=1}^C  \mytensor{W}_{t, k, y, x} \mytensor{X}_{k, y, x}
    = \mytensor{X} \times_1 \mymatrix{W}.
\end{equation}
Note that, here, we have \(x=y=1\) as we are considering a \(1 \times 1\) convolution. 

\subsubsection{Kruskal convolutions}
Let us now focus on how tensor decomposition can be applied to the convolutional kernels of CNNs and how this results not only in fewer parameters but also faster operations. The CP decomposition, for instance, allows separating the modes of a convolutional kernel, resulting in a Kruskal form. Albeit the application of tensor decomposition in the context of CNNs is relatively new and  of practical interest nowadays, the idea of separable representation of linear operators by means of tensor decomposition is not new (see \cite{8187112}) and in fact was one of the original motivations for tensor rank decomposition of the matrix-multiplication tensor (see \cite{GCP} for example).

In computer vision, the use of separable convolutions was proposed  in~\cite{rigamonti2013learning}, in the context of learnable filter banks. In the context of deep learning, \cite{jaderberg2014speeding} proposed to leverage redundancies across channels by exploiting separable convolutions. Furthermore, 
it is possible to start from a pre-trained convolutional kernel and apply CP decomposition to it in order to obtain a separable convolution.
This was proposed for 2D convolutions in~\cite{lebedev2015speeding}, where a CP decomposition of the kernel was achieved by minimizing the reconstruction error between the pre-trained weights and the corresponding CP approximation. The authors demonstrated both space savings and computational speedups. 
Similar results can be obtained using different optimization strategies such as the tensor power method~\cite{astrid2017cp}.

\begin{figure}[!h]
  \centering
  \includegraphics[width=1\linewidth,trim={0 150 0 150},clip]{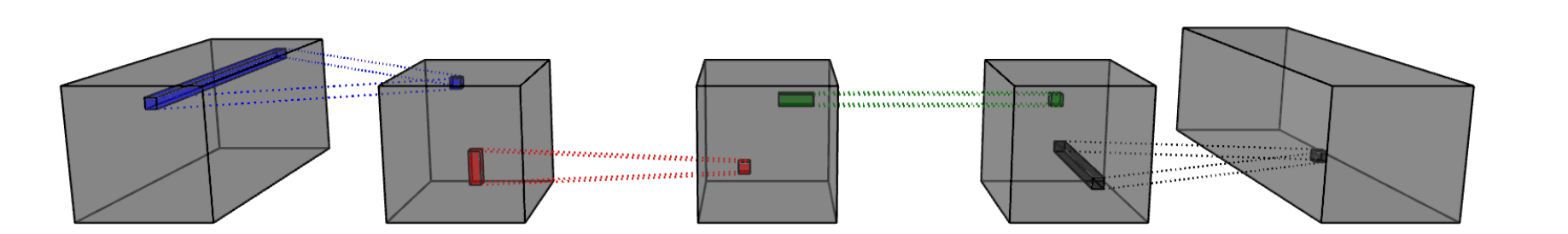}
  \caption{\textbf{Illustration of a 2D Kruskal convolution}, as expressed \cref{eq:kruskal-conv}, with matching colours. First, a $1\times 1$ convolution reduces the number of input channels to the rank (blue). Two depthwise convolutions are then applied on the height and width of the activation tensor (red and green). Finally, a second $1\times 1$ convolution restores the number of channels from the rank of the CP decomposition to the desired number of output channels (black).
  }
  \label{fig:kruskal-conv}
\end{figure}

An advantage of factorizing the kernel using a CP decomposition is that the resulting factors can be used to parametrize a series of efficient depthwise separable convolutions, replacing the original convolution~\cite{lebedev2015speeding}, as illustrated in Fig.~\ref{fig:kruskal-conv}. This can be seen by considering the expression of a regular convolution:
\begin{equation}
    \label{eq:convolution}
    \mytensor{F}_{t, y, x} = 
    \myblue{\sum_{k=1}^C}
    \myred{\sum_{j=1}^H}
    \mygreen{\sum_{i=1}^W}
    \mytensor{W}(t, \myblue{k}, \myred{j}, \mygreen{i})
    \mytensor{X}(\myblue{k}, \myred{j + y}, \mygreen{i + x}).
\end{equation}

A Kruskal convolution is obtained by expanding the kernel \(\mytensor{W}\) in the CP form as introduced in Section~\ref{sssec:CP}. By reordering the terms, we can then obtain an efficient reparametrization~\cite{lebedev2015speeding}:
\begin{equation}
    \label{eq:kruskal-conv}
    \mytensor{F}_{t, y, x} = 
    \underbrace{\sum_{r=1}^{R} \mymatrix{U}^{(T)}_{t, r}
    \underbrace{\left[\mygreen{\sum_{i=1}^W} \mymatrix{U}^{(W)}_{\mygreen{i}, r}
    \underbrace{\left(\myred{\sum_{j=1}^H} \mymatrix{U}^{(H)}_{\myred{j}, r}
    \underbrace{\left[\myblue{\sum_{k=1}^C}  \mymatrix{U}^{(C)}_{\myblue{k}, r}\mytensor{X}(\myblue{k}, \myred{j + y}, \mygreen{i + x})
     \right]}_{\myblue{1\times1 \text{ conv}}}
     \right)}_{\myred{\text{ depthwise conv}}}
    \right]}_{\mygreen{\text{ depthwise conv}}} 
    }_{1\times 1 \text{ convolution}}
\end{equation}

\subsubsection{Tucker convolutions}\label{sssec:tensorize_tucker}
As previously, we consider the convolution \(\mytensor{F} = \mytensor{X} \myconv \mytensor{W}\) described in (\ref{eq:convolution}). However, instead of a Kruskal structure,  the convolution kernel \(\mytensor{W}\) is assumed to admit a Tucker decomposition as follows:
\begin{equation}
    \label{eq:tucker-kernel}\nonumber
    \mytensor{W}(t, s, j, i) =
    \sum_{r_1=0}^{R_1}
    \sum_{r_2=0}^{R_2}
    \sum_{r_3=0}^{R_3}
    \sum_{r_4=0}^{R_4}
        \mytensor{G}_{r_1, r_2, r_3, r_4}
        \mymatrix{U}^{(T)}_{t, r_1} 
        \mymatrix{U}^{(C)}_{s, r_2} 
        \mymatrix{U}^{(H)}_{j, r_3} 
        \mymatrix{U}^{(W)}_{i, r_4}.
\end{equation}

This approach allows for an efficient reformulation~\cite{yong2016compression}: first, the factors along the spatial dimensions are absorbed into the core by writing
$
    \mytensor{H} = \mytensor{G} 
    \times_3 \mymatrix{U}^{(H)}_{j, r_3}
    \times_4 \mymatrix{U}^{(W)}_{i, r_4}.
$
By reordering the term, we can now see that a Tucker convolution is equivalent to first transforming the number of channels, then applying a (small) convolution before restoring the channel dimension from the rank to the target number of channels:
\begin{gather}\label{eq:tucker-conv}
    \mytensor{F}_{t, y, x} =
    \underbrace{
        \mygreen{\sum_{r_1=1}^{R_1}}
        \mymatrix{U}^{(T)}_{t, \mygreen{r_1}}
        \left[
            \underbrace{
                \myred{
                    \sum_{j=1}^H
                    \sum_{i=1}^W
                    \sum_{r_2=1}^{R_2}
                }
                \mytensor{H}_{\mygreen{r_1}, \myred{r_2}, \myred{j}, \myred{i}}
                \left[
                    \underbrace{
                        \myblue{\sum_{k=1}^C}
                        \mymatrix{U}^{(C)}_{\myblue{k}, \myred{r_2}}
                        \mytensor{X}(\myblue{k}, \myred{j + y}, \myred{i + x})
                    }_{\myblue{1 \times 1 \text{ conv}}}
                \right]
            }_{\myred{H \times W \text{ conv}}}
        \right]
    }_{\mygreen{1 \times 1 \text{ conv}}}.
\end{gather}

\begin{figure}[!h]
  \centering
  \includegraphics[width=1\linewidth,trim={0 130 0 150},clip]{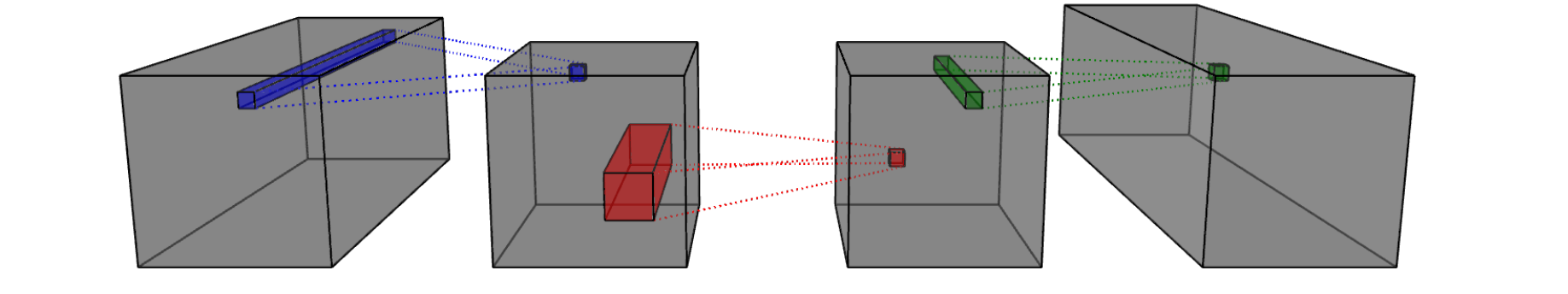}
  \caption{\textbf{Illustration of a Tucker convolution} expressed as a series of small, efficient convolutions, as illustrated in \cref{eq:tucker-conv}. Note that this is the approach taken by ResNet for the Bottleneck blocks. After the full kernel has been decomposed, the factors along the input and output channels are used to parametrize $1 \times 1$ convolutions, applied respectively first (blue) and last (green). The remaining two factors are absorbed into the core and used to parametrize a regular 2D convolution (red).}
  \label{fig:tucker-conv}
\end{figure}

Besides the aforementioned types of convolutions (i.e., bottleneck layers, separable convolutions), \cite{hayashi2019exploring} identify maximally resource-efficient convolutional layers utilizing tensor networks and decompositions. Tensor decompositions have also been used for unifying different successful architectures. Particularly, in \cite{chen2017sharing}, the block term decomposition
\cite{deLathhauwer2008uniqueness}
is used to unify different architectures that employ
residual connections, such as ResNet~\cite{he2016deep},  while proposing a new architecture that improves parameter efficiency.
This implies that searching over the space of tensor decompositions and their parameters might be an excellent proxy task to limit the search space for architecture search.

From a theoretical point of view, by using the CP decomposition to replace convolutional with low-rank layers results in deriving
the upper bound for the generalization error of the compressed network \cite{li2020understanding}. This bound improves previous results
on the compressibility and generalizability of neural networks \cite{arora2019implicit}. It is also related to the layers' rank, providing insights on designing neural networks (with low-rank tensors as layers).

\subsubsection{Multi-stage compression}
The factorization strategies mentioned above allow for efficient compression of pre-trained fully-connected and convolutional layers of deep neural networks by applying tensor decomposition to the learned weights. 
Fine-tuning allows recovery for loss of performance when increasing the compression ratio. However, it is also possible to apply multi-state compression~\cite{Gusak_2019_ICCV} to further compress without loss of performance. Instead of compressing then fine-tuning, the idea is to adopt an iterative approach by alternating low-rank factorization with rank selection and fine-tuning.

\subsubsection{General N-D separable convolutions and transduction}\label{sssec:separable}

Design of general (N-dimensional) fully separable, convolution has been introduced in the context of deep neural networks in \cite{kossaifi2020factorized}. 
Let us assume an \((N + 1)\myth\)--order input activation tensor \(\mytensor{X} \in \myR^{C \times D_1 \times \cdots \times D_{N}}\) corresponding to \(N\) dimensions with \(C\) channels.
The higher-order separable convolution, which we denote \(\Phi\), is defined by a kernel \(\mytensor{W} \in \myR^{T \times C \times K_1 \times \cdots \times K_{N}}\). 
By expressing this kernel in a factorized Kruskal form as \(\mytensor{W} = \mykruskal{\myvector{\lambda}; \,\,\mymatrix{U}^{(T)}, \mymatrix{U}^{(C)}, \mymatrix{U}^{(K_1)}, \cdots, \mymatrix{U}^{(K_{N})}}\), we obtain:
\begin{equation}\nonumber
    \begin{split}
    \Phi (\mytensor{X})_{t, i_1, \cdots, i_{N}}  & = 
    \sum_{r = 1}^{R}
    \sum_{s = 1}^C
     \sum_{i_1 = 1}^{K_1}
    \cdots
    \\ 
    &
    \cdots \sum_{i_{N} = 1}^{K_{N}} \lambda_r \bigl[\mymatrix{U}^{(T)}_{t, r} \mymatrix{U}^{(C)}_{s, r} 
    \mymatrix{U}^{(K_1)}_{i_1, r} 
    \cdots
    \mymatrix{U}^{(K_{N})}_{i_{N}, r} 
    \mytensor{X}_{s, i_1, \cdots, i_{N}} \bigr].
    \end{split}
\end{equation}

The separable higher-order convolution can be obtained by reordering the term into:
\begin{equation}
    \label{eq:ho-conv-2}
    \mytensor{F} = 
    \left(
        \rho \left( \mytensor{X} \times_1 \mymatrix{U}^{(T)} \right)
    \right) \times_1 \left( \mydiag(\myvector{\lambda}) \mymatrix{U}^{(C)} \right),
\end{equation}
where $\rho$ applies the 1D spatial convolutions:
\begin{equation}
    \label{eq:rho}
    \rho(\mytensor{X}) = 
    \left(
        \mytensor{X}
        \myconv_1 \mymatrix{U}^{(K_1)}
        \myconv_2 \mymatrix{U}^{(K_2)}
        \myconv \cdots
        \myconv_{N} \mymatrix{U}^{(K_{N})}
    \right).
\end{equation}

A factorized separable convolution can be extended from $N$-dimensions to $N+K$ ($N, K \in \myN$) by means of \emph{transduction} \cite{kossaifi2020factorized}. Since the convolution is fully separable, it is possible to add additional factors. In particular, to apply transduction from \(N\) to \(N+1\), it is sufficient to add an additional factor \(\mymatrix{U}^{K_{N+1}}\) and update \(\rho\) in equation~(\ref{eq:rho}) to:
\begin{equation}\nonumber
    \rho(\mytensor{X}) = 
    \left(
        \mytensor{X}
        \myconv_1 \mymatrix{U}^{(K_1)}
        \myconv_2 \mymatrix{U}^{(K_2)}
        \myconv \cdots
        \myconv_{N} \mymatrix{U}^{(K_{N})}
        \myconv_{N+1} \mymatrix{U}^{(K_{N+1})}
    \right).
\end{equation}

\subsubsection{Parametrizing full networks}

So far, we reviewed methods focusing on parametrizing a single layer. It is also possible to jointly parametrize a whole convolutional neural network with a single low-rank tensor. This is the idea of ``T-Net''~\cite{kossaifi2019tnet}, where all the convolutional layers of a stacked-hourglass
\cite{newell2016StackedHN}
(a type of U-Net \cite{Ronneberger2015unet}) are parametrized by a single 8\myth--order tensor. Each of the modes of this tensor models a modality of the network, allowing to jointly model and regularize all the layers. Hence correlations across channels, layers, convolutional blocks, and subnetworks of the model can be leveraged. Imposing low-rank constraints on this model tensor results in large parameter space savings with little to no impact on performance, or even improved performance, depending on the compression ratio. That is, small compression ratios result in improved performance, while it is possible to reach substantial compression ratios with only minor decreases in performance.  In addition, it is possible to partially reconstruct the full tensor to obtain a Tucker factorization of the kernel of each layer, thus allowing for speedups as described in section~\ref{sssec:tensorize_tucker}.

\subsubsection{Preventing degeneracies in factorized convolutions}
As we discussed above, when parametrizing DCNNs, one can either train  the factorized layers  from scratch in an end-to-end fashion via stochastic gradient descent, or decompose existing convolutional kernels to obtain the factorized version. 
In the latter case, the authors of~\cite{phan2020stable} showed that when applying decomposition to obtain a factorized convolution, degeneracy can appear in the decomposition due to diverging components. They tackle this issue and enable stable factorization of convolutional layers through sensitivity minimization.

\subsection{Robustness of deep networks}
Deep neural networks produce remarkable results in computer vision applications but are typically very vulnerable to noise, such as capture noise, corruption, adversarial attacks of domain shift. Below, we provide a brief review of tensor methods that robustify deep networks.

\subsubsection{Tensor Dropout}
One known way to increase robustness is to incorporate randomization during training. For example, Dropout~\cite{srivastava2014dropout} randomly drops activations, while DropConnect~\cite{wan2013regularization} applies the randomization to the weights of fully-connected layers.
However, randomizing directly in the original space results in sparse weights or activations (with zero-ed values), which affects the high-order statistics.

\begin{figure}
    \centering
    \includegraphics[width=1\linewidth]{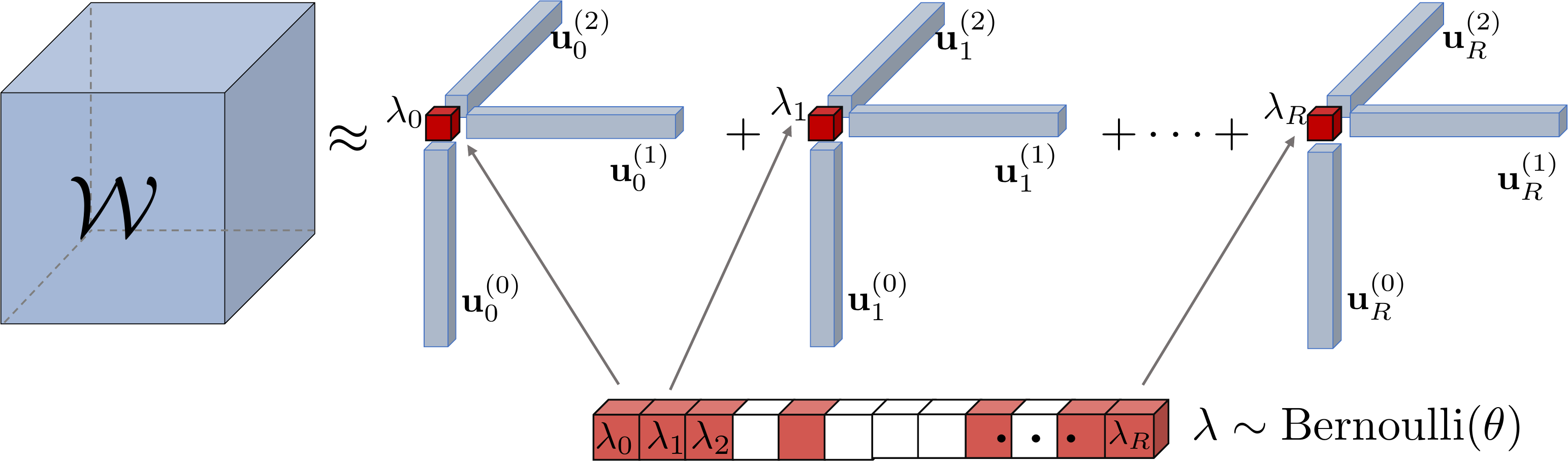}
    \caption{\textbf{Tensor Dropout on a CP decomposition.} Visualization of tensor dropout applied to a CP factorization. Here, a Bernouilli dropout is applied by first sampling a multivariate Bernoulli random vector $\myvector{\lambda}$. Its elements have value $1$ (\emph{red}) with probability $\theta$ and $0$ (\emph{white}) with probability $1 - \theta$. Each rank-$1$ CP component is kept or discarded according to the corresponding Bernoulli variable, thus inducing stochasticity on the rank of the decomposition.
    }
    \label{fig:overview-cp-dropout}
\end{figure}

\begin{figure}
    \centering
    \includegraphics[width=1\linewidth]{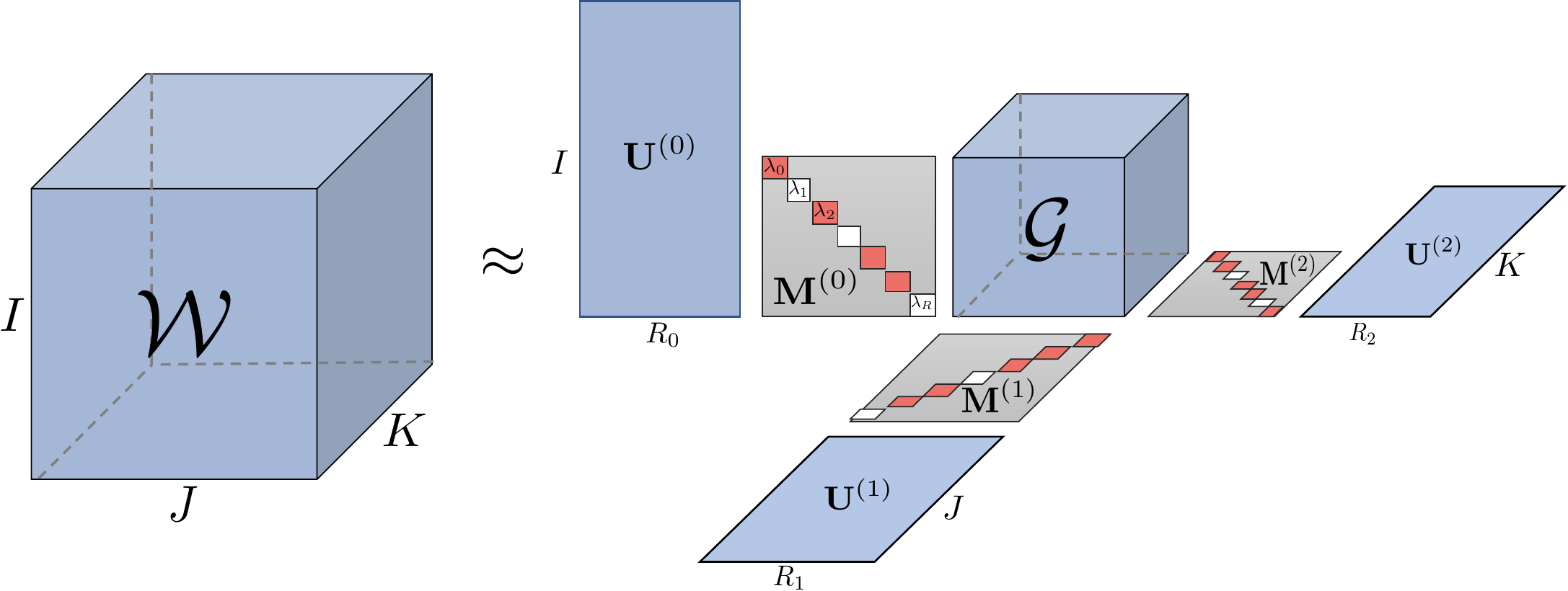}
    \caption{\textbf{Tensor Dropout on a Tucker decomposition.} Visualization of tensor dropout applied to a Tucker decomposition. Several types of randomization can be used. Here, a Bernouilli dropout is represented, applied in the latent subspace of the decomposition, by contracting the core along each mode with a series of (diagonal) Bernouilli sketching matrices, $\mathbf{M}^{(k)} = \mydiag({\myvector{\lambda}^{(k)}})$. }
    \label{fig:overview-tucker-dropout}
\end{figure}

Tensor dropout~\cite{kolbeinsson2020stochastically} is a principled way to incorporate randomization to robustify the model without altering its statistics. It consists of applying sketching matrices in the latent subspace spanned by a tensor decomposition, whether that's CP (see Fig.~\ref{fig:overview-cp-dropout}) or Tucker (see Fig.~\ref{fig:overview-tucker-dropout}), leading to significantly improved robustness to random noise (e.g., during capture) or adversarial attacks.

\subsubsection{Defensive Tensorization}
By combining the reparametrization approach described in section~\ref{sssec:separable} with tensor dropout, one can drastically improve the robustness of a DCNN to adversarial attacks. When combined with adversarial training, this approach proved successful against a wide range of attacks~\cite{bulat2020defensive}.

\subsubsection{Tensor-Shield}
In addition to the line of research that consists in modifying the weights or activations of the network to increase robustness to adversarial attacks, it is possible to modify the input imagee directly. For instance, Shield~\cite{das2018shield} uses a JPEG encoding to remove any potential adversarial noise. However, the defensive capacity of this approach is limited. Instead of a JPEG encoding, in \cite{entezari2020tensorshield} tensor decomposition techniques are utilized as a preprocessing step to find a low-rank approximation of images which can significantly discard high-frequency perturbations.

\subsection{Tensor structures in polynomial networks and attention mechanisms}
\label{ssec:review_polynomial_nets}

\notebooktiny{polynomial\_neural\_network\_astroid\_generation}

Apart from CNNs, tensor structures arise in
polynomial networks and attention mechanisms, capturing higher-order (e.g., multiplicative) interactions among data or parameters. Here, we provide an overview of how tensor methods are incorporated within such models.

\looseness-1 Multiplicative interactions and polynomial function approximation have long been used in the domain of machine learning. Group Method of Data Handling (GMDH)~\cite{ivakhnenko1971polynomial} proposes building quadratic polynomials from input elements. The elements in each quadratic polynomial are predetermined, which does not allow the method to capture data-driven correlations among different elements. The method was later extended in higher-order polynomials~\cite{oh2003polynomial}. Low-order polynomial networks were also used in pi-sigma~\cite{shin1991pi} and sigma-pi-sigma~\cite{li2003sigma}. The pi-sigma network includes a single hidden layer, and the output is the product of all hidden representations with fixed weights equal to $1$. The idea of sigma-pi-sigma is to sum multiple pi-sigma networks. In the context of kernel methods, Factorization Machines \cite{FM} and their higher-order extension \cite{HOFM} model both low- and higher-order interactions between variables, using factorized parameters via matrix/tensor decompositions. By employing factorized parameters, such models are 
able to estimate higher-order interactions even in tasks where data are sparse (e.g., recommender systems), unlike other kernel methods such as SVMs.

More recently, deeper neural networks have been proposed to capture multiplicative interactions \cite{srivastava2015highway, jayakumar2020Multiplicative}. 
For instance, Highway networks~\cite{srivastava2015highway} introduce an architecture that enables second-order interactions between the inputs. In \cite{dong2017more}, the authors augment the original convolution with a multiplicative interaction to increase representational power. In \cite{jayakumar2020Multiplicative}, the authors group several such multiplicative interactions as a second-order polynomial and prove that the second-order polynomials extend the class of functions that can be represented with zero error.  

Recent empirical~\cite{karras2018style, chrysos2020poly, chrysos2020naps} and theoretical~\cite{jayakumar2020Multiplicative} results indicate that multivariate higher-order polynomials expand the classes of functions that can be approximated with deep neural networks. Approximating a function using a multivariate higher-order polynomial involves learning the multivariate polynomial expansion parameters using input-output training data. Higher-order tensors naturally represent the parameters of such a multivariate polynomial \cite{sorber2014numerical}, whose dimensionality increases exponentially to the order of the polynomial approximation. This is another instance of the curse of dimensionality that tensor decompositions can mitigate. Moreover, in this setting, choosing different tensor decompositions results in different recursive equations that compose the polynomial function approximation and are hence linked to different neural network architectures.

\looseness-1Particularly, let $\binvar \in \realnum^d$ be the input, e.g., the latent noise in a generative model, or an image in a discriminative model, and let $\boutvar \in \realnum^o$ be the target, e.g., an output image or a class-label. A polynomial expansion\footnote{The Stone-Weierstrass theorem~\cite{stone1948generalized} guarantees that any smooth function can be approximated by a polynomial. Multivariate function approximation is covered by an extension of the Weierstrass theorem, e.g., in \cite{nikol2013analysis} (pg 19).} of the input $\binvar$ can be considered for approximating the target $\boutvar$. That is, a vector-valued function $\bm{G}(\bm{z}): \realnum^{d} \to \realnum^{o}$ expresses the high-order multivariate polynomial expansion:

\begin{equation}\label{eq:prodpoly_poly_general_eq}
    \boutvar = G(\binvar) = \sum_{n=1}^N \bigg(\bmcal{W}^{[n]} \prod_{j=2}^{n+1} \times_{j} \binvar\bigg) + \bm{\beta},
\end{equation}
where $\bm{\beta} \in \realnum^o$ and $\big\{\bmcal{W}^{[n]} \in  \realnum^{o\times \prod_{m=1}^{n}\times_m d}\big\}_{n=1}^N$ are the learnable parameters. The form of \eqref{eq:prodpoly_poly_general_eq} can approximate any smooth function (for large $N$). As aforementioned  the number of parameters required to accommodate all higher-order interactions of the input increases exponentially with the desired order of the polynomial which is impractical for high-dimensional data.
To make this model practical, its parameters can be reduced by making a) low-rank assumptions, and b) sharing factors between the decompositions of each parameter tensor, which accounts to jointly factorize all the parameter tensors $\bmcal{W}^{[n]}$. To illustrate this, a certain form of  CP decomposition with shared factors among low-rank decompositions of weight tensors \cite{chrysos2020poly}  is selected.  Concretely, the parameter tensor $\bmcal{W}^{[n]}$ is factorized using $n$ factor matrices. The first and the last factor matrices (called $\bm{U}\matnot{1}$ and $\bm{C}$ respectively) are shared across all parameter tensors. Then, depending on the interactions between the layers we want to forge, we can share the corresponding factor matrices. That results (see  \cite{chrysos2020poly, 9353253} for detailed derivation) in a simple recursive relationship, that can be expressed as: 
\begin{equation}
    \boutvar_{n} = \Big(\bm{U}\matnot{n}^T \binvar \Big)* \boutvar_{n-1} + \boutvar_{n-1},
    \label{eq:prodpoly_model1}
\end{equation}
for $n=2,\ldots,N$ with $\boutvar_{1} = \bm{U}\matnot{1}^T \binvar$ and $\boutvar = \bm{C}\boutvar_{N} + \bm{\beta}$. The parameters $\bm{C} \in \realnum^{o\times k}, \bm{U}\matnot{n} \in  \realnum^{d\times k}$ for $n=1,\ldots,N$ are learnable with $k \in \naturalnum$ the rank of the CP decomposition. 

\begin{figure}[!h]
    \centering
    \includegraphics[width=1\linewidth]{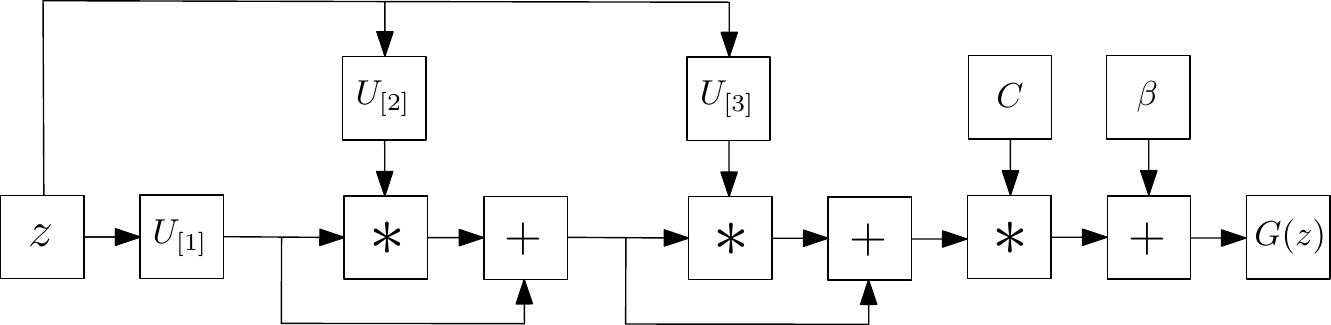}
\caption{Schematic illustration of the polynomial expansion for third order approximation.}
\label{fig:prodpoly_model1_schematic}
\end{figure}

Each recursive call in \eqref{eq:prodpoly_model1} expands the approximation order by one. Then, the final output $\boutvar$ is a polynomial of $N^{th}$ order. The recursive relationship of  \eqref{eq:prodpoly_model1} enables us to implement the polynomial using a hierarchical neural network, where the approximation order defines the depth of the network. A schematic assuming a third-order expansion ($N=3$) is illustrated in Fig.~\ref{fig:prodpoly_model1_schematic}. The tensor decomposition selected and the sharing of the factors have a crucial role in the resulting architecture. As aforementioned,  by selecting different decompositions, new architectures can emerge~\cite{chrysos2020poly}, while additional constraints can be placed on the structure, e.g., shift invariance by allowing factor matrices to implement convolutions (e.g., via im2col operator or a circulant operator).  

Higher-order polynomials can be represented as a product of lower-order polynomials, which have been extensively studied for both univariate and multivariate cases. Berlekamp's algorithm~\cite{berlekamp1967factoring} was one of the first for successfully factoring univariate polynomials over finite fields; the Cantor–Zassenhaus algorithm~\cite{cantor1981new} has since become more popular. Factorizing multivariate polynomials as a linear combination of univariate polynomials has recently been studied using tensor decompositions~\cite{dreesen2015decoupling}. The work of \cite{comon2015polynomial} demonstrates that the aforementioned factorization of multivariate polynomials is related to the joint CP decomposition. 

In deep polynomial networks, instead of using a single polynomial, also products of polynomials can be used \cite{chrysos2020poly}. That is,
the higher-order multivariate polynomial function is expressed as a product of low-order multivariate polynomials. This approach
enforces variable separability in two levels. First, in the parameters of low-order polynomials via tensor decomposition as aforementioned, and second by allowing approximation as a product of simpler functions. The product of polynomials increases the expansion order without increasing the number of layers significantly.

Furthermore, the original attention mechanism~\cite{bahdanau2014neural} is focused on neural machine translation, yet it expresses multiplicative interactions. The idea is to capture long-range dependencies that might exist in a temporal sequence. This was later extended to attention mechanisms for image and video data~\cite{wang2018non}.  Multi-head attention learns different matrices per head; concatenating those matrices, a third-order tensor is obtained. The authors utilize a CP decomposition to both reduce the parameters and reduce the redundancy in the learned matrices \cite{cordonnier2020multi}. In
 \cite{babiloni2020tesa}, where the authors extend self-attention to capture not only inter- but also intra-channel correlations by constructing an appropriate tensor such that the variance of spatial-and channel-wise features is regularized.


\section{Applications in Computer Vision}
\label{sec:applCV}
\looseness-1 In this section, we firstly discuss applications related to extracting multiple latent factors of variation from visual data (\cref{sec:applCV:multifac}). An overview of applications that focus on feature extraction with multilinear subspace learning is presented in \cref{sec:applCV:MSL}, while applications in inverse problems such as denoising and completion are presented in \cref{sec:applCV:inverse}.  In \cref{subsec:tensoreg_visualclass}, tensor regression applications in visual classification are discussed.  Finally, in \cref{subsec:deepnetsparam}, some recent applications of tensor-parametrized deep learning architectures are briefly presented

\subsection{Multiple factors analysis in computer vision}
\label{sec:applCV:multifac}

\notebook{tensorfaces}

Extracting multiple hidden factors of variation from visual data is a first fundamental step in a large number of computer vision applications. Consider the problem of view- and pose-invariant visual object classification, where one needs to classify visual objects captured across different perspectives and poses. To achieve this, it is required to design or learn data representations that are {\it invariant}  to view and pose changes, and at the same time, capture discriminant information regarding the object's category.  Structure from motion (which aims to estimate 3D structures from two-dimensional image sequences) is another computer vision task where disentanglement of multiple factors is required. More specifically, one needs to disentangle shape and motion factors and subsequently exploit them in reconstructing 3D visual objects. Face editing \cite{Wang2019AnAN,abrevaya2018multilinear} and talking faces synthesis \cite{kefalas2020speech} are further problems requiring the manipulation of multiple latent variables.

\looseness-1Multiple factors analysis via tensor methods is achieved using two different approaches. The first and historically oldest approach flattens visual data to vectors, which are then arranged into a tensor according to the factors involved in their formation to prepare the data for subsequent analysis. The second, more recent approach assumes a tensor-structured latent space in deep autoencoder-based architectures \cite{kingma2013autoencoding}).

The TensorFaces~\cite{vasilescu2002tensorfaces} is probably the most popular method in the first category and has been widely employed to analyze visual data of known sources of variation. For instance PIE~\cite{sim2002pie} and Multi-PIE~\cite{gross2010multipie} databases contain a number of people (i.e., multiple identities) captured under different poses and illuminations, displaying a variety of facial expressions. Likewise, the Weizmann face dataset \cite{moses1996weizmann} consists of $28$ subjects in $5$ viewpoints, $4$ illuminations, and $3$ expressions. Following the TensorFaces framework, such datasets are represented by a tensor with order equal to the number of sources of variation, and vectorized images are arranged into its fibers according to the labels that describe the sources of variation. The  Weizmann dataset, for example, 
would represented by a $28\times5\times4\times3\times7943$ tensor $\mytensor{X}$, where each image consists of $7943$ pixels.
TensorFaces uncover the latent factors of variation by seeking a decompostion of the form:
\begin{equation}\label{eq:tensorfaces_ex}
\mytensor{X} = \mytensor{B} \times_1 \mymatrix{U}_{people} \times_2 \mymatrix{U}_{views} \times_3 \mymatrix{U}_{illumns} \times_4 \mymatrix{U}_{expres} \times_5 \mymatrix{U}_{pixels},
\end{equation}
where matrices $\mymatrix{U}$ capture different latent factors of variation and $\mytensor{B}$ corresponds to multilinear mixing tensor that entangles the factors of variation. Such factor matrices are employed next for extraction of invariant features.
Similar analysis has also been done in analysis of motion signatures \cite{vasilescu2002motion} and gait sequences \cite{lee2005gait},  disentangling pose and body shape \cite{hasler2010multilinear, chen2013tensor}, 3D head pose estimation \cite{derkach2019tensor}, face transfer  \cite{vlasic2006face}, appearance-based tracking \cite{zhang2011visual, shao2004appearance}, structure from motion, and optical flow \cite{ramachandran2010fast,liu2003accurate}. A compositional variant of TensorFaces has also been proposed ~\cite{vasilescu2019compositional}. TensorFaces and related approaches \cite{Qiu_2015,vasilescu2019compositional} are supervised learning models and hence suitable for analysis of visual data collected in tightly controlled conditions.

Unsupervised variants of TensorFaces have been proposed in
\cite{wang2018disentangling,wang2017visualstructure},\cite{vasilescu2007multilinearprojection}
(extending \cite{tenenbaum2000separating} and \cite{tang2013_t_analyzers}). Such methods uncover multiple latent factors of variation from vector data samples in an unsupervised setting (i.e., without the presence of labels). It is only assumed that the number of sources of variation is known. More specifically,
let $\bm{X} \in \mathbb{R}^{d \times N}$  be a matrix of observations, where each of the $N$ columns represents a vectorized image of $d$  pixels. Each image is assumed to be formed by the entanglement of  $M-1$ different, yet unknown sources of variation as:
\begin{equation} \label{eq0}
\myvector{x}_i = \mytensor{B} \times_2 \myvector{u}^{(2)}_i \times_3  \myvector{a}^{(u)}_i \times_4 \dots \times_m \myvector{u}^{(M)}_i.
\end{equation}
Each factor $\myvector{u}_i^{(m)}$ accounts for a source of variability in the data, while the multiplicative interactions of these factors through tensor $\mytensor{B} \in \mathbb{R}^{d \times K_2 \times \cdots \times K_M}$  emulate the entangled variability giving rise to the rich structure of image $\myvector{x}_i$. For a set of images represented by the observation matrix $\bm{X}$, the above model is written in matrix form as \cite{wang2018disentangling,wang2017visualstructure}:
\vspace{-6pt}
\begin{equation} \label{eq1}
\mytensor{X} = \mymatrix{B_{(1)}} ( \mymatrix{U}^{(2)} \odot \mymatrix{U}^{(3)} \cdots \odot \mymatrix{U}^{(M)}) , 
\end{equation}
where $\mymatrix{B_{(1)}} \mathbb{R}^{d  \times K_2 \cdot K_3 \cdots K_M} $ is the mode-$1$ unfolding of $\bm{\mathcal{B}}$ and  $\{ \bm{U}^{(m)} \}_{m=2}^M \in  \mathbb{R}^{K_m  \times N}$ gathers the variation coefficients for all images across $M-1$ modes of variation. Observe that this model is different from the Tucker decomposition and does not require labels for the data tensor formation, in contrast to TensorFaces.
To find the unknown latent factors and the mixing operator,  the reconstruction error $ \Vert \mytensor{X} - \mymatrix{B_{(1)}} ( \mymatrix{U}^{(2)} \odot \mymatrix{U}^{(3)} \cdots \odot \mymatrix{U}^{(M)} ) \Vert_F^2$ is minimized. Having found the disentangled modes of variation, they can subsequently be used to drive shape from shading, face editing (expression transfer), and estimation of surface normals \cite{wang2018disentangling,wang2017visualstructure}
and viewpoint and illumination invariant face recognition \cite{vasilescu2007multilinearprojection}.

In the second family of methods, the multilinear structure is imposed on the latent space of deep autoencoder-based models. More specifically,  \cite{Wang2019AnAN,abrevaya2018multilinear, hong2020adaptive}, employ CNN-based encoders to obtain a set of $M-1$ latent vector representations $\{\myvector{u}_i^{(m)}\}_{m=2}^M$ for each image using label information. Each factor accounts for a source of variability in the data, while the multiplicative interactions of these factors emulate the entangled variability using (\ref{eq0}). The entangled latent variable is fed into a decoder network that reconstructs each image. In this setting, the autoencoder parameters and the tensor $\mytensor{B}$ are trained end-to-end. Such models can be used then for image editing (e.g., face editing) via manipulating the latent factors of variation.
A similar parameterization of the latent space of conditional deep variational autoencoders has been employed in \cite{markos} allowing to generate images formed by combinations of sources of variation that have not been observed in training data.

\subsection{Dimensionality reduction via tensor component analysis}
\label{sec:applCV:MSL}

\looseness-1
Tensor component analysis such as tensor extensions of PCA and LDA (cf. \cref{sec:replearning:tensorcomponent}) and decompositions have found a surge of applications in dimensionality reduction and features extraction from tensor-valued visual data. Subsequently, the extracted features are utilized for downstream tasks such as classification, using learning models such as nearest-neighbors, SVMs, or neural networks.
In {\it human sensing} applications, tensor component analysis methods are applied directly on face or body silhouette images, and the extracted features are used for face recognition, and body/gait analysis \cite{lu2008mpca,tao2007general,lu2007uncorrelatedGait,lu2007boostingGait,lu2006mpcaRecognition,lu2009boostingDisc}. Besides modeling variation on texture, tensor component analysis models, such as the HOSVD, have been used to build statistical shape models in the context of Active Appearance Models (AAM) \cite{macedo2006expression,feng2019unified} instead of the linear model typically used~\cite{fast_aam_2016}.
Furthermore, the Tucker decomposition has been used for image compression  \cite{inoue2005dsvd,ballester2020tthresh}.  

In the domain of {\it hyperspectral image analysis}, tensor component analysis methods can be used to reduce the statistical redundancy resulting from high interband correlations \cite{richards1999remote}. For instance, in \cite{zhang2012tensor}, discriminant features are extracted for classification by applying the tensor discriminative locality alignment (TDLA) method
onto a data tensor consisting of hyperspectral image patches.

Variants of tensor networks and decompositions have been further employed for tasks such as tumor and legion detection from visual data \cite{selvan2020tensor,papastergiou2018tensor}, showing benefits both in terms of performance and computational cost.  Unsupervised coupled matrix-tensor factorization methods have also been proposed towards identifying coherent brain regions across subjects that activate when similar stimuli are provided \cite{papalexakis2014turbo}.

\subsection{Tensor methods in inverse problems}
\label{sec:applCV:inverse}
\looseness-1 Inverse problems in imaging span a wide range of tasks, such as image restoration via denoising and completion of missing visual data (imputation). By representing visual as tensors, models addressing inverse problems exploit low-rank tensor decompositions (\cref{sec:replearning:tensordecomp}), or rely on robust variants (Sections \ref{sec:replearning:robust} \ref{sec:replearning:diclearn}) to estimate a low-rank tensor which forms a basis from which missing data can be imputed or denoised. Along this line of research, several methods for image and video inpainting and/or denoising have been developed by estimating the low-rank tensor using either nuclear norm-based regularizers, or by assuming an explicit low-rank tensor structure described by decompositions such CP, Tucker,  Tensor Ring, or TT decomposition \cite{ravindran2018video,yuan2019tensor,zdunek2020image,hu2016twist,liu2012tensor,hu2015new_lowrank,Gandy_2011_low_n_rank,liu2014generalized_HOOI_completion,bo2015provable}.  For the same task, methods based on tensor-structured separable dictionary learning (Section \ref{sec:replearning:diclearn}) have also been proposed \cite{bahri2019robustKC,bahri2017robustKDC,Hawe_2013_Separable}.
In the context of hyperspectral image analysis, data are corrupted with more complex noise distributions specific to the domain, such as impulse noise and stripings. To denoise, such data, regularized variants of the CP decomposition have been employed
 \cite{xue2019nonlocal,liu2012denoising,fan2017hyperspectral}. Experimental results indicate increased stability and accuracy compared to methods that do not exploit the tensor structure of the data.
 
In \cite{zhou2016linked}, several extensions of multilinear component analysis methods are presented, emphasizing the discovery of common features shared by multiple data views - such as common and individual feature analysis, with applications in MRI denoising and completion. 
\subsection{Tensor regression for visual classification}
\label{subsec:tensoreg_visualclass}
\looseness-1
Low-rank tensorregression models are well-suited for classification of large-scale visual data, such as face and body images and videos, as well as hyperspectral and medical images. In \cite{guo2012tensor},  low-rank tensor regression models with automatic rank selection are applied on problems such as face classification and pose estimation, while in \cite{paredes2013multilinear}, a multilinear multitask learning framework is used for pain estimation from videos.  Low-rank quantile tensor regression is also utilized in \cite{lu2020high} and applied for crowd counting on image datasets.   In \cite{makantasis2018tensor},  tensor logistic regression is employed towards  land-cover classification from 
hyperspectral data. It effectively reduces the number of parameters compared to methods that treat hyperspectral images as vectors from $C\prod_{i=1}^N {I_i}$ to $C\sum_{i=1}^N I_i$, where $C$ indicates the number of classes and $N$ is the order of the tensor describing the data. Classification of fMRI images may relate to several target variables, for example, detecting and identifying disease and brain damage and human brain activity that can be associated with stimuli context.  A variety of tensor methods has been used to this end, such as tensor extensions to SVM \cite{he2014dusk} and regularized low-rank tensor regression models \cite{zhou2013tensor,song2017multilinear,batmanghelich2011regularized}.

\subsection{Deep networks parametrization using tensor decomposition}
\label{subsec:deepnetsparam}

The vast majority of the deep learning models discussed in \cref{sec:deeparch} have been introduced in the context of image classification, and their effectiveness have been studied in large-scale standard benchmark datasets such as ImageNet. Neural network compression using CP decomposition has also been applied on land-cover classification  ~\cite{makantasis2018tensor}.
 Furthermore, CNNs with tensor regression layers have also been used for various applications such as affect estimation from faces~\cite{mitenkova2019valence,kossaifi2020factorized}. Randomized TRL has been employed in age prediction from MRI scans \cite{kolbeinsson2020stochastically}, significantly improving the results with respect to state-of-the-art algorithms.

Mobilenet V1~\cite{mobilenet} and Mobilenet V2~\cite{mobilenet-v2} are seminal light-weight architectures allowing mobile and embedded vision applications. Interestingly, these architectures can be interpreted as special cases of Kruskal and Tucker convolutions described in Section~\ref{subsec:parametrizingconvl}. The interested reader is referred to~\cite{kossaifi2020factorized,hayashi2019exploring} for details.

Parametrizing one or several layers with a low-rank tensor factorization has several advantages, namely  robustness to noise, regularizing effect, computational speedups, and reduction in the number of parameters.  Another advantage is conceptual: the structure induced by the factorization can be used to adapt a model to different domains or efficiently share information between tasks. For instance, when applying a Tucker decomposition to a tensor, we obtain a latent subspace represented by the core tensor. We can project to and from that space using the factors of the decomposition, which are typically projection matrices. This enables various applications, including cross-task and cross-domain incremental learning, typically by allowing weight sharing through joint factorization.
\cite{yang2017deep} propose weight sharing 
in multitask learning and~\cite{chen2017sharing} 
propose sharing residual units.
Most recently, \cite{bulat2019incremental} proposed to learn a task agnostic core, which is shared between tasks and domains. This shared subspace can be specialized to the new task or domain by learning a new set of task-specific projection matrices.

\section{Challenges and Practitioner's Guide} 
\label{sec:challenges}
\looseness-1

Tensor methods are handy tools for learning from multi-dimensional data and a convenient framework for infusing more structure in deep neural networks, leveraging redundancy to obtain more compact models, and increasing robustness. 
However, some challenges may currently prevent their wider adoption, especially in the context of deep learning. This section highlights such significant challenges and guides practitioners on how to tackle them in practice.

\textbf{Rank selection}. Most tensor methods discussed in this paper require knowledge or estimation of the rank of the decomposition. However, determining the rank of a tensor is, in general, NP-hard~\cite{haastad1990tensor,hillar2013most} and usually heuristics are employed in practice. While the choice of exact rank within a deep framework is less crucial as all the parameters are trained end-to-end jointly; rank selection remains a significant challenge.

In practice, one can start with selecting a rank such that $90\%$ of the parameters are preserved and incrementally reduce the rank, e.g., through multi-stage compression~\cite{Gusak_2019_ICCV}. When starting from a pre-trained networks, it is also possible to approximate the rank by applying  Bayesian matrix factorization~\cite{NIPS2012_26337353} to the unfolding of the weight tensor to get an approximation of the rank~\cite{yong2016compression,Gusak_2019_ICCV}. Alternatively, a Lasso-type penalty can be added to the loss function to automatically set some of the components to zero~\cite{kossaifi2020factorized}.

\textbf{Which decomposition to use?} Another significant open question in tensor methods is the choice of decomposition. While there are heuristics developed through intuition and experimental work, there is no principled way to choose the most appropriate decomposition.  

In practice, as a rule of thumb, the Tucker decomposition is well suited for subspace learning as it allows to project to and from the latent subspace represented by its core. The CP decomposition is well suited for learning interpretable latent factors; however, it is less suited for decomposing tensors which have some very large modes and some smaller ones, since the same rank is used for all dimensions. The Tensor-Train decomposition is best when the goal is to achieve very high compression ratios, as it disentangles the dimensions. In fact, it may be desirable to jointly model dimensions if we want to preserve the correlations between them~\cite{novikov2015tensorizing}.

Similar intuition applies to the choice of  tensor decomposition in the context of deep learning, albeit with some additional considerations. A CP factorization can be seen as a MobileNet-v2 block, where the rank is equal to the product of the expansion factor with the number of input channels. This means that for achieving speedups, one should choose the rank accordingly, e.g., to be a power of $2$. CP is a beneficial choice when combined with transduction to train, e.g., on a static domain such as images, and then generalize to videos~\cite{kossaifi2020factorized} during inference. A Tucker convolution gives more flexibility by allowing each mode to have a different rank, but the rank chosen along the input and output channels should also be chosen with speed in mind. For efficient learning of convolutional LSTMs in higher-orders, convolutional tensor-train decompositions have proven successful~\cite{NEURIPS2020_9e1a3651}.

As we established in Section~\ref{sec:deeparch},
there is an equivalence between choosing the tensor decomposition applied to convolutional kernels and choosing an efficient convolutional block, and an equivalence between finding the decomposition's rank and finding the block's parameters.
Clearly, choosing the appropriate tensor decomposition and its rank depends on the tensor structure, and therefore the two problems are intertwined. However, there is currently a lack of theory to guide optimal choices and enable efficient search over model architectures.  Furthermore, auto-tensorization is of particular interest. In other words, given a (pre-trained) existing network architecture, an interesting question is how to automatically choose the decomposition and its rank to approximate the network in a compressed form optimally.

\textbf{Training tensorized deep neural networks}. 
Training tensorized deep neural networks remains a challenging task due to mainly numerical issues.  For instance, composing several tensor contractions can lead to instability, with the gradient vanishing or exploding.  This is particularly true when using lower precisions (e.g., int8) motivating the need for principled ways to initialize the factors of tensor decompositions, and the need for better normalization methods that can take into account the structure.

In practice, using full precision (i.e., `float64') can alleviate most of such numerical issues. However, in order to speed up learning and inference, modern deep learning systems often rely on reduced precision, e.g., float32, or even adopt quantization of the deep models to int16, int8, or just binary, which in turn can cause numerical stability issues. To enable robust and stable learning in such settings, one can use automatic mixed-precision software tools (available for instance through NVIDIA's Amp). Automatic mixed-precision relies on two parts: using reduced precision when this can be done without affecting performance and automatic loss scaling to preserve small gradient values. On the hardware side, better handling of mathematical operations allows for striking a balance between speed and precision, for example by using tensor floats. Tensor floats on 32 bits (TF32), for instance, use the same mantissa as half-precision (Float16) but still use an 8-bit exponent to support a larger numerical range. Alternatively, it is possible to perform the calculations in log-space. While such an approach improves numerical stability, it limits the range to positive values only. 

Finally, it should be noted that a careful initialization of the factors of the decomposition is crucial to ensure that the values of the reconstruction are in the expected range. It is well known that deep convolutional neural networks need to be carefully initialized, e.g., using the He~\cite{He_2015_ICCV} or Glorot~\cite{pmlr-v9-glorot10a} initialization. These typically use a Gaussian distribution with a zero-mean and small variance. It is possible to perform variance analysis and initialize the factors directly, so the reconstruction has zero-mean and a given variance.

\ifCLASSOPTIONcaptionsoff
  \newpage
\fi

{\small
\bibliographystyle{IEEEtran}
\bibliography{egbib}
}

\end{document}